\documentclass[fleqn,10pt]{wlscirep}
\usepackage[utf8]{inputenc}
\usepackage[T1]{fontenc}
\usepackage{bm}
\usepackage{times}
\usepackage{helvet}
\usepackage{courier}
\usepackage{times}
\usepackage{latexsym}
\usepackage{amsmath}
\usepackage{algorithm}
\usepackage{algpseudocode}
\usepackage{url}
\usepackage{multirow}
\usepackage{enumitem}
\usepackage{array}
\usepackage[symbol]{footmisc}
\usepackage{threeparttable}
\usepackage{graphicx}
\usepackage{dblfloatfix} 
\usepackage{caption}
\usepackage[labelformat=simple]{subcaption}

\usepackage{adjustbox}
\usepackage{multirow}
\usepackage{hyperref}
\newcommand{\revised}[1]{\textcolor{black}{#1}}

\title{Why Do Neural Language Models Still Need Commonsense Knowledge to Handle Semantic Variations in Question Answering?}

\author[1]{Sunjae Kwon}
\author[1]{Cheongwoong Kang}
\author[1]{Jiyeon Han}
\author[1,2*]{Jaesik Choi}
\affil[1]{Korea Advanced Institute of Science and Technology, Graduate School of Artificial Intelligence, Daejeon, Republic of Korea}
\affil[2]{Director, Explainable Artificial Intelligence Center, KAIST}

\affil[*]{e-mail: jaesik.choi@kaist.ac.kr}

\begin{abstract}
Many contextualized word representations are now learned by intricate neural network models, such as masked neural language models (MNLMs) which are made up of huge neural network structures and trained to restore the masked text. Such representations demonstrate superhuman performance in some reading comprehension (RC) tasks which extract a proper answer in the context given a question. However, identifying the detailed knowledge trained in MNLMs is challenging owing to numerous and intermingled model parameters. This paper provides new insights and empirical analyses on commonsense knowledge included in pretrained MNLMs. First, we use a diagnostic test that evaluates whether commonsense knowledge is properly trained in MNLMs. We observe that a large proportion of commonsense knowledge is not appropriately trained in MNLMs and MNLMs do not often understand the semantic meaning of relations accurately. In addition, we find that the MNLM-based RC models are still vulnerable to semantic variations that require commonsense knowledge. Finally, we discover the fundamental reason why some knowledge is not trained. We further suggest that utilizing an external commonsense knowledge repository can be an effective solution. We exemplify the possibility to overcome the limitations of the MNLM-based RC models by enriching text with the required knowledge from an external commonsense knowledge repository in controlled experiments.
\end{abstract}
\begin{document}

\flushbottom
\maketitle

\thispagestyle{empty}

\section{Introduction}
One of the long-standing problems in natural language processing (NLP) is to teach machines to effectively understand language and infer knowledge \cite{winograd1972understanding}. 
% RCQA Change
In NLP, reading comprehension (RC) is a task to predict the correct answer in the associated context for a given question. RC is widely regarded as an evaluation benchmark for a machine's ability of natural language understanding and reasoning \cite{richardson2013mctest}.
%In NLP, reading comprehension (RC) is a task to predict the correct answer in the associated context for a given question. RC is widely regarded as an evaluation benchmark for a machine's ability of natural language understanding and reasoning \cite{richardson2013mctest}.

Neural language models (NLMs) that consist of neural networks to predict a word sequence distribution have widely been utilized in natural language understanding tasks \cite{radford2018improving}. In particular, masked neural language models (MNLMs) including \textit{Bidirectional Encoder Representations from Transformers} (BERT), which are trained to restore the randomly masked sequence of words, have recently led to a breakthrough in various RC tasks\cite{devlin2019bert}. However, the \textit{black box} nature of the neural networks prohibits analyzing which type of knowledge leads to performance enhancement and which type of knowledge remains untrained.

Recently, there are active efforts to understand which information is trained in the pretrained NLMs \cite{shi2016does,adi2017fine,perone2018evaluation, conneau2018you,Sahin2019LINSPECTORMP,hewitt2019structural, liu2019linguistic, hahn2019tabula, tenney2019you,tenney2019bert,manning2020emergent, kim2020pre}. Existing studies mainly focus on exploring whether a trained model embodies linguistic features for semantic analysis such as tense analysis \cite{shi2016does, conneau2018you} and named entity recognition (NER) \cite{Sahin2019LINSPECTORMP,liu2019linguistic, tenney2019bert}, and for syntactic analysis such as part-of-speech tagging \cite{shi2016does,Sahin2019LINSPECTORMP, liu2019linguistic, tenney2019bert}, chunking \cite{hahn2019tabula} and parsing \cite{Sahin2019LINSPECTORMP,liu2019linguistic, tenney2019you, tenney2019bert, hewitt2019structural, manning2020emergent, kim2020pre}. One common approach for the linguistic probing is to verify the existence of simple linguistic features by training simple classifiers upon the MNLMs for each task \cite{conneau2018you}.
%\jaesik{Please cite at least 10 papers here for each different task.} 
% Rechcek
%On the other hand, it is observed that the behavioral patterns of the BERT's self-attention heads capture syntactic information well, even without additional training on the linguistic features \cite{clark2019does}.
%In contrast,  \cite{clark2019does} finds that one can well capture syntactic information by observing the self-attention heads' the behavior patterns of the BERT.

Commonsense knowledge, defined as `information that people are supposed to know in common\cite{nilsson1998artificial}', is known to be another essential factor for natural language understanding and reasoning in the RC task \cite{mihaylov2018knowledgeable}. A recent study shows how to attain commonsense knowledge from pretrained MNLMs without additional training procedures \cite{davison2019commonsense}. However, to the best of our knowledge, detailed analysis on which type of knowledge is trained and untrained in the MNLMs has not yet been thoroughly examined and clearly discovered.

The focus of our paper is to verify how much the MNLM-based RC models answer or process the complicated RC tasks by understanding semantic relations among the words. To answer this problem, we raise the following questions regarding the semantic understanding of MNLMs:

\vspace{-0.3em}
\begin{enumerate}[itemsep=-0.1em]
    \item Do MNLMs understand various types of commonsense knowledge, especially relations of entities? (Section~\ref{sec:knowledge_probing_test})
    \item Do MNLMs distinguish some semantically related relations well? (Section~\ref{sec:synonym and antonym})
    %\item How do MNLM-based RC models solve problems across different levels of difficulty? (Section~\ref{sec:difficulty_word_overlap})
    \item \revised{What are the challenging RC-task questions for the MNLM-based RC models? (Section~\ref{sec:difficulty})}
\end{enumerate}
\vspace{-0.3em}
\revised{To answer Questions 1 and 2, we introduce a \textit{knowledge probing test} designed to analyze whether an MNLM understands structured commonsense knowledge such as semantic triples in an external repository, specifically ConceptNet \cite{speer2017conceptnet}. Experimental results on the knowledge probing test reveal that MNLMs understand some types of semantic knowledge. However, unexpectedly, we also observe that MNLMs have a lot of missing or untrained knowledge, and thus cannot precisely distinguish simple semantic concepts such as opposite relations. In addition, we can notice that when fintuning MNLMs as a commonsense knowledge base, not only probing performance but also models help to distinguish opposite relations.}
%\jaesik{even some opposite relations $\rightarrow$ even simple semantic concepts such as opposite relations}.

%For Questions 3 and 4, we first define the difficulty of an RC problem with the words overlapped between the context and the question.\jaesik{Please rewrite this sentence} Then, we analyze how the MNLMs perform on different levels of difficulty and investigate which type of questions be critical limitations of the current MNLMs. As a result of the analyses, we observe that the lexical variation is a crucial determinant in the difficulties of the RC task. In addition, we clarify that the questions that require commonsense knowledge are challenging for the MNLM-based RC models.\jaesik{Please clarify what commonsense knowledge here.}

For Question 3, we first explore a possible factor for determining the difficulty level of RC questions. Herein, we postulate that the lexical variation between a question and a context is the factor. Indeed, we observe that the lexical variation correlates with the difficulty level of RC questions. On top of that, we show a difficult question may require additional inference procedures to solve it. 
Inspired by these observations, we categorize RC questions into six question types based on the required information to solve the questions. Then, we clarify that the questions, which require commonsense knowledge to solve them, are still challenging for the existing MNLM-based RC models.

% we classify types of the questions and clarify that commonsense knowledge type questions, 
% require additional inference to solve the questions, are still challenging for current MNLM-based RC models.
%\jaesik{define commonsense knowledge} 

% Based on the above results, we propose a solution that can ameliorate the limitations of the current MNLMs by integrating knowledge originated from an external commonsense repository. To verify our solution, we conduct two experiments. 
% Firstly, we manually convert words in the question to integrate the knowledge that is required to solve the problem.
% Secondly, we propose a neural network architecture that complements MNLMs with the external commonsense repository. In both experiments, we observe that MNLMs could be complemented by integrating commonsense knowledge. 

Finally, by analyzing the result of the knowledge probing test and the observed frequency of subject and object entity pairs, we find that MNLMs' way of learning knowledge is substantially affected by the conditional probability of the entity pairs. Based on this finding, we explain why an external knowledge repository is needed to overcome the limitations of MNLMs. In addition, we conduct controlled experiments to show that an external knowledge repository can be helpful to overcome the limitation of MNLM-based RC models. In the experiments, we enrich the incorrectly predicted questions with required commonsense knowledge from the external knowledge repository. The results show that the incorrectly predicted questions are properly answered by MNLM-based RC models without any change of the models.

%In addition, we conduct controlled experiments . We suggest a way to enrich text by using a simple rule and knowledge triples extracted from an external knowledge repository.
%From the controlled experiments, we attest that the limitation of the MNLM-based RC models can be allayed by complementing text with knowledge triples.
%\jaesik{a controlled experiment $\rightarrow$ controlled experiments} 

Main contributions in this paper are as follows:
\vspace{-0.3em}
\begin{itemize}[itemsep=-0.1em]
    \item From the experimental results of the knowledge probing test on the commonsense knowledge of ConceptNet, we decisively observe that MNLMs have a lot of missing or untrained knowledge. 
    \item By analyzing the results of the MNLM-based RC models, we observe a new finding that the existing MNLMs have critical limitations when solving questions requiring commonsense knowledge. 
    \item To the best of our knowledge, it is the first approach to empirically explain the fundamental reasons why a large portion of commonsense knowledge is not learned by the existing MNLMs and discuss why external commonsense knowledge repositories are still required. Moreover, we show that MNLMs can be complemented by integrating an external commonsense knowledge in the actual RC-task.
\end{itemize}
\vspace{-0.3em}
The paper is organized as follows. Section~\ref{sec:background} briefly describes the notions required to readily understand our paper. Section~\ref{sec:knowledge_probing} introduces our knowledge probing test and demonstrates the results of the test. Then, we present the performance of the MNLM models on different difficulties of RC problems in Section~\ref{sec:difficulty}. Section~\ref{sec:discussion} discusses the reasons why external commmonsense repositories are needed and suggests a possible direction to overcome the limitation of the existing MNLM-based RC models. Finally, the conclusion is stated in Section~\ref{sec:conclusion}.
%\jaesik{define commonsense knowledge} 
\section{Background}
\label{sec:background}
\subsection{Masked Neural Language Models}
We consider an MNLM that calculates the probability distribution over the sequence of words with a neural network. We mainly discuss three types of MNLMs, BERT and its two variations: 1) BERT, 2) \textit{A Robustly Optimized BERT Pretraining Approach} (RoBERTa) \cite{liu2019RoBERTa} and 3) \textit{A Light BERT} (ALBERT) \cite{lan2019albert}. Detailed structural information on the experimental models used in this paper is described in Table~\ref{tab:model_structures}. 
%Appendix~\ref{apx:detail_information_of_the_experimental_mnlms}.
%Especially, we mainly discuss following three MNLMs: 1) BERT, 2) \textit{A Robustly Optimized BERT Pretraining Approach} (RoBERTa) \cite{liu2019RoBERTa} and 3) \textit{A Light BERT} (ALBERT) \cite{lan2019albert}. In this section, we will briefly introduce models that used in this paper, 

BERT is made up of the transformer architecture \cite{vaswani2017attention}. The model has $L$ transformer layers. Each layer comprises $S$ self-attention heads and $H$ hidden dimensions. In addition, the input of the model is a conjunction of two sentences $A_1,...,A_N$ and $B_1,...,B_M$, where each token is split into WordPiece \cite{schuster2012japanese} with a vocabulary of 30,000 tokens. Special delimiter tokens `[CLS]' and `[SEP]', which indicate `classification token' and `sentence separate token' respectively, are adopted to integrate two sentences into the ensuing input: 
\begin{gather*}
[CLS],A_1,...,A_N,[SEP],B_1,...,B_M,[SEP]
\end{gather*}
By adding delimiter tokens, the final number of tokens of the input sequence should be $N+M+3$. Two objectives are used to pretrain BERT model: 1) the masked language model (MLM) loss and 2) the next sentence prediction (NSP) loss. Different from traditional language models that optimize the likelihood of the next word prediction, BERT is optimized with the MLM loss. With the MLM loss, tokens in the text are randomly masked with a special token `[MASK]' at a designated proportion, and BERT is optimized with the cross-entropy loss to predict the correct tokens for the masked input.
On the other hand, NSP loss is a binary classification loss to determine whether sentences $A$ and $B$ are naturally observed in the data sequence. In a positive example, $A$ and $B$ are consecutive sentences. In contrast, $B$ is randomly selected from another document in a negative example.
We adopt two BERT models (BERT$_{base}$ and BERT$_{large}$) to investigate the results. The pretrain data of these models include an integration of two different corpora (English Wikipedia and Book Corpus \cite{zhu2015aligning}) which approximates 16GB. 

RoBERTa has the same structure of transformers with BERT. However, there are several changes to refine the original BERT. First, different from BERT that fixes masked tokens during the entire training procedure, RoBERTa changes the masked tokens through the process of learning. Next, the NSP loss of BERT is no longer used in RoBERTa. Instead, RoBERTa is trained on a single sequence of document consists of up to 512 tokens. In addition, during the training, RoBERTa uses a much larger batch size compared with BERT to reduce training time. Furthermore, RoBERTa's vocabulary uses byte pair encoding (BPE) \cite{sennrich2016neural} instead of the word piece token used in BERT. Finally, RoBERTa is pretrained with approximately 160GB of data including the BERT pretraining corpora as well as three additional corpora (CommonCrawl News dataset\cite{ccnews}, open-source recreation of the WebText corpus \cite{Openwebtext} and STORIES corpus \cite{trinh2018simple}). We utilize two RoBERTa models (RoBERTa$_{base}$ and RoBERTa$_{large}$).

ALBERT also consists of the structure of transformers. Nevertheless, some changes are adopted to amend the original BERT. First, in order to reduce the number of parameters, ALBERT decreases token embedding size and shares the parameters of attention and feed-forward networks across all transformer layers. Not only that, instead of NSP loss, ALBERT is trained with sentence order prediction (SOP) loss that predicts the natural order of two sentences $A$ and $B$ come from the same document. To compare the effects of data size, we utilize three ALBERT version 1 models (ALBERT1$_{base}$, ALBERT1$_{large}$, and ALBERT1$_{xlarge}$) pretrained with the BERT pretrain data and three ALBERT version 2 models (ALBERT2$_{base}$, ALBERT2$_{large}$, and ALBERT2$_{xlarge}$) pretrained with the RoBERTa pretrain data.

\subsection{Generative Pre-Training Models}
\revised{We also discuss Generative Pre-Training (GPT) models, which are the state-of-the-art generative NLMs. Herein, we utilize the largest models of each version of GPT models: GPT1 \cite{radford2018improving}, GPT2\cite{radford2019language} and GPT3\cite{brown2020language}. 
%Detailed information on the hyperparameters and the pretraining data of the models is reported in Table~\ref{tab:model_structures}.
}

\revised{GPT models are built on the transformer architecture. Unlike the aforementioned BERT families, GPT models are trained to predict the next token for the given sequence of words as the traditional language models do. For example, if we have a sentence $A_{1...N}$ and the token sequence $A_{1...N-1}$ is given, then the GPT models are trained to maximize the probability $P(A_N|A_{1...N-1})$ such that the token $A_N$ will be observed.}

\revised{The most significant differences between each GPT version are the number of parameters and the pretraining data as summarized in Table~\ref{tab:model_structures}. The number of parameters becomes larger in the order of GPT1 to GPT3 (GPT1: 119M, GPT2: 1.5B and GPT3: 175B), as well as the size of the pretraining data (GPT1: 4GB, GPT2: 40GB and GPT3: 570GB). Among the GPT models, GPT3 has been reported to have impressive zero-shot and few-shot inference performances in previous studies \cite{brown2020language, da2021understanding, sainz2021ask2transformers}. In this paper, we use \textit{Davinci}, which is the largest model among the GPT3 models provided by OpenAI API\cite{openai}. Note that the GPT3 API is provided for a fee and the pretrained parameters are not openly available.}

\subsection{Commonsense Knowledge Repositories}
It is important to determine an external resource where we can extract commonsense knowledge. ConceptNet, a part of an \textit{open mind commonsense} (OMCS) \cite{singh2002open} project, is a knowledge base designed to help computers understand commonsense knowledge shared by people. It has been widely exploited as a commonsense knowledge repository in previous studies \cite{wang2018yuanfudao,guan2019story,talmor2019commonsenseqa,petroni2019language,kassnerS20negated, jiang2020can,shin2020eliciting,bouraoui2020inducing}. Commonsense knowledge in ConceptNet is represented as semantic triples which constitute a subject entity, an object entity, and a relation between the entities. ConceptNet includes commonsense knowledge that originates from several resources: crowdsourcing, expert-creating, and games with a purpose. We utilize ConceptNet 5.6.0 \cite{conceptnet5.6.0} version for experiments. In this paper, we conduct the knowledge probing test on 32 relations. Detail information of the relations that we use can be found in Appendix~\ref{apx:details_on_the_templates}.

%\footnote{ConceptNet 5.6.0, \url{https://s3.amazonaws.com/conceptnet/downloads/2018/edges/conceptnet-assertions-5.6.0.csv.gz}} a semantic network 

\section{Probing Commonsense Knowledge in MNLMs}
\label{sec:knowledge_probing}
This section investigates which types of commonsense knowledge are well trained and contained in the pretrained MNLMs. 
Clarifying the knowledge included in the MNLMs is difficult for ensuing reasons. First, there is a disparity between the input format of MNLMs that is natural language and the structure of ConceptNet knowledge that is made up of semantic triples.

%For instance, the inputs of MNLMs are natural language format while each knowledge of ConceptNet has semantic triple format. In addition, due to the fact that MNLMs consist of a bunch of complicatedly intermingled model parameters, it is difficult to analyze which information is trained in the MNLMs. % \textit{black box} nature of the neural networks makes difficult to ~. % one is disparity between since each knowledge has a semantic triple form while the MNLMs have complex and intermingled model parameters. 
The Cloze test \cite{chapelle1990cloze}, known to be a reliable assessment for the language ability of a participant, is a task wherein one fills in the correct answer for the blank in the text. In the following example, ``children and \_ are opposite.'', the answer word would be `adults' rather than `kids'. To infer the correct answer, we must know not only the meaning of each word but also the semantic relation between the words. 

\revised{Currently, several studies suggest the methodologies to gauge relational knowledge from pretrained MNLMs with the Cloze test approaches \cite{davison2019commonsense, petroni2019language,kassnerS20negated, jiang2020can, bouraoui2020inducing, shin2020eliciting, zhong2021factual}. 
% single manual template: LAMA
LAMA (LAnguage Model Analysis) probe\cite{petroni2019language} is an early study of the Cloze style probing approach. Herein,  probing has been conducted by a Cloze test with manually designed templates. The results show that some factual knowledge can be recalled from the MNLMs without finetuning procedures. However, in this case, the results highly depend on the designed set of templates. To ameliorate this, LM Prompt And Query Archive (LPAQA) suggests to create a set of candidate prompts for each relation by using text-mining and paraphrasing approaches \cite{jiang2020can}. Among the multiple candidate prompts, one may select the top-K prompts. The authors claim that they can achieve higher performance when the probing results of the retrieved text and the manually designed text are ensembled. Recently, the gradient-based methods have been proposed to find the optimal relational prompts for each model \cite{shin2020eliciting,zhong2021factual}. AutoPrompt is a method to automatically generate prompts for a diverse set of tasks, based on a gradient-guided search\cite{Wallace2019Triggers}. The authors suggest to generate a prompt for each relation by adding ``trigger'' tokens that can be any token in the vocabulary. Herein, all tokens are set to `[MASK]' token in the beginning then recurrently updated to optimize the probability of the answer label of the training data. OptiPrompt is another gradient-based prompt engineering approach \cite{zhong2021factual}. Different from AutoPrompt which optimizes discretely, OptiPrompt suggests to directly optimize the input vector space. As a result, they can find real-valued inputs for extracting factual knowledge. 
However, existing methods have some drawbacks to be adopted to our experimental setting. First of all, the gradient-based prompt searching can generate semantic and grammatical inscriptions such as ``''(),ex-,Liverpool'' which are far from the natural sentences \cite{Wallace2019Triggers}. 
On top of that, as GPT3 API is provided for a fee, it requires tremendous cost to paraphrase the prompts suitable for each model, which is also the case for the ensemble-based method.
Finally, the gradient-based approaches require `white-box' access to the NLMs to calculate gradient. Thus, it is hard to be applied to the extremely large NLMs such as GPT3 which are not publicly accessible. 
Therefore, we try to conduct the knowledge probing, named as \textit{knowledge probing test}, by taking into account 1) semantic and grammatical plausibility, 2) not relying on a single prompt for each relation, 3) minimizing experimental cost, and 4) black-box setting for the parameter accessibility of the very large NLMs. }

In the knowledge probing test, we first transform a semantic triple $(s,r,o)$ into a sentence that can be used as an input to a designated MNLM. Herein, a sentence is created via predefined predicate templates collected from frequently used patterns representing particular relations in the OMCS dataset\cite{OMSC}. To be specific, for a give triple we generate masked sentence for the knowledge probing test by the following procedures. 

\revised{First of all, we find the most grammatically plausible candidate sentence for each template. To this end, from an original template, grammatically diversified sentences are generated by grammar transformation rules\cite{davison2019commonsense}. The sentence with the lowest perplexity \cite{radford2018improving} on the pretrained LM among the generated sentences is selected as the candidate sentence of a template. Then, the most semantically probable one is picked among candidate sentences originated from the original templates. Specifically, the sentence with the lowest perplexity is selected as the masked sentence of a triple. Through this process, we can create the most grammatically and semantically natural masked sentences for each triple. Herein, GPT1 is used as the pretrained LM for perplexity calculation. The detailed procedure of the grammatical transformation is described in Appendix~\ref{apx:details_on_the_grammar_transformation}. In addition, we provide several examples for the results of the knowledge probing test in Appendix~\ref{apx:qualitative_results}. The details on the original templates are presented in Appendix~\ref{apx:details_on_the_templates}.}

Our paper reports following fundamental limitations of the existing MNLMs in literature, veiled behind empirical successes of NLMs, which have not been extensively explored yet: 1) Even if MNLMs predict correct answers on knowledge triples, it may not guarantee MNLMs accurately understand attributes of subject entities, 2) MNLMs have a hard time to discern semantically related relations such as opposite relation pairs.

%In contrast, our paper reveals several fundamental limitations of the current MNLMs which are not extensively explored yet due to empirical successes of neural language models.

\subsection{Probing on Various Types of Relations}
\label{sec:knowledge_probing_test}

% We provide additional examples in Appendix~\ref{apx:additional_probabilistic_distribution}. 

The result of the knowledge probing test, we use hits@K metric \cite{bordes2013translating} that measures the ratio of correctly predicted answers, in the top K predictions, out of all true answers from the ConceptNet repository. Table~\ref{tab:hits@K} presents macro-average, an equally weighted average of the result of relations, and micro-average, a weighted average of the results of the relations according to their frequencies. We use the macro average as the main yardstick because there is a large variation in the number of examples in each relation. For example, the experimental results are greatly influenced by relations with a high proportion such as `RelatedTo' and `HasContext'. Individual results for each relation are listed in Appendix~\ref{apx:quantitative_analysis_for_probabilistic_distributions}. \revised{Note that, our knowledge probing test is in the recent research lines in that Cloze test is used. Some of the recent studies mainly focus on the positive results that NLMs are able to infer factual knowledge  \cite{davison2019commonsense,petroni2019language}. On the other hand, a recent study demonstrates that the existing NLMs predict almost the same results in a given sentence and its negated sentence \cite{kassnerS20negated}.}

% The effects of the selected templates

% NMI 
First, we analyze the effect of the size for each type of MNLM through the performance of knowledge probing test. From the experimental results in Table~\ref{tab:hits@K}, for the same type of MNLMs, the larger models generally perform better. 
\revised{Especially, the largest models of each type of MNLMs, BERT$_{large}$, RoBERTa$_{large}$, and ALBERT2$_{xlarge}$, show performances around 40 at hits@100. Moreover, the performance of GPT3 clearly surpasses that of GPT1 and GPT2, and hits@100 shows that the ability to extract factual knowledge is substantially higher than that of the other models.}
% BERT, ALBERT2, RoBERTa 설명
% ALBERT1 설명 및 해설
Meanwhile, the results of ALBERT1 models demonstrate somewhat different from those of the other models. Firstly, the overall performance is less than the other type of models. In particular, the hits@100 performance of ALBERT1 is less than 30, which is about 25\% less than the other models. Next, it can be seen that ALBERT1$_{xlarge}$ has lower performance than ALBERT1$_{base}$ and ALBERT1$_{large}$, which consist of fewer parameters and layers than those of ALBERT1$_{xlarge}$. 
% ALBERT2와 ALBERT1 비교하면서 추가 데이터의 역할 설명
However, we can also find that such issue is ameliorated in the results of the ALBERT2 models trained on larger corpus than ALBERT1 models. Therefore, from these observations, we can achieve the following inferences. First, commonsenese knowledge can be extracted from MNLMs without further fine-tuning. Next, typically a deeper and larger model shows better performance compared to smaller models. However, for models with very deep and large sizes such as ALBERT$_{xlarge}$ models, it is difficult to learn commonsense knowledge with a relatively smaller dataset. Finally, MNLMs can learn more commonsense if they are trained on the larger pretrain dataset.

% Recheck
Although the average hits@100 performances above 50\% may seem to be high, considering the average number of answers provided by ConceptNet (see Appendix~\ref{apx:details_on_the_templates}) is less than 5, the listed models cannot predict all 5 of the confident answers correctly within top 100 predicted words. In addition, large fluctuation can be found in the quantitative results for each relation. Some relations (`Entails', `AtLocation'...) show below 20\% in hits@100 while some (`NotCapableOf', `MadeOf', `ReceivesAction') show above at least 60\%.
\begin{figure*}[ht!] 
    \begin{center}
    \includegraphics[width=\linewidth]{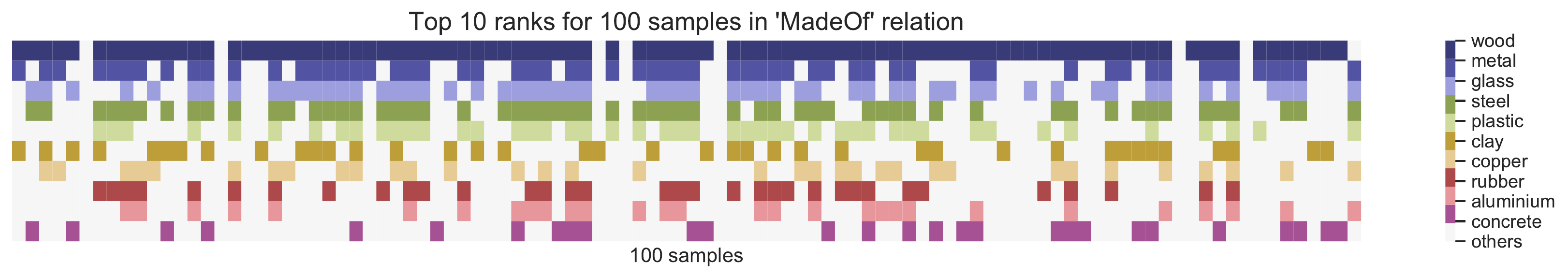}
    \end{center}
%\caption{Color coded results on the BERT$_{large}$ model prediction for 100 samples in the 'MadeOf' relation. Colors are labeled when each sample (x-axis) contain the top 10 most-frequent words. We can notice that top 10 words are redundantly observed in the high rank.}
\caption{\textbf{Color-coded results of the BERT$_{base}$ model's predictions on 100 samples in the `MadeOf' relation}. This figure shows whether each sample (x-axis) contains certain object words (y-axis) in the top 10 predictions. Each color represents the 10 most frequently observed words in the predictions on the `MadeOf' relation.}
\label{fig:made_of_color_coded}
\end{figure*} %%% Figure 9.
\begin{figure*}[ht!] 
    \begin{center}
    \includegraphics[width=\linewidth]{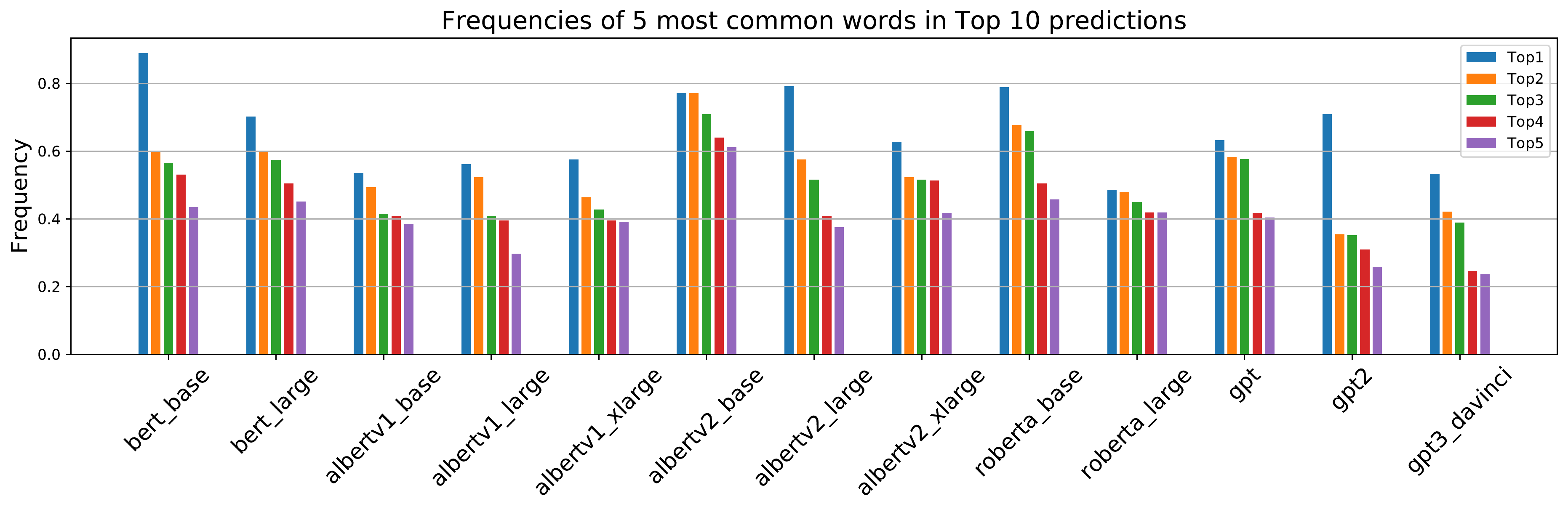}
    \end{center}
%\caption{Color coded results on the BERT$_{large}$ model prediction for 100 samples in the 'MadeOf' relation. Colors are labeled when each sample (x-axis) contain the top 10 most-frequent words. We can notice that top 10 words are redundantly observed in the high rank.}
\caption{\textbf{Frequencies of the top 5 frequently occurred words in top 10 predictions}. X-axis indicates the top 5 most commonly observed words in top 10 predictions of each model. Y-axis is a frequency ratio of the commonly observed words. }
\label{fig:topk_word_image}
\end{figure*} %%% Figure 9.

% Furthermore, we suspect that the semantic understanding of MNLMs about relations is not as accurate as expected despite the high hit ratios. `MadeOf' relation is an illustrative example. While it shows relatively high hits@10 performances in MNLMs, we commonly observe that some specific words are repeated across different subjects. We provide detailed figures in Appendix ~\ref{apx:MadeOf}. As shown in Fig.~\ref{fig:topk_word_image}, `wood', `metal' and `glass' frequently appear as high-rank predictions in more than 70\% of sampled subjects. Therefore, our observations say that the prediction tends to follow the marginal distribution of `MadeOf' relation instead of reflecting the conditional distribution of a subject. This can be problematic when those frequent words are definitely incorrect answers. For example, `wood' is actually predicted as the most probable answer for the question ``What is butter made of?'' where the human can easily notice `wood' is an inadequate answer. 

Furthermore, we conjecture that the semantic understanding of MNLMs about relations is not as accurate as expected despite the high hit ratios. An illustrative example is `MadeOf' relation which consistently shows the best performance in almost all MNLMs. Indeed, `MadeOf' relation has the highest performance in all MNLMs based on hits@10 among the relations with more than 50 examples. However, when we have a closer look at the predictions from MNLMs, it is commonly observed that some specific words are repeated across different subjects. 

Especially the BERT$_{base}$ model, which achieves the highest hits@10 in `MadeOf' relation, presents a noticeable result. Figure~\ref{fig:made_of_color_coded} shows the appearance of the 10 most frequent words, in the top 10 predictions of the BERT$_{base}$ model for 100 samples of the `MadeOf' relation. In more than 70\% of sampled subjects, `wood', `metal', and `glass' appear as high-rank predictions. Therefore, our observations say that the prediction tends to follow the marginal distribution of `MadeOf' relation instead of reflecting the conditional distribution of a subject. This can be problematic when those frequent words are definitely incorrect answers. For example, `wood' is predicted as the most probable answer for the question ``What is butter made of?'' where a human can easily notice `wood' is an inadequate answer. \revised{As Figure~\ref{fig:topk_word_image} shows, such repetition of the predicted words can be commonly observed among the MNLMs. Note that, even if the size of the model increases ($base$ < $large$ < $xlarge$; GPT1 < GPT2 < GPT3) or the model is trained on the larger dataset (ALBERT1 < ALBERT2; GPT1 < GPT2 < GPT3), the marginal prediction issue of the `MadeOf' relation is not fundamentally solved. We provide detailed information about the frequently observed object words for each model in Appendix~\ref{apx:MadeOf}.} 
%The top 10 overlapping of predicted words for each MNLM model and their observation ratio can be found in the Appendix~\ref{apx:MadeOf}.

\subsection{Probing the Relationship Between Two Opposite Relations}\label{sec:synonym and antonym}
\begin{figure}[!t]
  \adjustbox{minipage=2em,raise=-\height}{\subcaption{} \label{fig:opposite_test_move}}%
  \raisebox{-\height}{\includegraphics[width=.45\linewidth]{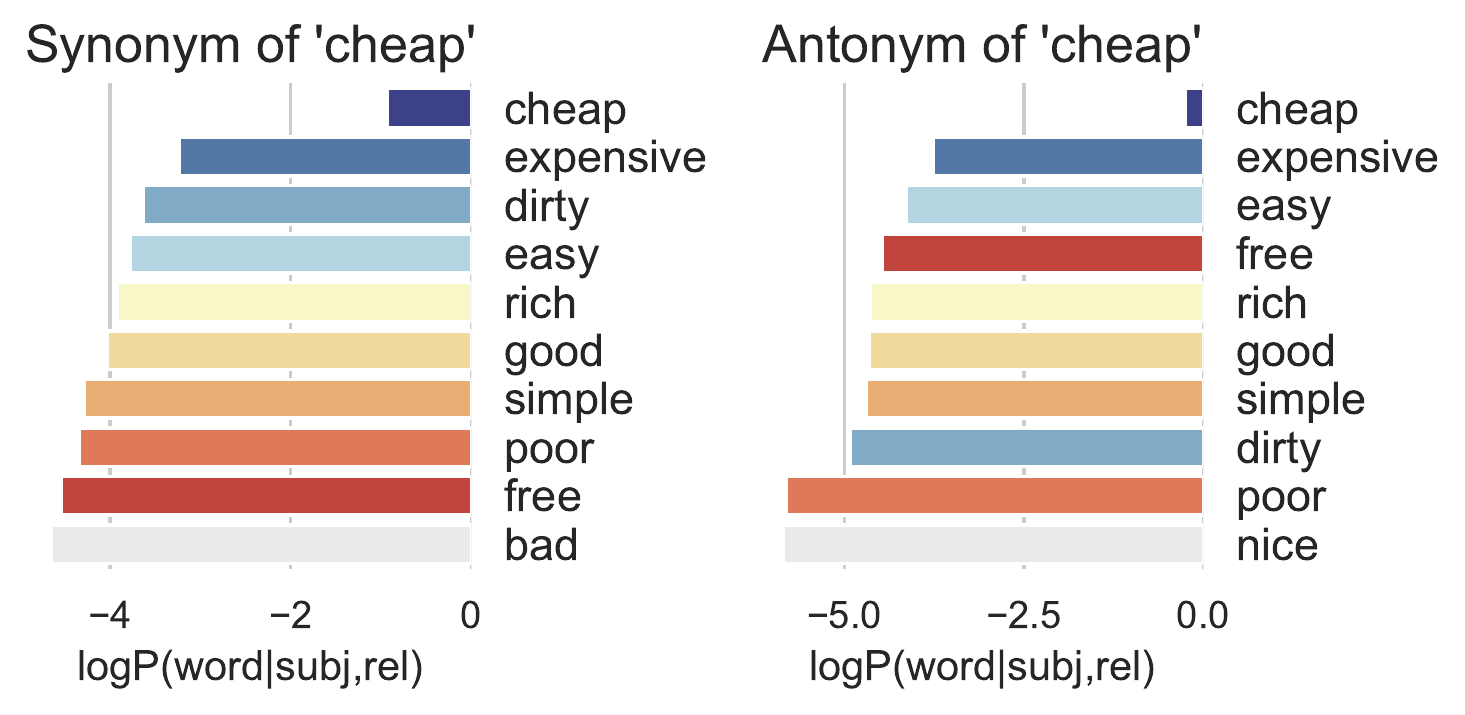}}
% 
%  \newline
  \adjustbox{minipage=2em,raise=-\height}{\subcaption{} \label{fig:opposite_test_trust}}%
  \raisebox{-\height}{\includegraphics[width=.45\linewidth]{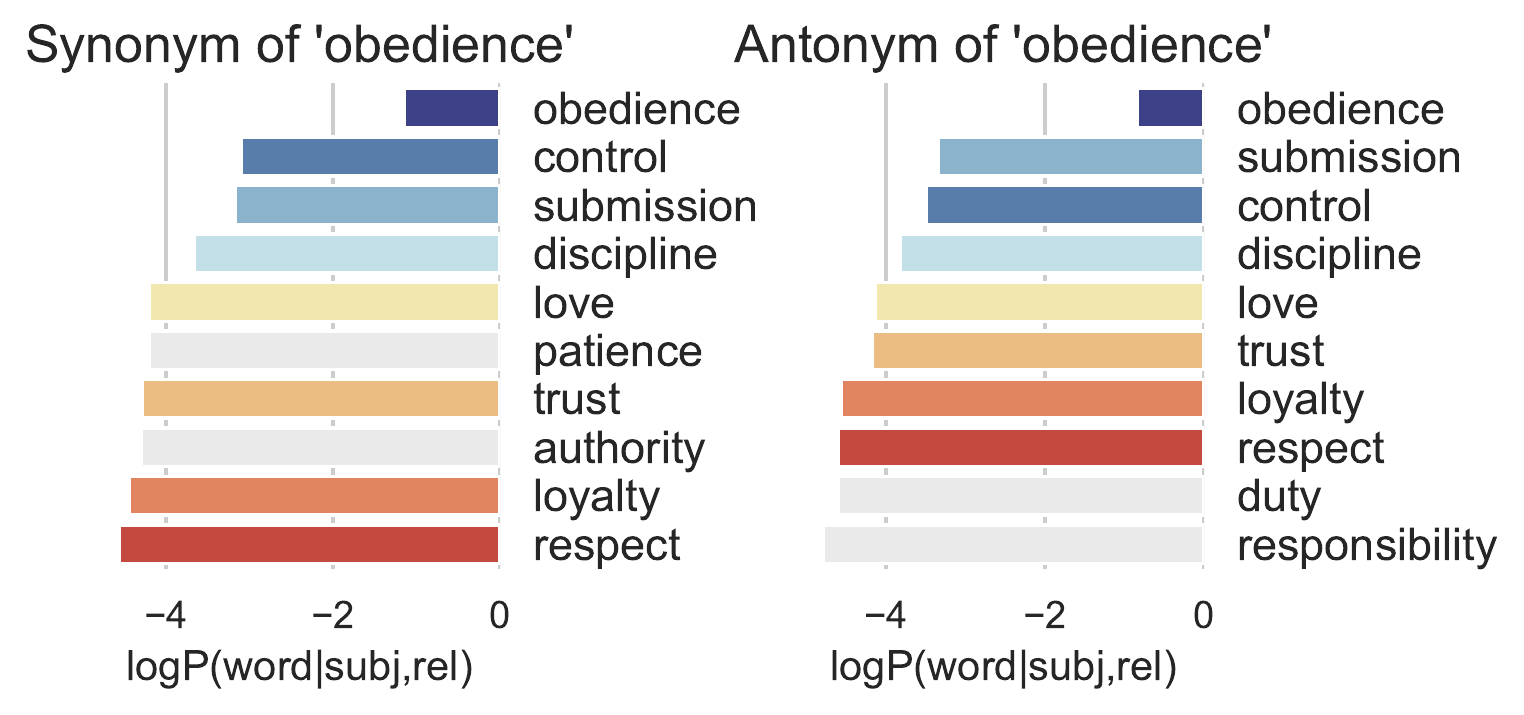}}
  \caption{\textbf{Results of the BERT$_{base}$ model on the top 10 words on the opposite relations on subject words}. \textbf{(a)} `cheap' and \textbf{(b)} `obedience'. Words commonly observed in both results are painted in the same color, and the other words are in light gray.}
  \label{fig:antonym_synonym_overlapping}
\end{figure}
\revised{So far, we have discussed the behavior of MNLMs for each relation. Here, we address the following question ``Do MNLMs precisely understand the semantic differences between relations?'' To answer the question, we compare the results from the knowledge probing test of the four following pairs of opposite relations on the same subject: `Synonym / Antonym', `HasProperty / NotHasProperty', `Desire / NotDesire' and `CapableOf / NotCapableOf.' 
Different from the previous study\cite{kassnerS20negated}, our experimental environment is more restrictively controlled that there must exist answers for both relations of an opposite relation pair to the same subject. The results for each relation of an opposite relation pair should be clearly distinguished if the MNLMs understand subtle semantic differences between the opposite relations. }

%Although our experimental setting seems similar to the previous study\cite{kassnerS20negated}, our experimental  Therefore, different from the previous study where 

%it is possible to clearly observe whether MNLMs distinguish the subtle semantic differences between opposite relation pairs.
%if MNLMs clearly distinguish the subtle semantic differences between opposite relation pairs, then the predicted results should be different. 
%although our experimental setting seems similar to the previous study\cite{kassnerS20negated}.

Figure~\ref{fig:antonym_synonym_overlapping} indicates illustrative examples in both of the opposite relations on the same subject words. Unexpectedly, there are words simultaneously predicted in the opposite relations. The quantitative results in Table~\ref{tab:overlapping_results} show that words with high probabilities of two opposite relations are common in many cases. The finding supports that the MNLMs may not clearly distinguish the subtle meaning of the opposite relations. \revised{This phenomenon is not fundamentally solved either as the size of the model grows or as the training data increases as seen in the results of the ALBERT and GPT models.}

Furthermore, in order to demonstrate that such overlapping issue is undesirable and even problematic, we measure the ratio of incorrect answers by grading the predictions with answers from the opposite relation that can be regarded as wrong answers. Table~\ref{tab:intergrade_results} summarizes the result. 
%Among the opposite relation pairs, the answer object of the `Synonym / Antonym' is incompatible while the others rarely but possibly have the identical answers. For this reason, we conduct the experiments on the `Synonym / Antonym' pair. Hits@K, in this case, can be interpreted as the incorrect rate.
Among the opposite relation pairs, the answer object of the `Synonym / Antonym' pair is incompatible while the other opposite pairs can have identical answers. For example, there is an identical answer `pregnant' for a given subject `person' and the `Desire / NotDesire' opposite relations. For this reason, we conduct the experiments on the `Synonym / Antonym' pair. Hits@K, in this case, can be interpreted as the Miss@K, which is the ratio of predicted words definitely incorrect.
% New 
% NMI

The experimental results demonstrate that MNLMs have a relatively high incorrect ratio considering it is unnecessary. In addition, as the size of the model and the training data increases, the performance of the models enhances. However, at the same time, the incorrect rate also increases generally. Thus, from these results, it is difficult for MNLMs to discern a precise difference between semantically related relationships specifically opposite relations. Additional examples of the overlapped predictions between the opposite relations can be found in Appendix~\ref{apx:additional_examples_on_overlapping}.
%As shown in Table~\ref{tab:intergrade_results}, the incorrect rate is high even when comparing the ALBERT$_{xlarge}$ with BERT$_{base}$, considering that the no-hit is desirable. In addition, as shown in Appendix~\ref{apx:quantitative_analysis_for_probabilistic_distributions}, as the size of the model increases, the performance of each relation enhances, while the incorrect rate also increases as shown in the Appendix~\ref{apx:synonym antonym grading}. Thus, we argue that MNLMs with the current training scheme do not discriminate opposite relations well.

\subsection{The Impact on the Masked Sentence Selection Procedure}
\revised{Herein, we analyze the impact of the selected templates in the reported performance of the LMs. For this, we analyze the impact of the masked sentence selection in the knowledge probing test.
More specifically, for the BERT models, we compare the probing performance of the selected masked sentence with the average probing performance of the candidate sentences derived from all original templates. The results in Table~\ref{tab:impact_of_template_selection} demonstrate that the selected masked sentences show substantially higher performance compared to the candidate sentences. From this, we can see that the selected sentences are not only the most natural among the candidate sentences, but also effective for extracting factual knowledge in general.}

\subsection{The Impact of the Fine-tuning on the ConceptNet}
\revised{The previous experimental sections are performed in a zero-shot environment without additional fine-tuning on pretrained MNLMs for the knowledge probing test. However, the distribution of per-train corpora of MNLMs has a large discrepancy from that of ConceptNet. Therefore, we need to verify that the limitations we point out above arise from the difference between the train corpora of MNLMs and ConceptNet. For this, we fine-tune BERT models on ConceptNet triples and analyze the results.}

\revised{First of all, we randomly divide all of the triples included in the knowledge probing test into 3 folds. After that, we conduct fine-tuning and evaluating of MLNMs by designating each fold as train, validation, and test sets. To be specific, to evaluate each fold, we specify train, validation, and test sets as follows: 1) Train: Fold 0, Validation: Fold 1, Test: Fold 2, 2) Train: Fold 1, Validation: Fold 2, Test: Fold 0, 3) Train: Fold 2, Validation: Fold 0, Test: Fold 1.} \revised{During fine-tuning, models are trained to predict the answer objects of a masked sentence. Note that, we set the hyper-parameters for fine-tuning of models to the default values of OptiPrompt \cite{zhong2021factual}. Finally, the performance of fine-tuned model is calculated by averaging the test performance for each fold.}

\revised{Table~\ref{tab:hits@K_finetune} is the experimental results on the knowledge probing test of pretrained BERTs and fine-tuned BERTs. After fine-tune, it can be seen that the performance is significantly enhanced ($p < 0.05$) in both BERT$_{base}$ and BERT$_{large}$ models. Especially, the performances of hits@100 show over 70\% for both models.}

\revised{Tables~\ref{tab:overlap@K_finetune} and Tables~\ref{tab:miss@K_finetune} show the results of experiments on overlapping between opposite relations for the fine-tuned models. Here, we can see that miss@1 of the fine-tuned model is significantly reduced compared to before training. It is assumed that this is because the prediction probability of the correct answer object increases as the prediction rate of the Synonym and Antonym relationships increases. Nevertheless, in the case of overalp@K, the problem is not fundamentally solved even after fine-tuning, and in the case of synonym-antonyms, it can be seen that the overlap ratio increases after fine-tuning.}

\revised{Overall, we can see that when the model is fine-tuned to ConceptNet, not only the knowledge probing performance is improved, but also the miss-overlapped ratio between some opposite relations can be reduced. These results imply the importance of a tailored dataset similar to the distribution of a target task. However, even after fine-tuning, it can be seen that the problem of overlapping is still observed in the model output results in the opposite relationship is not fundamentally solved. Detailed experimental results for each relation in the folds of the models can be found in the Appendix~\ref{apx:finetuning_on_KB}.}

\section{Which Types of Questions Are Still Challenging for MNLMs?}\label{sec:difficulty}
\revised{In recent years, MNLMs have led to breakthroughs in RC task even beating human-level performance\cite{radford2018improving,devlin2019bert,lan2019albert}. It is widely known that not only syntactic but also comprehensive semantic knowledge including commonsense knowledge is required to solve the task accurately. Then, do these results mean that the linguistic understanding of the model exceeds that of humans who can infer the correct answer based on background including commonsense knowledge? In this section, we will figure out which types of questions are still challenging for the existing MNLMs. Especially, we scrutinize the problems that require factual background knowledge, such as a ConceptNet triple, among the keywords in a question-context pair.}

% MNLMs still have incomplete commonsense knowledge even though MNLM-based RC models have outperformed previous approaches in RC tasks . 
% We assume that if there is a high lexical variation between question and context then the question is more likely to need commonsense knowledge. 
%We assume that if keywords of a question is paraphrased in a context then additional inference, includes commonsense knowledge, is required in order to identify them.
% We assume that if keywords of a question are paraphrased in a context then additional inference, including commonsense knowledge, is required in order to identify them.
% We assume that if words in the context are paraphrased in the question, then additional inferences, including commonsense knowledge, are required to identify them.
% In other words, the questions with high lexical variations are more likely to be challenging compared to the questions with a few lexical variations. Section~\ref{apx:difficulty_word_overlap} analyzes the correlation between the lexical variation of a question and a context, and the difficulty level of the question based on predictions of MNLM-based RC models. Subsequently, we analyze the correlation between whether a question requires commonsense knowledge and the difficulty level of the question based on model predictions.

\revised{The experiments are conducted on the following environment. First of all, we use two widely known MRC benchmarks, the Stanford Question Answering Dataset (SQuAD) 2.0\cite{rajpurkar2018know} and the Reading comprehension with Commonsense Reasoning Dataset (ReCoRD) \cite{zhang2018record}.
SQuAD comprises two types of questions: \textit{has answer} and \textit{no answer}. A \textit{has answer} question contains an answer in the corresponding context. A \textit{no answer} question does not have a contextual answer. On the other hand, ReCoRD is a task that finds the answer in the entities of the context for a given question, and it is known that more commonsense reasoning is needed to solve the problem compared to the SQuAD. ReCoRD is a multiple choice MRC task. ReCoRD contains a cloze-test style question and the correct answer for a question will be the one of entities for the given context\cite{wang2019superglue}. In our experiment, to use the same performance evaluation criteria as SQuAD, we conduct ReCoRD experiments with an answer span prediction approach. We analyze the sampled question-context pairs from the development set of each dataset since we cannot access the test set.} 

\revised{In terms of training models, we only concentrate on the largest models of MNLMs ( BERT$_{large}$, ALBERT1$_{xlarge}$, ALBERT2$_{xlarge}$ and RoBERTa$_{large}$). This is because not only MNLMs show substantially higher performance than GPT 1 and 2 models in natural language understanding tasks\cite{devlin2019bert, van2019does, clark2020electra}, but also there is no way to finetune a GPT 3 model. Detailed analysis for the point of view of lexical overlapping ratio between question and context of each dataset and a data sampling method are described in Appendix~\ref{apx:difficulty_word_overlap}.}

% \subsection{Which Types of Questions Are Still Challenging for MNLMs?}
% \label{sec:question_types}
We first classify the questions into the following six classes by referring to the question types of SQuAD \cite{rajpurkar2016squad}. \revised{Detailed explanations of the question types are described with examples in Appendix~\ref{apx:details_on_RC_question_types}}: 
\begin{itemize}
    \item The \textit{synonymy} class indicates the existence of a synonym relation between an answer sentence and a question.
    \item The \textit{commonsense knowledge} class indicates that commonsense is required to solve a question.
    \item The \textit{no semantic variation} class denotes that a question does neither belong to \textit{synonymy} nor \textit{commonsense knowledge} type.
    \item The \textit{multiple sentence reasoning} class indicates that there are anaphora or clues scattered across multiple sentences.
    \item The \textit{typo} class denotes a typographical error in a question or an answer sentence.
    \item The \textit{others} class indicates that the presented answers are incorrectly tagged.
\end{itemize}

Then, we manually label the question types to see which types are more challenging than other types for the MNLM-based RC models. \revised{We report the analysis results on the proportions of the question types of SQuAD in Table~\ref{tab:question_types_squad} and ReCoRD in Table~\ref{tab:question_types_record}. 
We investigate the proportions of question types comparing questions incorrectly predicted by the models to questions correctly predicted by the models.}
%We compare the proportions of questions that are correctly answered by each model with the incorrectly answered questions.

The results show that there are obvious differences in the proportion of the question types between the correctly predicted questions and the incorrectly predicted questions. Across all models, the proportion of \textit{no semantic variation} questions accounts for more than 50\% in the correctly predicted questions, while the proportion decreases to about 30\% in the incorrectly predicted questions. In other words, the easier questions have a higher proportion of the \textit{no semantic variation} type questions.
%the questions that can be easily solved by the models compared to the questions failed to predict by the models. 
Likewise, the proportion of \textit{commonsense knowledge} questions comprises approximately 22\% to 25\% in the correctly predicted questions, whereas the proportion increases to about 50\% in incorrectly predicted questions. 

\revised{On the other hand, we see that commonsense type questions account for the majority in ReCoRD. Moreover, there is an extremely large portion of semantic variation type questions. In the case of no semantic variation type questions, we notice that all the models answer the same questions incorrectly. In contrast, in case of the commonsense knowledge type questions, BERT, ALBERT1, ALBERT2, and RoBERTa incorrectly predict problems 42, 40, 34, and 29, respectively. This result indicates that the models with the larger size of the pretrained corpora have higher performance on commonsense type questions.}

From these results, we can infer that commonsense type questions can be substantially affected by the size of the pretrained corpus, that is, the size of knowledge contained in the corpus. Therefore, it is still challenging for the MNLM-based RC models to deal with questions with semantic variations, or questions that require commonsense knowledge to solve. This is an important discovery because it suggests that MNLMs should have the capacity to address commonsense knowledge, to overcome the limitations of MNLMs.
%especially commonsense knowledge required questions.

\section{Main Results and a Possible Solution}
\label{sec:discussion}
%\subsection{What are the Fundamental Limitations of Current MNLM Learning?}
Section~\ref{sec:knowledge_probing} reveals that there are still limitations of the existing MNLMs even though they have the potential to infer commonsense knowledge. \revised{Furthermore, the results in Section~\ref{sec:difficulty} indicate that the \textit{commonsense knowledge} type is the most dominant among the questions that MNLM-based RC models have hard time to answer correctly.} We conjecture that the existing MNLMs are heavily trained to learn observed information in the corpus and co-occurrences of the words rather than precise meanings of relations. Here, we first analyze observed joint frequencies of subject and object entities in the BERT pretrain data (English Wikipedia and Book Corpus). Subsequently, we discuss the impact of word frequency on knowledge probing results. 
Finally, we suggest a potential solution and a direction to handle the \textit{semantic variation} type questions by utilizing an external commonsense repository. 
%From the observations in both sections, we can infer that the lack of semantic knowledge of MNLMs accounts for the limitation on the RC task, it is important to improve this.

% \begin{figure*}[ht!] %%%
%     \begin{center}
%     \end{center}
% \caption{\textbf{The architecture of our commonsense knowledge incorporated question answering model.}}
% \label{fig:memory_rc}
% \end{figure*}

\subsection{Why MNLMs still Need External Commonsense Repository?}
\label{sec:frequency}
% 어떤 단어 pair 들이 잘 학습되는지가 중요
% BERT pretrain 데이터 분석. 
% Distant supervision의 가정을 참조하여, 단어가 한 문장에 함께 나오는 경우 성능이 높다고 가정. 
% 1) joint frequency가 높으면 잘나옴
% 2) P(Object|subject)가 높으면 잘 나옴. 특히, rare pattern의 경우 학습이 잘 되는 경향이 있음. 

% 우리는 모델이 commonsense knowledge triple 학습하기 위한 조건들을 분석한다. 이때, relation의 경우 exclusive pattern이 없기 때문에 관측 빈도를 계산하기가 어렵다. 따라서, 우리는 relation extraction  이를 위해, 우리는 기존에 제안된 원거리 지도학습 가정을 unlabeled 데이터의 subject와 object가 함게 등장하는 sentence들은 차용한다. 따라서, 우리는 약 16GB BERT pretrain dataset에서 subject entity들의 관측 빈도와 subject object의 결합 관측 빈도를 계산한다. 최종적으로, 결합 관측 빈도가 1회 이상인 triple들의 knowledge probing 결과를 바탕으로 model's top 100 prediction에서 answer object가 있는지를 분석한다. 
% 
\begin{figure*}[ht!] 
    \begin{center}
    \includegraphics[width=\linewidth/2]{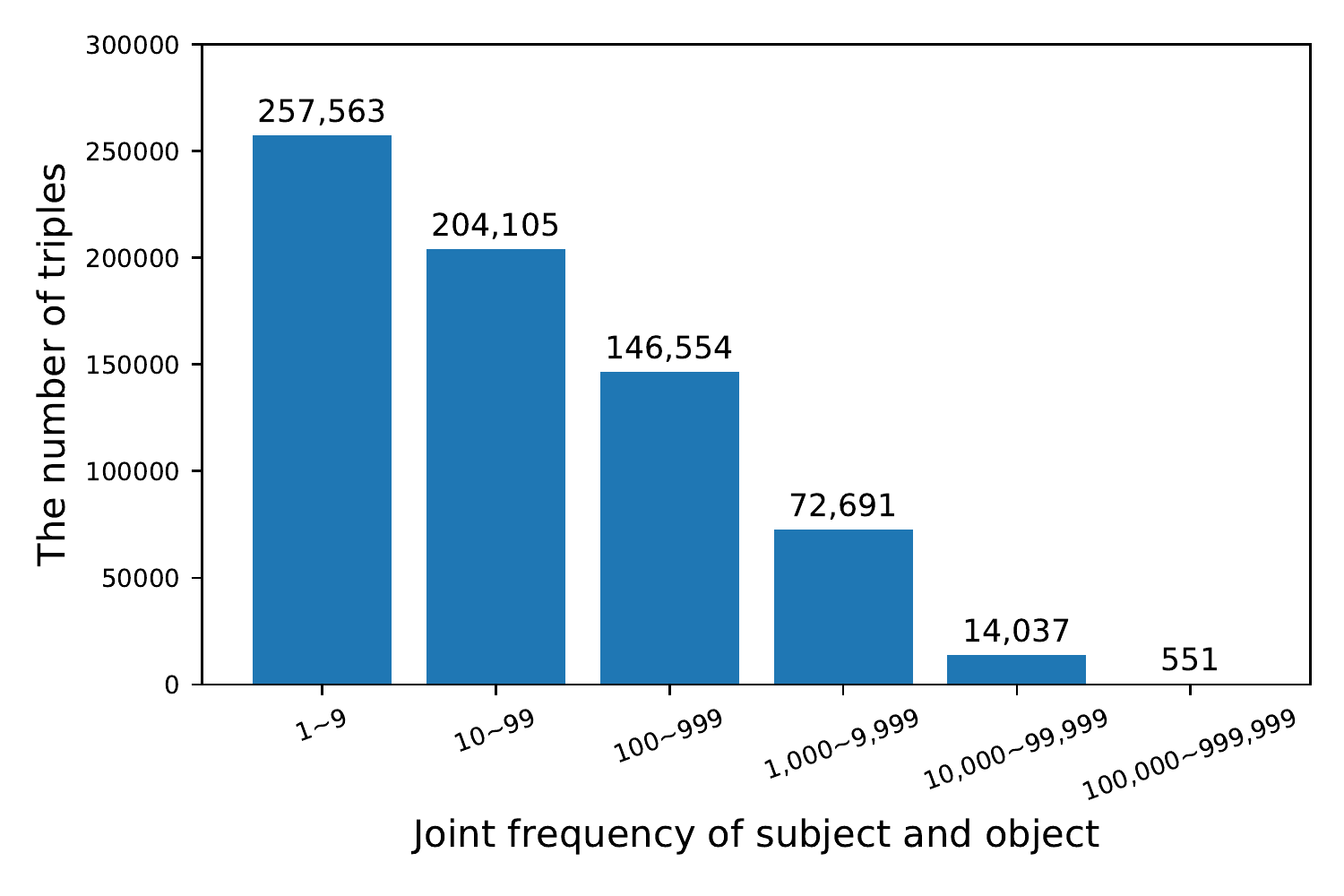}
    \end{center}
%\caption{Color coded results on the BERT$_{large}$ model prediction for 100 samples in the 'MadeOf' relation. Colors are labeled when each sample (x-axis) contain the top 10 most-frequent words. We can notice that top 10 words are redundantly observed in the high rank.}
\caption{\textbf{Statistics of the triples in the BERT pretrain datset.} X-axis indicates the frequency with which subject and object entities are observed together. Y-axis is the number of triples for each section of the joint frequency of subject and object. The joint frequency and the proportion of triples for each frequency sector have a negative correlation.}
\label{fig:entity_pair_statistics}
\end{figure*} %%% Figure 9.

\begin{figure*}[!ht]

\centering
% \begin{subfigure}[b]{0.45\textwidth}
% \includegraphics[width=\linewidth]{images/BERT_base_entity_pair_answer_ratio.pdf}
% \caption{BERT$_{base}$}
% \end{subfigure}
\begin{subfigure}[b]{0.45\textwidth}
\includegraphics[width=\linewidth]{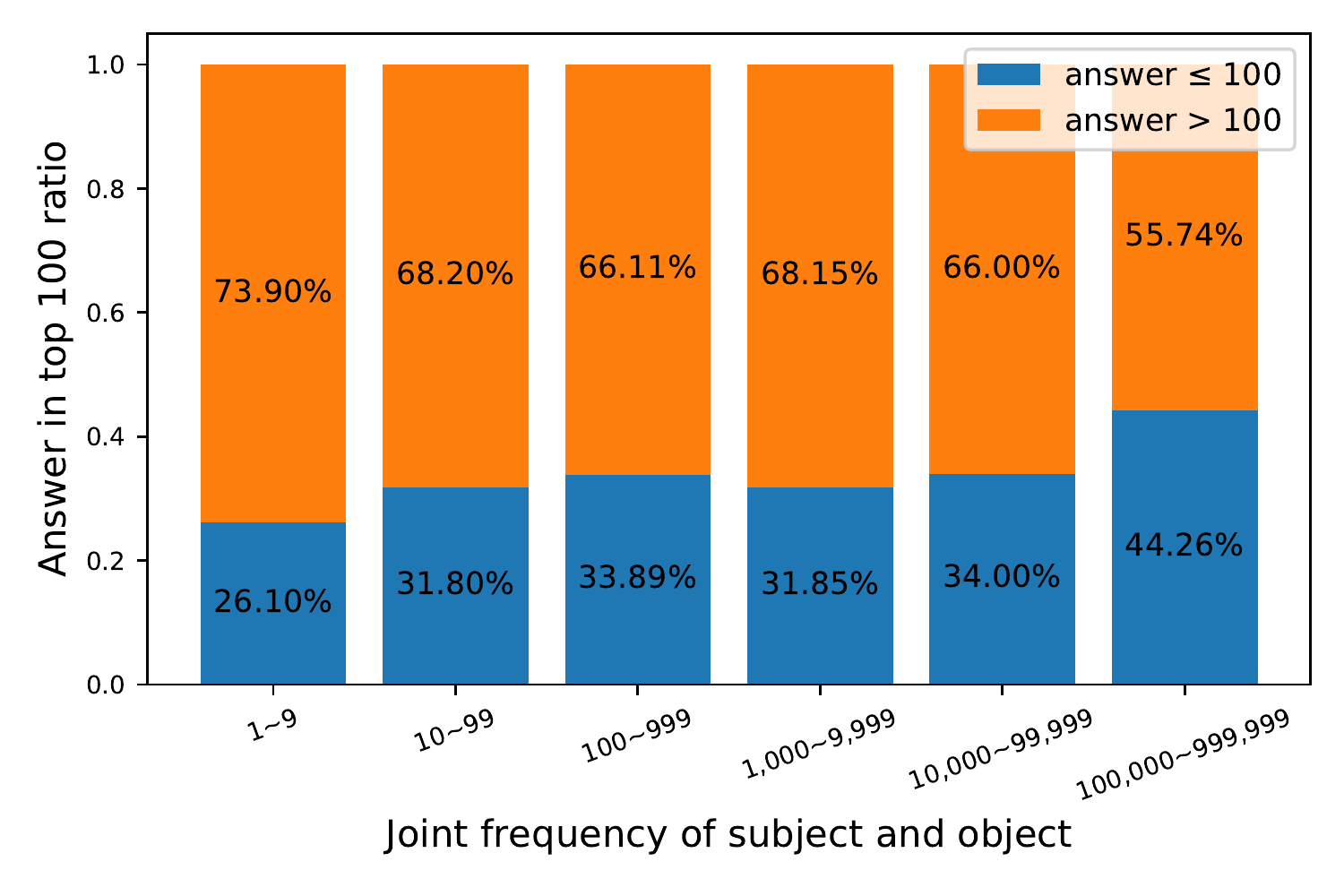}
\caption{BERT$_{large}$}
\end{subfigure}
% \begin{subfigure}[b]{0.45\textwidth}
% \includegraphics[width=\linewidth]{images/high_diff_color.pdf}
% \caption{`High'}
% \end{subfigure}
% \newline
% \begin{subfigure}[b]{0.45\textwidth}
% \includegraphics[width=\linewidth]{images/ALBERT_base_entity_pair_answer_ratio.pdf}
% \caption{ALBERT1$_{base}$}
% \end{subfigure}
% \begin{subfigure}[b]{0.45\textwidth}
% \includegraphics[width=\linewidth]{images/ALBERT_large_entity_pair_answer_ratio.pdf}
% \caption{ALBERT1$_{large}$}
% \end{subfigure}
\begin{subfigure}[b]{0.45\textwidth}
\includegraphics[width=\linewidth]{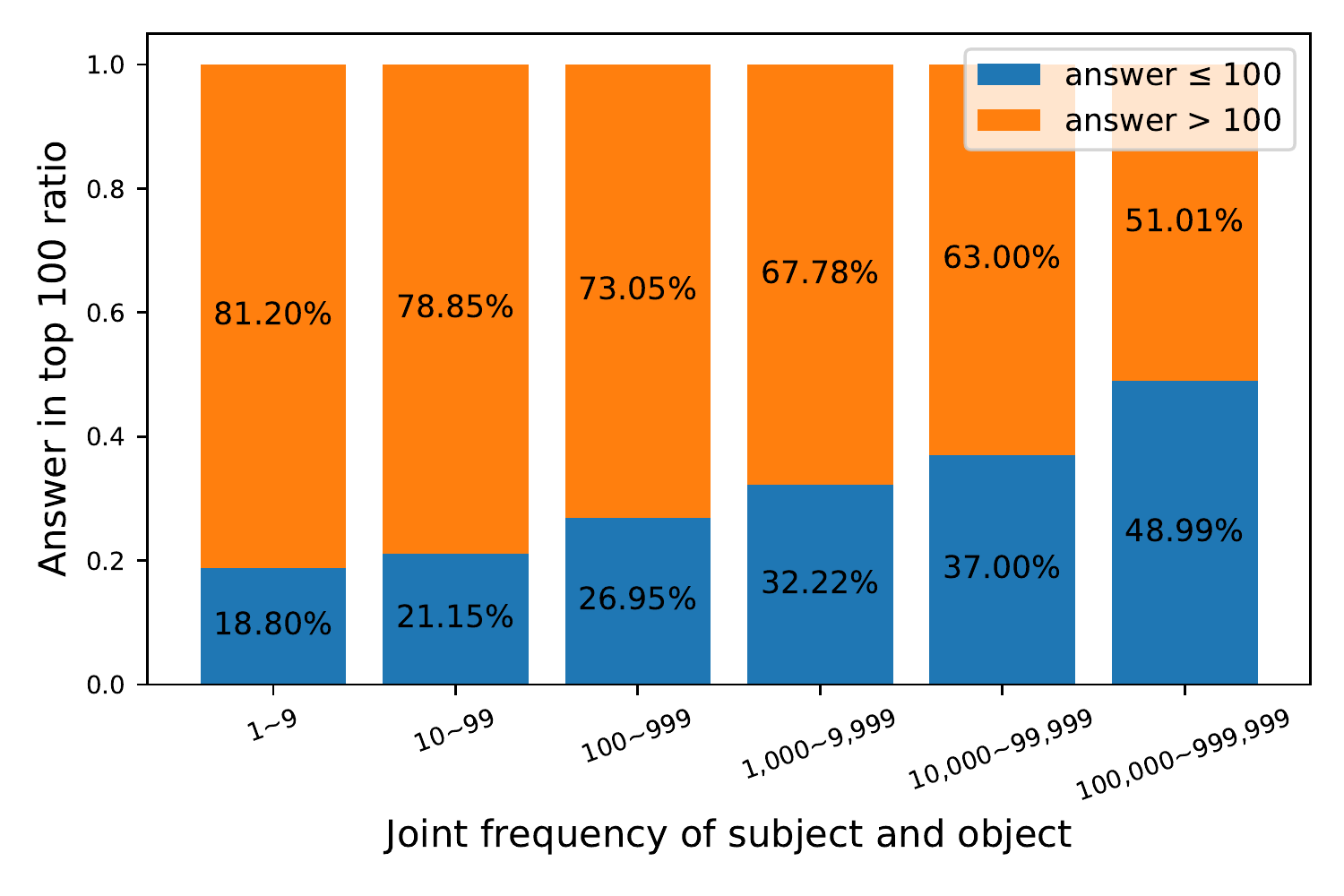}
\caption{ALBERT1$_{xlarge}$}
\end{subfigure}
\caption{\textbf{The proportion of knowledge triples whose objects are in top 100 predictions for each model:} \textbf{(a)} BERT$_{large}$, \textbf{(b)} ALBERT1$_{xlarge}$. X-axis indicates the frequency with which subject and object entities are observed together. Y-axis is the proportion of which the answer object can be found in top 100 model's prediction for each frequency section. Blue bars indicate that the answer is in the top 100 predictions, while orange bars mean that the answer is not in the top 100 predictions. The results show that the joint frequency of subject and object affects the knowledge probing performance.}
\label{fig:bar_plot_of_joint_freq_and_preformance}
\end{figure*}

\begin{figure*}[!ht]
\centering
% \begin{subfigure}[b]{0.45\textwidth}
% \includegraphics[width=\linewidth]{images/BERT_base_conditional_heatmap.pdf}
% \caption{BERT$_{base}$}
% \end{subfigure}
\begin{subfigure}[b]{0.40\textwidth}
\includegraphics[width=\linewidth]{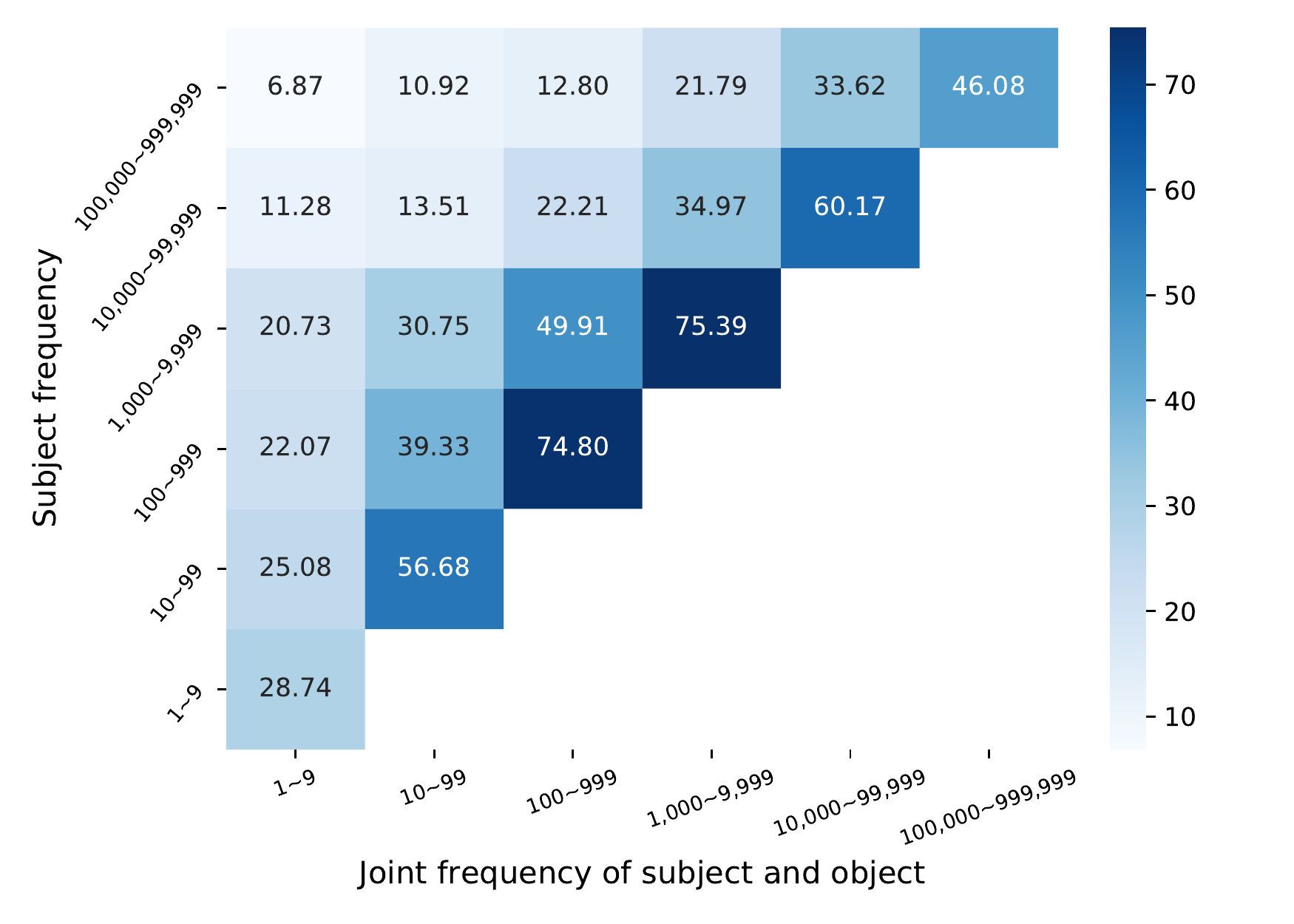}
\caption{BERT$_{large}$ in top 100}
\end{subfigure}
\begin{subfigure}[b]{0.40\textwidth}
\includegraphics[width=\linewidth]{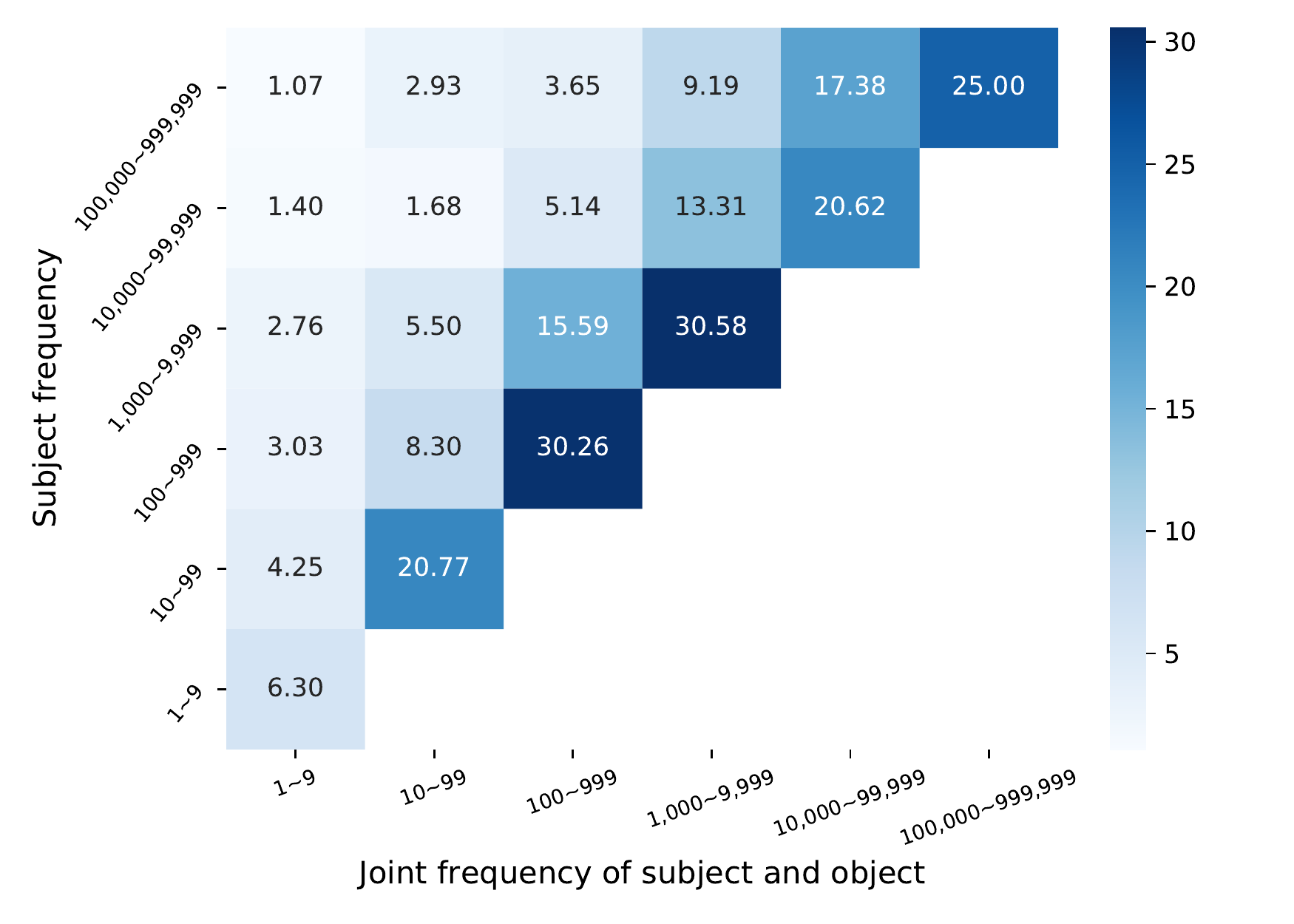}
\caption{BERT$_{large}$ in top k}
\end{subfigure}
\\
% \begin{subfigure}[b]{0.45\textwidth}
% \includegraphics[width=\linewidth]{images/high_diff_color.pdf}
% \caption{`High'}
% \end{subfigure}
% \newline
% \begin{subfigure}[b]{0.45\textwidth}
% \includegraphics[width=\linewidth]{images/ALBERT_base_conditional_heatmap.pdf}
% \caption{ALBERT1$_{base}$}
% \end{subfigure}
% \begin{subfigure}[b]{0.45\textwidth}
% \includegraphics[width=\linewidth]{images/ALBERT_large_conditional_heatmap.pdf}
% \caption{ALBERT1$_{large}$}
% \end{subfigure}
\begin{subfigure}[b]{0.40\textwidth}
\includegraphics[width=\linewidth]{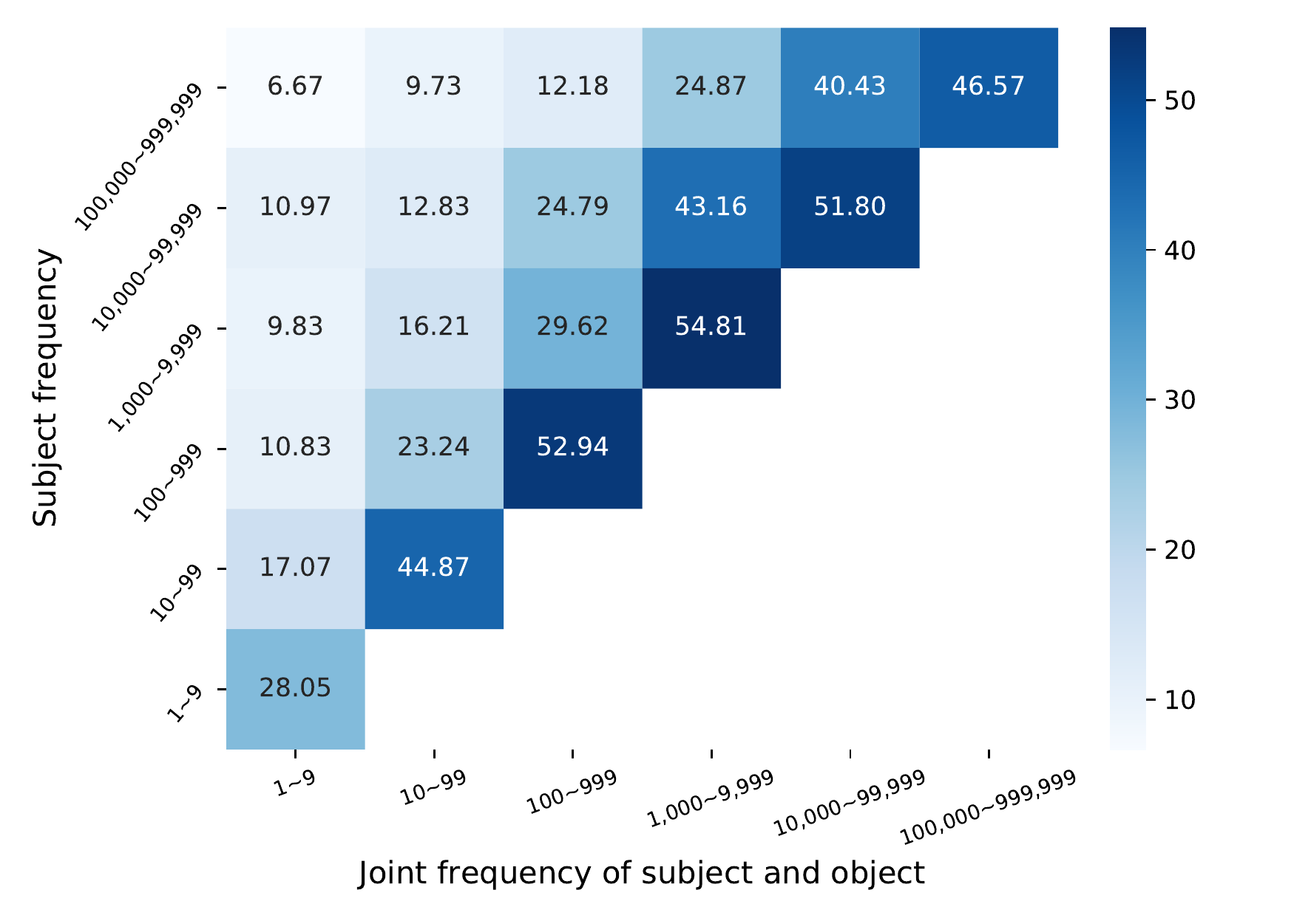}
\caption{ALBERT1$_{xlarge}$ in top 100}
\end{subfigure}
\begin{subfigure}[b]{0.40\textwidth}
\includegraphics[width=\linewidth]{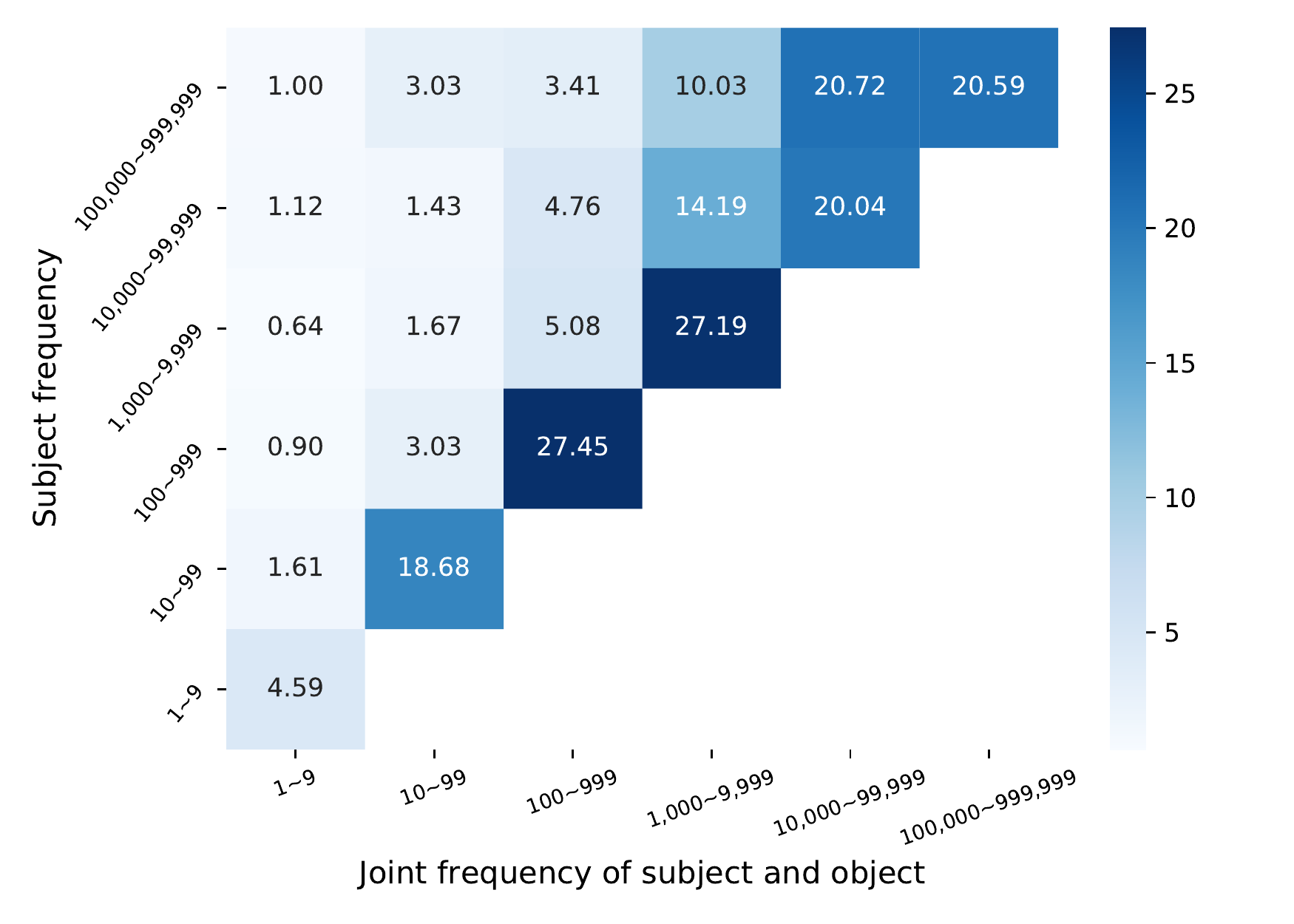}
\caption{ALBERT1$_{xlarge}$ in top k}
\end{subfigure}
\caption{\revised{\textbf{Correlations of the joint frequency of subject and object, and frequency of subject:} \textbf{(a)} BERT$_{large}$ in top 100, \textbf{(b)} BERT$_{large}$ in top k, \textbf{(c)} ALBERT1$_{xlarge}$ in top 100, \textbf{(d)} ALBERT1$_{xlarge}$ in top k. X-axis indicates the frequency with which subject and object entities are observed together. Y-axis is the frequency of the subject. Values of heatmap mean the proportion of knowledge triples whose objects can be found in top 100 or top k model’s predictions for each frequency grid. Herein, the `k' indicates the number of the ground truth objects for each subject and relation pair. The results show that the performance increases as the subject observation frequency is lower in the same joint frequency section of subjects and objects.} }

\label{fig:frquency_performance_heat_map}
\end{figure*}
Herein, we analyze conditions for the MNLMs to learn the commonsense knowledge triples. Since there are no exclusive patterns for each relation, it is difficult to calculate the exact observation frequencies of the triples directly. Therefore, we adopt an assumption of the distant supervision paradigm of relation extraction tasks \cite{mintz2009distant}. The assumption is that if sentences include a pair of entities that have a specific relation, then some of the sentences probably represent the relation of the entity pair. The more sentences contain the entity pair, the more sentences are likely to represent the relation. \revised{For this reason, we calculate the joint frequency of entity pairs of the ConceptNet from the BERT pretrain dataset. In this section, top 100 indicates that the answer object can be found in the top 100 prediction of a model. Furthermore, top k denotes that the answer object can be found in the number of objects of a given subject and relation pair.}
%The assumption is that if two entities have a specific relation then it is likely to represent the relation.  
%Thus, if there are sentences containing a given entity pair, some of the sentences may include a relation of the entity pair what we want to know.
%Therefore, we calculate the observed frequency for each entity of ConceptNet and 

Figure~\ref{fig:entity_pair_statistics} presents the statistics for the number of triples in each interval of the joint frequency of subject and object. Herein, many triples lack of the joint frequency.
Then, we analyze the probing results of BERT and ALBERT1 models which are trained on the BERT pretrain dataset. 
Finally, we attest whether there is an answer object in the model's top 100 or top k predictions based on the knowledge probing results of triples whose joint frequency of the subject and object entity pair is at least 1 time. In this section, we only show the results of the BERT$_{large}$ and ALBERT1$_{xlarge}$ models since the experimental results are consistent in the other models. Experimental results of the other models are demonstrated in Appendix~\ref{apx:frequency_performance}. 

\revised{Figure~\ref{fig:bar_plot_of_joint_freq_and_preformance}, the main observation of our study, shows that the joint frequency of subjects and objects obviously affects performances of the knowledge prediction. Nevertheless, there are some entity pairs with very low frequency (1 to 9) which succeed to predict the answer in the top 100. We analyze why this happens and report the results in Figure~\ref{fig:frquency_performance_heat_map}. The results show that, in the same joint frequency section, as the subject observations frequency increases, performance tends to decrease. Therefore, we can see that the performance is positively affected by the joint frequency of the subject and object, while negatively affected by the subject frequency. It demonstrates that commonsense knowledge trained in MNLMs can be influenced by the conditional probability of the object given the subject.}

From these findings, training MNLMs on a larger dataset will be useful to enhance learning commonsense knowledge as the joint frequency of entity pairs is expected to increase in the larger dataset. However, it can be inefficient to simply increase the size of the dataset to sufficiently train relatively rare entity pairs.
This is because, most entity pairs in the larger dataset can be redundant since the number of co-occurrences of word pairs follows the power-law distribution\cite{pennington2014glove}, similar to the well-known Zipf's law\cite{powers1998applications}.
Moreover, it does not guarantee that the conditional probability of entity pairs in the larger dataset will increase.
%although the frequency of rarely observed entity pairs may increase in the larger dataset.  
 %In other words, appropriately observed entity pairs 
%the ratio between fre  the number of frequently observed entity pairs is relatively small, and overwhelmingly less than rarely observed entity pairs. 
%On the other hand, even if the size of the dataset increases, rarely observed entity pairs will be still rare compared to the frequent pairs, if the distribution of data does not change much.
%in a sense that they may not be used to learn not-so-correctly trained relations 
%according to the Zipf’s law \cite{powers1998applications,pennington2014glove}. 
%In the case of entity pairs with high subject frequency and low joint frequency, it is difficult to anticipate significant changing of conditional probability of the pairs by simply multiplying the dataset size if the words distribution is almost constant. 
Therefore, learning commonsense knowledge apart from the distribution of training corpora will be an important factor for overcoming the limitations of the existing MNLMs.
%Therefore, defining the facts we want to know and using them separately from the distribution of the training corpora will be an important factor for overcoming the limitations of existing MNLMs. 
Taking all these into consideration, it is still hard for the existing MNLMs to learn a significant portion of commonsense knowledge and may need help from external knowledge repositories or other tailored special purpose datasets.

% The results of finetuning 

\subsection{Can An External Commonsense Repository Help for MNLMs to Solve the Questions including the Semantic Variation?}
\label{sec:complement_NLMs_with_knowledge}
As discussed, it is clear that there are limitations in the existing MNLM-based RC models to infer semantic variation type questions.
However, it is yet obscure whether the explicit complement of necessary knowledge can be a solution to the limitations of the MNLM-based RC models. 
Therefore, we conduct a controlled experiment to verify whether the external commonsense knowledge can help ameliorate the limitations. Specifically, we integrate adequate knowledge by enriching the text of a question or a context without additional training or changing the model. \revised{Examples in Table~\ref{tab:example_of_manually_integrating_in_squad} and Table~\ref{tab:example_of_manually_integrating_in_record} illustrate how the required knowledge is integrated into the text of each dataset. First, we find the subject word of each required knowledge triple from the question and the context. Then, the word is enriched with the relation and the object of the triple. Specifically, the relation is converted into natural language using the templates. A detailed algorithm of the knowledge integration is provided in Appendix~\ref{apx:details_on_the_integrating_external_commonsense_repository_test}.}

%on the hardest problems in sector A. Details on the experimental settings and examples are provided in Appendix~\ref{apx:details_on_the_integrating_external_commonsense_repository_test}.

\revised{We analyze the \textit{semantic variation} type questions to figure out whether clues for each question can be found among knowledge triples in the ConceptNet. As a result we find that 244 questions out of 684 semantic variation type questions of SQuAD have clues in the ConceptNet. We also observe 104 questions out of 200 semantic variation type questions of ReCoRD have clues in the ConceptNet. For the evaluation, the questions are divided into two groups according to their exact match (EM) \cite{rajpurkar2016squad} scores. The first group is `incorrect' questions whose predicted answers are not fully matched with the ground truth answers. The other group is `correct' questions whose predicted answers are fully matched with the ground truth answers.}
Then, we evaluate our baseline models with the selected data in EM and F1 score of Equation~(\ref{F1}), that is a harmonic mean of recall (Equation~(\ref{RECALL})) and precision (Equation~(\ref{PRECISION})). We use BERT $_{large}$, ALBERT1 $_{xlarge}$, ALBERT2 $_{xlarge}$, and RoBERTa $_{large}$, as our baseline models. In addition, we conduct one-tailed t-test for the significant test.
\begin{equation} \label{PRECISION} 
\hspace*{2.1cm} 
\text{Precision}=\frac{\text{\#\,of\,words\,in\,predicted\,answer\,matched\,with\,words\,in\,the\,ground\,truth}}{\text{\#\,of\,words\,in\,predicted\,answer}}
\end{equation}
\begin{equation} \label{RECALL}
\hspace*{2.35cm} 
\text{Recall}=\frac{\text{\#\,of\,words\,in\,predicted\,answer\,matched\,with\,words\,in\,the\,ground\,truth}}{\text{\#\,of\,words\,in\,the\,ground\,truth}}
\end{equation}
\begin{equation} \label{F1}
\hspace*{5.8cm} 
\text{F1}=\frac{2 \times \text{Precision} \times \text{Recall}}{\text{Precision}+\text{Recall}}
\end{equation}

\revised{Table~\ref{tab:result_of_manual_integration} indicates the results of the knowledge integration. The experimental results show that the performance of incorrect questions are significantly improved from all experimental models ($p$ < 0.05) by adapting knowledge integrated text without any modifications in the model architecture on both SQuAD and ReCoRD. On the other hand, as a result of the knowledge integration, the performance of correct questions was slightly decreased. Nevertheless, from the experimental results, we can see that the improvement in the performance of incorrect questions is prominently higher compared to the performance decreasing of the correct questions. }

\revised{What should be noted are the results of the ALBERT1 and ALBERT2 models. In particular, the performance improvement in knowledge integration on ALBERT1 is consistently higher than that of ALBERT2. In the previous sections, we showed that ALBERT1 was trained with the small amount of text corpora on the same topology compared to ALBERT2, and this can result in 1) poor probing performance, and 2) relatively susceptible to solving the semantic variation type questions. Considering all these points, we can infer that ALBERT1, which lacks knowledge of learning, is more sensitive to the knowledge contained in the context.}

\section{Conclusion and Future Work}
\label{sec:conclusion}
We investigate which types of commonsense knowledge are trained in the pretrained MNLMs by using the knowledge probing test. We find that MNLMs understand some commonsense knowledge while the trained knowledge is not precise enough to distinguish opposite relations. We also find that the questions requiring commonsense knowledge are still challenging to existing MNLM-based RC systems. 
Finally, we empirically demonstrate that the limitations of MNLMs can be complemented by integrating a commonsense knowledge repository. To the best of our knowledge, our study is the first to report the fundamental reason why existing MNLMs do not include a large portion of commonsense knowledge yet. 

%\revised{ First of all, ~. In addition, in ~. }
%We believe that further analysis of this will be necessary in the future.}

\revised{Although the aforementioned observations, some questions are still left behind us. First of all, in Section~\ref{sec:frequency}, we analyze the impact of commonsense explicitly trained in the corpus, although commonsense is found implicitly in the corpus, in many cases. However, due to the nature of implicit commonsense, it is extremely difficult to directly measure the influence on the corpus, and many previous studies have tried to evaluate it based on indirect approaches such as probing. 
To ameliorate this, we intensively analyze pairs with high hit scores among very low frequency subject-object pairs. In particular, we can analyze the correlation with relatively frequently observed entity pairs by using the adjacency graph structure for co-occurrence frequencies between entities. In addition, it is required to verify the direct causality between the ability of knowledge replication and the performance of the downstream tasks.} It is also a question to investigate how MNLMs learned to discern a subtle semantic difference between opposite relations from unlabeled text. Further, an automatic solution for the knowledge integrating remains to be developed as future work.
% In the future, we will study how MNLMs learned to discern a subtle semantic difference between opposite relations from unlabeled text. Moreover, we are going to make an automatic solution for the knowledge integrating.

\bibliography{main.bib}

\noindent\textbf{Acknowledgements}\\
This work is supported by IITP grant funded by the Korea government (MIST) (2017-0-00255, Autonomous digital companion framework and application), IITP grant funded by the Korea government (MIST) (2017-0-01779, XAI), Artificial Intelligence Graduate School Program (KAIST) (2019-0-00075), and NAVER Corp.\\

\noindent\textbf{Author contributions}\\
S. Kwon, C. Kang, J. Han and J. Choi designed research. S. Kwon, C. Kang and J. Han performed research. S. Kwon, C. Kang and J. Choi analyzed the data. S. Kwon, J. Han and J. Choi revised the manuscript. S. Kwon, C. Kang, J. Han and J. Choi wrote the paper. S. Kwon, C. Kang, J. Han and J. Choi performed revision. 
\\

\noindent\textbf{Data and code availability}\\
All of data used in our manuscript are available in referred web link or in the supplementary information. The related codes are available in \url{https://zenodo.org/record/3982680#.XzUvU6dxeUl}
\\
% Y.Y. conceptualized the idea. Y.Y., H.-T.Z. and L.Y. initialized, conceived and supervised
% the project. L.Y., H.X. and S.L. collected data. Y.Y., M.W., Y.G. and C.S. discovered key
% features and the clinical route. L.Y., H.-T.Z., Yang Xiao, L.M., H.X., J.G. and Y.Y. drafted
% the manuscript. All authors provided critical review of the manuscript and approved the
% final draft for publication.

\noindent\textbf{Competing interests}\\
The authors declare no competing interests.

% \noindent\textbf{Supplementary information (optional)}
% If your article requires supplementary information, please include these files for peer-review. Please note that supplementary information will not be edited.

% \begin{table}[]
% \centering
% % \small
% \caption{Results of micro average and macro average hits@K for the ConceptNet relations. }
% %The macro avg. equally average the results of all relations, while the micro avg. weighted average the results of the relations according to their portion.}
% \label{tab:hits@K}
% \begin{tabular}{c|l|ccc}
% \hline
% \multicolumn{2}{c|}{\multirow{2}{*}{Model}} & \multicolumn{3}{c}{Hits@K} \\ \cline{3-5} 
% \multicolumn{2}{c|}{} & 1 & 10 & 100 \\ \hline
% \multirow{5}{*}{\begin{tabular}[c]{c}Micro\\average\end{tabular}} & BERT$_{base}$ & 5.93  & 17.36 & 34.33\\
%  & BERT$_{large}$ & 5.08 & 16.78 & 33.36  \\\cline{2-5}
%  & ALBERT$_{base}$ & 5.15& 15.05& 31.56  \\
%  & ALBERT$_{large}$ & \textbf{8.26}& 19.87& 35.80\\
%  & ALBERT$_{xlarge}$ & 8.05& \textbf{20.34}& \textbf{35.94}  \\\hline\hline
% \multirow{5}{*}{\begin{tabular}[c]{c}Macro\\average\end{tabular}} & BERT$_{base}$ & 5.06 & 18.84 & 39.49\\
%  & BERT$_{large}$ & 6.68  & 20.04 & 41.94 \\ \cline{2-5}
%  & ALBERT$_{base}$ & 4.64& 16.54& 38.89\\ 
%  & ALBERT$_{large}$ & 6.18& 18.93& 39.88\\
%  & ALBERT$_{xlarge}$ & \textbf{7.57}& \textbf{22.19}& \textbf{42.87} \\\hline
% \end{tabular}
% \end{table}
\newpage

\begin{table}[h!]
\centering
\caption{\textbf{Detail structures of MNLMs and NLMs}}
\label{tab:model_structures}
\begin{tabular}{l|cccc}
\hline
\multicolumn{1}{c|}{Model} & Parameters & Layers & Hidden & Data\\
\hline\hline
BERT$_{base}$ & 108M & 12 & 768 & 16GB\\
BERT$_{large}$ & 334M & 24 & 1,024 & 16GB\\
\hline
ALBERT1$_{base}$ & 12M & 12 & 768 & 16GB\\
ALBERT1$_{large}$ & 18M & 24 & 1,024 & 16GB\\
ALBERT1$_{xlarge}$ & 60M & 24 & 2,048 & 16GB\\
\hline
ALBERT2$_{base}$ & 12M & 12 & 768 & 160GB\\
ALBERT2$_{large}$ & 18M & 24 & 1,024 & 160GB\\
ALBERT2$_{xlarge}$ & 60M & 24 & 2,048 & 160GB\\
\hline
RoBERTa$_{base}$ & 108M & 12 & 768 & 160GB\\
RoBERTa$_{large}$ & 334M & 24 & 1,024 & 160GB\\
\hline
GPT1 & 119M & 12 & 768 & 4GB\\
GPT2 & 1,558M & 25 & 1,600 & 40GB\\
GPT3 & 175B & 96 & 12,288 & 570GB\\
\hline
\end{tabular}
\end{table}

\begin{table}[h!]
\centering
\begin{threeparttable}
% \small
\caption{\textbf{Result of micro average and macro average performance of 1) the average hits@K performance of candidate masked sentences for each triple (`AVG') and 2) the hits@K of the selected masked sentence (`SEL').}}
%The macro avg. equally average the results of all relations, while the micro avg. weighted average the results of the relations according to their portion.}
\label{tab:impact_of_template_selection}
\begin{tabular}{c|c|ccc|ccc}
\hline
\multicolumn{1}{c|}{\multirow{3}{*}{Model}} & \multicolumn{1}{c|}{\multirow{3}{*}{\begin{tabular}[c]{@{}c@{}}Sentence\\Selection\end{tabular}}}  & \multicolumn{6}{c}{Hits@K}  \\ \cline{3-8}
\multicolumn{1}{c|}{} & &\multicolumn{3}{c|}{Micro average} & \multicolumn{3}{c}{Macro average} \\\cline{3-8}
\multicolumn{1}{c|}{} & &  1 & 10 & 100 & 1 & 10 & 100 \\ \hline\hline

% BERT$_{base}$ & \multicolumn{1}{c|}{\multirow{2}{*}{AVG}} & 5.26$\pm$2.55 & 13.78$\pm$5.73 & 28.57$\pm$12.83 & 4.49$\pm$2.76 & 13.88$\pm$6.84 & 30.94$\pm$9.35\\
% BERT$_{large}$ &  & 5.00$\pm$3.65 & 13.94$\pm$8.19 & 28.60$\pm$13.33& 5.27$\pm$3.95 & 15.15$\pm$9.77 & 32.64$\pm$11.17\\
% \hline
% BERT$_{base}$ & \multicolumn{1}{c|}{\multirow{2}{*}{SEL}} & \textbf{7.39$\pm$0.03} & \textbf{19.07$\pm$0.05} & \textbf{36.24$\pm$0.06} & \textbf{6.33$\pm$1.11} & \textbf{18.19$\pm$2.27} & \textbf{37.39$\pm$3.16}\\
% BERT$_{large}$ && \textbf{6.89$\pm$0.03} & \textbf{19.00$\pm$0.05} & \textbf{35.98$\pm$0.06} & \textbf{7.35$\pm$1.21} & \textbf{20.04$\pm$2.30} & \textbf{39.63$\pm$3.23}\\
BERT$_{base}$ & \multicolumn{1}{c|}{\multirow{2}{*}{AVG}} & 5.26 & 13.78 & 28.57 & 4.49 & 13.88 & 30.94\\
BERT$_{large}$ &  & 5.00 & 13.94 & 28.60& 5.27& 15.15 & 32.64\\
\hline
BERT$_{base}$ & \multicolumn{1}{c|}{\multirow{2}{*}{SEL}} & \textbf{7.39} & \textbf{19.07} & \textbf{36.24} & \textbf{6.33} & \textbf{18.19} & \textbf{37.39}\\
BERT$_{large}$ && \textbf{6.89} & \textbf{19.00} & \textbf{35.98} & \textbf{7.35} & \textbf{20.04} & \textbf{39.63}\\
\hline
\end{tabular}
% {\footnotesize
% % \begin{tablenotes}
% % % \item[] Data presented as indicate $\pm$ mean of standard error of mean
% % %\item[\textdagger] Questions succeed or failed to predict by all experimental models commonly
% % % \item[\textdaggerdbl] Symbol 1
% % % \item[\S] Symbol 1  
% % % \item[$\|$] Symbol 1
% % % \item[$\!$\#] Symbol 1          
% % \end{tablenotes}
% }
% {\small
% \begin{tablenotes}
% \item[a] testing
% \end{tablenotes}
% }
\end{threeparttable}
\end{table}

\begin{table}[h!]
\centering
\begin{threeparttable}
% \small
\caption{\textbf{Results of micro average and macro average hits@K for the ConceptNet relations. }}
%The macro avg. equally average the results of all relations, while the micro avg. weighted average the results of the relations according to their portion.}
\label{tab:hits@K}
\begin{tabular}{c|ccc|ccc}
\hline
\multicolumn{1}{c|}{\multirow{3}{*}{Model}} & \multicolumn{6}{c}{Hits@K}  \\ \cline{2-7}
\multicolumn{1}{c|}{}&\multicolumn{3}{c|}{Micro average} & \multicolumn{3}{c}{Macro average} \\\cline{2-7}
\multicolumn{1}{c|}{} &  1 & 10 & 100 & 1 & 10 & 100 \\ \hline\hline
% BERT$_{base}$ & 3.57$\pm$0.02&12.29$\pm$0.04&28.57$\pm$0.06&4.32$\pm$0.67&17.30$\pm$2.40&38.47$\pm$3.47\\
% BERT$_{large}$ & 3.44$\pm$0.02&12.24$\pm$0.04&27.86$\pm$0.06&6.27$\pm$1.13&18.81$\pm$2.39&40.16$\pm$3.12\\
% \hline
% ALBERT1$_{base}$&5.82$\pm$0.03&15.27$\pm$0.05&31.15$\pm$0.06&2.73$\pm$0.43&10.48$\pm$1.19&29.26$\pm$2.14\\
% ALBERT1$_{large}$ &2.66$\pm$0.02&10.82$\pm$0.04&26.44$\pm$0.06&2.13$\pm$0.33&9.92$\pm$0.95&29.93$\pm$1.84\\
% ALBERT1$_{xlarge}$ &2.01$\pm$0.02&7.17$\pm$0.03&19.93$\pm$0.05&3.08$\pm$0.66&9.77$\pm$0.99&25.68$\pm$1.78 \\
% \hline
% ALBERT2$_{base}$ &4.04$\pm$0.03&8.35$\pm$0.03&25.56$\pm$0.05&4.49$\pm$0.68&17.23$\pm$1.72&39.06$\pm$2.22\\
% ALBERT2$_{large}$ &6.28$\pm$0.03&17.00$\pm$0.05&33.88$\pm$0.06&5.62$\pm$0.76&18.18$\pm$1.59&40.78$\pm$2.39\\
% ALBERT2$_{xlarge}$ &6.43$\pm$0.03&17.37$\pm$0.05&33.49$\pm$0.06&6.78$\pm$0.88&20.71$\pm$1.70&42.21$\pm$2.40\\
% \hline
% RoBERTa$_{base}$ & 1.08$\pm$0.01&5.65$\pm$0.03&19.67$\pm$0.05&4.21$\pm$0.78&13.63$\pm$1.63&33.58$\pm$2.50\\
% RoBERTa$_{large}$ & 1.20$\pm$0.01&8.35$\pm$0.03&25.56$\pm$0.05&5.12$\pm$0.84&17.23$\pm$1.72&39.06$\pm$2.22\\
% \hline
% GPT1 & 1.20$\pm$0.01&8.35$\pm$0.03&25.56$\pm$0.05&5.12$\pm$0.84&17.23$\pm$1.72&39.06$\pm$2.22\\
% GPT2 & 1.20$\pm$0.01&8.35$\pm$0.03&25.56$\pm$0.05&5.12$\pm$0.84&17.23$\pm$1.72&39.06$\pm$2.22\\
% GPT3 & 1.20$\pm$0.01&8.35$\pm$0.03&25.56$\pm$0.05&5.12$\pm$0.84&17.23$\pm$1.72&39.06$\pm$2.22\\
BERT$_{base}$ & 7.39$\pm$0.03 & 19.07$\pm$0.05 & 36.24$\pm$0.06 & 6.33$\pm$1.11 & 18.19$\pm$2.27 & 37.39$\pm$3.16\\
BERT$_{large}$ & 6.89$\pm$0.03 & 19.00$\pm$0.05 & 35.98$\pm$0.06 & \textbf{7.35$\pm$1.21} & 20.04$\pm$2.30 & 39.63$\pm$3.23\\
\hline
ALBERT1$_{base}$ & 8.36$\pm$0.04 & 19.21$\pm$0.05 & 36.33$\pm$0.06 & 3.58$\pm$0.64 & 13.51$\pm$1.33 & 30.82$\pm$2.05\\
ALBERT1$_{large}$ & 5.46$\pm$0.03 & 15.55$\pm$0.05 & 32.04$\pm$0.06 & 3.62$\pm$0.60 & 11.91$\pm$1.22 & 30.45$\pm$1.65\\
ALBERT1$_{xlarge}$ & 4.44$\pm$0.03 & 11.29$\pm$0.04 & 24.82$\pm$0.06 & 3.58$\pm$0.55 & 11.98$\pm$1.33 & 26.25$\pm$1.85\\
\hline
ALBERT2$_{base}$ & 6.01$\pm$0.03 & 17.06$\pm$0.05 & 34.70$\pm$0.06 & 4.82$\pm$0.58 & 16.18$\pm$1.35 & 36.83$\pm$2.04\\
ALBERT2$_{large}$ & 8.52$\pm$0.04 & 21.36$\pm$0.05 & 39.38$\pm$0.06 & 6.10$\pm$0.68 & 19.13$\pm$1.42 & 39.40$\pm$2.01\\
ALBERT2$_{xlarge}$ & \textbf{8.83$\pm$0.04 }& 21.93$\pm$0.05 & 39.23$\pm$0.06 & 7.29$\pm$0.90 & \textbf{21.32$\pm$1.67} & 41.73$\pm$1.92\\
\hline
RoBERTa$_{base}$ & 2.95$\pm$0.02 & 10.26$\pm$0.04 & 26.71$\pm$0.06 & 4.53$\pm$0.72 & 14.13$\pm$1.53 & 32.94$\pm$2.25\\
RoBERTa$_{large}$ & 3.89$\pm$0.03 & 14.83$\pm$0.05 & 34.27$\pm$0.06 & 5.94$\pm$0.83 & 18.21$\pm$1.67 & 39.67$\pm$2.21\\
\hline
GPT1 & 0.30$\pm$0.01 & 3.75$\pm$0.02 & 15.82$\pm$0.05 & 0.98$\pm$0.30 & 6.85$\pm$1.07 & 24.79$\pm$2.12\\
GPT2 & 3.73$\pm$0.03 & 21.55$\pm$0.05 & 43.32$\pm$0.06 & 3.01$\pm$0.63 & 16.43$\pm$1.72 & 37.50$\pm$2.43\\
GPT3 & 3.58$\pm$0.02 & \textbf{24.07$\pm$0.06} & \textbf{52.13$\pm$0.06} & 3.69$\pm$0.78 & 18.89$\pm$1.97 & \textbf{44.81$\pm$2.52}\\
\hline
\end{tabular}
{\footnotesize
\begin{tablenotes}
\item[] Data presented as mean $\pm$ standard error of mean
%\item[\textdagger] Questions succeed or failed to predict by all experimental models commonly
% \item[\textdaggerdbl] Symbol 1
% \item[\S] Symbol 1  
% \item[$\|$] Symbol 1
% \item[$\!$\#] Symbol 1          
\end{tablenotes}
}
% {\small
% \begin{tablenotes}
% \item[a] testing
% \end{tablenotes}
% }
\end{threeparttable}
\end{table}

\begin{table*}[h!]
\footnotesize
\centering
\begin{threeparttable}
\caption{\textbf{Results of overlapping ratio at top K predictions between the opposite relations. }}
\label{tab:overlapping_results}
\begin{tabular}{c|c|ccc}
\hline
\multirow{2}{*}{Relation} &\multirow{2}{*}{Model}& \multicolumn{3}{c}{Overlap@K\tnote{*}} \\
\cline{3-5}
    &   &   1   &   10  &   100\\\hline \hline
\multirow{13}{*}{Synonym} & BERT$_{ base }$ & 45.21$\pm$0.99 & 44.19$\pm$0.66 & 47.77$\pm$0.61 \\
& BERT$_{ large }$ & 37.00$\pm$0.96 & 41.12$\pm$0.62 & 46.09$\pm$0.56 \\
\cline{2-5}
& ALBERT1$_{ base }$ & 55.48$\pm$1.02 & 42.51$\pm$0.68 & 49.39$\pm$0.60 \\
& ALBERT1$_{ large }$ & 46.26$\pm$1.03 & 38.82$\pm$0.69 & 46.74$\pm$0.60 \\
& ALBERT1$_{ xlarge }$ & 35.03$\pm$0.98 & 34.22$\pm$0.66 & 40.97$\pm$0.61 \\
\cline{2-5}
& ALBERT2$_{ base }$ & 48.77$\pm$1.03 & 43.01$\pm$0.64 & 49.38$\pm$0.57 \\
& ALBERT2$_{ large }$ & 62.24$\pm$1.00 & 55.99$\pm$0.48 & 61.08$\pm$0.37 \\
& ALBERT2$_{ xlarge }$ & 52.13$\pm$1.03 & 52.90$\pm$0.55 & 57.66$\pm$0.45 \\
\cline{2-5}
& RoBERTa$_{ base }$ & 21.76$\pm$0.82 & 34.36$\pm$0.41 & 39.63$\pm$0.32 \\
& RoBERTa$_{ large }$ & 19.29$\pm$0.78 & 31.47$\pm$0.35 & 36.61$\pm$0.25 \\
\cline{2-5}
& GPT1 & 54.56$\pm$0.99 & 56.50$\pm$0.78 & 57.99$\pm$0.74 \\
& GPT2 & 56.32$\pm$0.98 & 59.25$\pm$0.74 & 62.27$\pm$0.67 \\
& GPT3 & 47.45$\pm$0.99 & 32.66$\pm$0.25 & 41.62$\pm$0.19 \\
\cline{2-5}
\hline
\multirow{13}{*}{Desires} & BERT$_{ base }$ & 57.89$\pm$11.64 & 50.00$\pm$3.90 & 65.79$\pm$1.45 \\
& BERT$_{ large }$ & 21.05$\pm$9.61 & 46.32$\pm$4.60 & 65.74$\pm$2.18 \\
\cline{2-5}
& ALBERT1$_{ base }$ & 78.95$\pm$9.61 & 72.63$\pm$3.96 & 72.79$\pm$1.14 \\
& ALBERT1$_{ large }$ & 26.32$\pm$10.38 & 62.63$\pm$4.25 & 73.32$\pm$2.25 \\
& ALBERT1$_{ xlarge }$ & 15.79$\pm$8.59 & 55.26$\pm$3.85 & 67.00$\pm$2.16 \\
\cline{2-5}
& ALBERT2$_{ base }$ & 31.58$\pm$10.96 & 58.95$\pm$3.66 & 69.37$\pm$1.58 \\
& ALBERT2$_{ large }$ & 42.11$\pm$11.64 & 60.53$\pm$3.63 & 69.63$\pm$1.92 \\
& ALBERT2$_{ xlarge }$ & 57.89$\pm$11.64 & 58.42$\pm$4.73 & 71.42$\pm$2.30 \\
\cline{2-5}
& RoBERTa$_{ base }$ & 52.63$\pm$11.77 & 48.95$\pm$3.74 & 61.79$\pm$2.16 \\
& RoBERTa$_{ large }$ & 5.26$\pm$5.26 & 41.58$\pm$4.41 & 58.26$\pm$1.94 \\
\cline{2-5}
& GPT1 & 52.63$\pm$11.77 & 58.42$\pm$3.18 & 62.84$\pm$1.37 \\
& GPT2 & 57.89$\pm$11.64 & 58.95$\pm$3.23 & 64.89$\pm$1.64 \\
& GPT3 & 42.11$\pm$11.64 & 45.79$\pm$3.36 & 54.05$\pm$2.29 \\
\cline{2-5}
\hline
\multirow{13}{*}{HasProperty} & BERT$_{ base }$ & 20.93$\pm$6.28 & 26.74$\pm$2.70 & 37.05$\pm$2.55 \\
& BERT$_{ large }$ & 23.26$\pm$6.52 & 20.47$\pm$2.82 & 31.93$\pm$2.71 \\
\cline{2-5}
& ALBERT1$_{ base }$ & 16.67$\pm$5.82 & 35.00$\pm$3.53 & 49.40$\pm$2.72 \\
& ALBERT1$_{ large }$ & 14.29$\pm$5.46 & 30.00$\pm$3.25 & 41.33$\pm$2.89 \\
& ALBERT1$_{ xlarge }$ & 9.52$\pm$4.58 & 19.76$\pm$3.43 & 33.69$\pm$2.94 \\
\cline{2-5}
& ALBERT2$_{ base }$ & 19.05$\pm$6.13 & 25.48$\pm$3.72 & 33.38$\pm$3.38 \\
& ALBERT2$_{ large }$ & 19.05$\pm$6.13 & 26.43$\pm$3.88 & 34.19$\pm$3.57 \\
& ALBERT2$_{ xlarge }$ & 14.29$\pm$5.46 & 20.71$\pm$3.33 & 30.07$\pm$3.10 \\
\cline{2-5}
& RoBERTa$_{ base }$ & 11.63$\pm$4.95 & 14.88$\pm$2.69 & 23.91$\pm$2.71 \\
& RoBERTa$_{ large }$ & 4.65$\pm$3.25 & 11.40$\pm$1.99 & 23.14$\pm$2.66 \\
\cline{2-5}
& GPT1 & 25.58$\pm$6.73 & 30.93$\pm$2.84 & 38.84$\pm$2.42 \\
& GPT2 & 30.23$\pm$7.09 & 31.40$\pm$2.61 & 37.67$\pm$2.44 \\
& GPT3 & 27.91$\pm$6.92 & 30.00$\pm$2.53 & 35.72$\pm$2.27 \\
\cline{2-5}
\hline
\multirow{13}{*}{CapableOf} & BERT$_{ base }$ & 0.00$\pm$0.00 & 62.00$\pm$8.60 & 69.80$\pm$3.25 \\
& BERT$_{ large }$ & 40.00$\pm$24.49 & 56.00$\pm$5.10 & 73.60$\pm$2.54 \\
\cline{2-5}
& ALBERT1$_{ base }$ & 40.00$\pm$24.49 & 46.00$\pm$5.10 & 60.80$\pm$5.46 \\
& ALBERT1$_{ large }$ & 0.00$\pm$0.00 & 50.00$\pm$10.00 & 70.40$\pm$3.20 \\
& ALBERT1$_{ xlarge }$ & 80.00$\pm$20.00 & 54.00$\pm$5.10 & 60.80$\pm$2.24 \\
\cline{2-5}
& ALBERT2$_{ base }$ & 40.00$\pm$24.49 & 50.00$\pm$7.75 & 67.40$\pm$1.63 \\
& ALBERT2$_{ large }$ & 20.00$\pm$20.00 & 32.00$\pm$8.00 & 52.40$\pm$2.25 \\
& ALBERT2$_{ xlarge }$ & 20.00$\pm$20.00 & 50.00$\pm$5.48 & 68.20$\pm$2.35 \\
\cline{2-5}
& RoBERTa$_{ base }$ & 20.00$\pm$20.00 & 44.00$\pm$6.00 & 55.40$\pm$3.59 \\
& RoBERTa$_{ large }$ & 40.00$\pm$24.49 & 36.00$\pm$7.48 & 55.80$\pm$2.40 \\
\cline{2-5}
& GPT1 & 20.00$\pm$20.00 & 36.00$\pm$8.72 & 61.40$\pm$2.38 \\
& GPT2 & 20.00$\pm$20.00 & 36.00$\pm$9.27 & 60.20$\pm$2.35 \\
& GPT3 & 20.00$\pm$20.00 & 28.00$\pm$5.83 & 52.20$\pm$1.46 \\
\cline{2-5}
\hline
\end{tabular}
{\footnotesize
\begin{tablenotes}
\item[] Data presented as mean $\pm$ standard error of mean
\item[*] Overlap@K = The ratio of overlapping words in top K predictions for two opposite relations.
\end{tablenotes}
}
\end{threeparttable}
\end{table*}

\begin{table}[h!]
%\scriptysize
\centering
\begin{threeparttable}

\caption{\textbf{Results of missed prediction ratio at top K predictions between the opposite relations. }}
\label{tab:intergrade_results}
\begin{tabular}{c|ccc|ccc}
\hline
\multirow{3}{*}{Model} & \multicolumn{6}{c}{Miss@K\tnote{*}} \\
\cline{2-7}
& \multicolumn{3}{c|}{Synonym / Antonym} & \multicolumn{3}{c}{Antonym / Synonym} \\
\cline{2-7}

& 1 & 10 & 100 & 1 & 10 & 100 \\
\hline
\hline
BERT$_{ base }$ & 9.40$\pm$0.50 & 29.12$\pm$0.77 & 50.91$\pm$0.84 & 7.02$\pm$0.41 & 17.64$\pm$0.59 & 32.16$\pm$0.74 \\
BERT$_{ large }$ & 12.89$\pm$0.57 & 32.24$\pm$0.79 & 53.48$\pm$0.84 &  4.95$\pm$0.34 & 17.51$\pm$0.61 & 31.63$\pm$0.75 \\
\hline
ALBERT1$_{ base }$ & 1.76$\pm$0.24 & 18.78$\pm$0.68 & 38.46$\pm$0.84 & 9.77$\pm$0.46 & 16.93$\pm$0.58 & 29.64$\pm$0.71 \\
ALBERT1$_{ large }$ & 3.72$\pm$0.33 & 19.42$\pm$0.68 & 39.95$\pm$0.85 &  7.43$\pm$0.41 & 14.83$\pm$0.55 & 26.45$\pm$0.69\\
ALBERT1$_{ xlarge }$ & 9.50$\pm$0.51 & 26.32$\pm$0.76 & 45.97$\pm$0.86 & 5.55$\pm$0.37 & 13.52$\pm$0.53 & 27.23$\pm$0.70 \\
\hline
ALBERT2$_{ base }$ & 6.45$\pm$0.42 & 24.34$\pm$0.75 & 43.35$\pm$0.86 &  8.72$\pm$0.44 & 17.57$\pm$0.60 & 31.93$\pm$0.74\\
ALBERT2$_{ large }$ & 9.69$\pm$0.52 & 29.42$\pm$0.79 & 51.51$\pm$0.86 & 8.10$\pm$0.43 & 23.09$\pm$0.67 & 40.16$\pm$0.76 \\
ALBERT2$_{ xlarge }$ & 12.72$\pm$0.58  & 33.34$\pm$0.82  & 54.71$\pm$0.85  & 8.34$\pm$0.44  & 23.39$\pm$0.67  & 40.66$\pm$0.78 \\
\hline
RoBERTa$_{ base }$ & 2.90$\pm$0.29 & 12.24$\pm$0.55 & 28.34$\pm$0.76 & 1.33$\pm$0.18 & 10.34$\pm$0.48 & 27.74$\pm$0.69 \\
RoBERTa$_{ large }$ & 10.48$\pm$0.53 & 30.85$\pm$0.80 & 52.75$\pm$0.86 & 10.56$\pm$0.48 & 23.25$\pm$0.67 & 39.78$\pm$0.77 \\
\hline
GPT1 & 2.88$\pm$0.27 & 13.03$\pm$0.56 & 32.29$\pm$0.80 &  5.02$\pm$0.34 & 16.57$\pm$0.57 & 31.77$\pm$0.72\\
GPT2 & 5.56$\pm$0.38  & 18.02$\pm$0.65  & 39.11$\pm$0.84  & 2.73$\pm$0.26  & 17.83$\pm$0.61  & 38.34$\pm$0.75 \\
GPT3 & 3.06$\pm$0.30  & 18.91$\pm$0.80  & 52.98$\pm$1.22  & 0.24$\pm$0.08  & 6.45$\pm$0.43  & 39.02$\pm$1.00 \\
\hline

\end{tabular}
{\footnotesize
\begin{tablenotes}
\item[] Data presented as mean $\pm$ standard error of mean
\item[*] Miss@K = The ratio of words that are considered right when graded with the opposite relation (undesirable) in top K predictions.
\end{tablenotes}
}
\end{threeparttable}
\end{table}

\begin{table}[h!]
\centering
\footnotesize
\begin{threeparttable}
% \small
\caption{\textbf{Comparison of macro average hits@K on pretrained and fune-tuned BERT models. Herin, Avg. indicates average.}}
%The macro avg. equally average the results of all relations, while the micro avg. weighted average the results of the relations according to their portion.}
\label{tab:hits@K_finetune}
\begin{tabular}{c|c|c|cccccccccccc}
\hline
\multirow{3}{*}{\begin{tabular}[c]{@{}c@{}}Train \\ Status\end{tabular}} & \multirow{3}{*}{Model} & \multirow{3}{*}{\begin{tabular}[c]{@{}c@{}}Avg. \\ Type\end{tabular}} & \multicolumn{12}{c}{Hits@K} \\ \cline{4-15} 
 & & & \multicolumn{4}{c|}{1} & \multicolumn{4}{c|}{10} & \multicolumn{4}{c}{100} \\ \cline{4-15} 
 & & & Fold 0 & Fold 1 & Fold 2 & \multicolumn{1}{c|}{Avg.*} & Fold 0 & Fold 1 & Fold 2 & \multicolumn{1}{c|}{Avg.} & Fold 0 & Fold 1 & Fold 2 & Avg. \\ \hline\hline
\multirow{4}{*}{Pretrained} & \multirow{2}{*}{BERT$_{base}$} & Micro & 6.67 & 5.05 & 6.19 & \multicolumn{1}{c|}{5.97} & 17.72 & 16.31 & 17.38 & \multicolumn{1}{c|}{17.14} & 36.13 & 33.78 & 34.80 & 34.90 \\
 & & Macro & 7.93 & 7.74 & 7.79 & \multicolumn{1}{c|}{7.82} & 19.69 & 19.65 & 19.62 & \multicolumn{1}{c|}{19.65} & 37.29 & 37.25 & 37.21 & 37.25 \\ \cline{2-15} 
 & \multirow{2}{*}{BERT$_{large}$} & Micro & 8.00 & 5.51 & 7.31 & \multicolumn{1}{c|}{6.94} & 20.03 & 17.65 & 18.27 & \multicolumn{1}{c|}{18.65} & 38.17 &36.31 & 36.07 & 36.85 \\
 & & Macro & 7.27 & 7.16 & 7.24 & \multicolumn{1}{c|}{7.22} & 19.31 & 19.27 & 19.19 & \multicolumn{1}{c|}{19.26} & 37.09 & 36.90 & 36.88 & 36.96\\ \hline
\multirow{4}{*}{Fine-tuned} & \multirow{2}{*}{BERT$_{base}$} & Micro & 17.73 & 16.40 & 16.25 & \multicolumn{1}{c|}{16.79} & 42.52 & 40.67 & 40.39 & \multicolumn{1}{c|}{41.19} & 66.87 & 62.82 & 63.80 & 64.50 \\
 & & Macro & 29.58 & 29.60 & 29.77 & \multicolumn{1}{c|}{29.65} & 52.20 & 52.22 & 52.16 & \multicolumn{1}{c|}{52.19} & 70.88 & 70.97 & 70.89 & 70.91 \\ \cline{2-15} 
 & \multirow{2}{*}{BERT$_{large}$} & Micro & 21.39 & 19.14 & 19.55 & \multicolumn{1}{c|}{20.03} & 46.93 & 43.63 & 43.93 & \multicolumn{1}{c|}{44.83} & 72.23 & 68.20 & 70.23 & 70.22 \\
 & & Macro & 32.41 & 32.35 & 32.53 & \multicolumn{1}{c|}{32.43} & 55.86 & 55.79 & 55.96 & \multicolumn{1}{c|}{55.87} & 74.28 & 74.26 & 74.23 & 74.26 \\ \hline
\end{tabular}
{\footnotesize
\begin{tablenotes}
\item[*] Average of the performances of Fold 1, 2 and 3
%\item[\textdagger] Questions succeed or failed to predict by all experimental models commonly
% \item[\textdaggerdbl] Symbol 1
% \item[\S] Symbol 1  
% \item[$\|$] Symbol 1
% \item[$\!$\#] Symbol 1          
\end{tablenotes}
}
% {\small
% \begin{tablenotes}
% \item[a] testing
% \end{tablenotes}
% }
\end{threeparttable}
\end{table}

\begin{table}[h!]
\centering
\begin{threeparttable}
% \small
\caption{\textbf{Results of overlapping ratio at top K predictions of fine-tuned BERT models between the opposite relations. Herein, `-' indicates when it is impossible to evaluate overlap@K since there is no example of an opposite relation for all subject-relation pairs in the same test fold.}}
%The macro avg. equally average the results of all relations, while the micro avg. weighted average the results of the relations according to their portion.}
\label{tab:overlap@K_finetune}
\begin{tabular}{c|c|ccc|ccc|ccc}
\hline
\multicolumn{1}{c|}{} & \multicolumn{1}{c|}{} & \multicolumn{9}{c}{{Overlap@K*}} \\
\cline{3-11}
\multicolumn{1}{c|}{} & \multicolumn{1}{c|}{} & \multicolumn{3}{c|}{1} & \multicolumn{3}{c|}{10} & \multicolumn{3}{c}{100} \\
\cline{3-11}
\multicolumn{1}{c|}{\multirow{-3}{*}{Relation}} & \multicolumn{1}{c|}{\multirow{-3}{*}{Model}} & Fold 0 & Fold 1 & Fold 2 & Fold 0 & Fold 1 & Fold 2 & Fold 0 & Fold 1 & Fold 2 \\
\hline \hline
\multirow{2}{*}{Desires} & BERT$_{base}$ & 0.00 & 100.00 & 50.00 & 55.00 & 60.00 & 50.00 & 63.50 & 72.50 & 67.50 \\
 & BERT$_{large}$ & 100.00 & 100.00 & 50.00 & 75.00 & 75.00 & 60.00 & 62.50 & 68.00 & 64.50 \\
 \hline
\multirow{2}{*}{HasProperty} & BERT$_{base}$ & 100.00 & 20.00 & 0.00 & 60.00 & 38.00 & 40.00 & 43.00 & 52.00 & 45.67 \\
 & BERT$_{large}$ & 0.00 & 40.00 & 0.00 & 30.00 & 20.00 & 30.00 & 32.00 & 41.60 & 45.67 \\
 \hline
\multirow{2}{*}{CapableOf} & BERT$_{base}$ & - & 100.00 & - & - & 60.00 & - & - & 78.00 & - \\
 & BERT$_{large}$ & - & 100.00 & - & - & 70.00 & - & - & 76.00 & - \\
 \hline
\multirow{2}{*}{Synonym} & BERT$_{base}$ & 78.82 & 80.00 & 70.59 & 69.88 & 66.94 & 68.37 & 71.32 & 70.59 & 70.17 \\
 & BERT$_{large}$ & 47.06 & 54.44 & 44.94 & 51.47 & 50.28 & 49.55 & 56.16 & 56.76 & 55.16 \\
\hline

\end{tabular}
{\footnotesize
\begin{tablenotes}
\item[*] Overlap@K = The ratio of overlapping words in top K predictions for two opposite relations.
\end{tablenotes}
}
% {\small
% \begin{tablenotes}
% \item[a] testing
% \end{tablenotes}
% }
\end{threeparttable}
\end{table}

\begin{table}[h!]
\centering
\begin{threeparttable}
% \small
\caption{\textbf{Results of missed prediction ratio at top K predictions of fine-tuned BERT models between the opposite relations.}}
%The macro avg. equally average the results of all relations, while the micro avg. weighted average the results of the relations according to their portion.}
\label{tab:miss@K_finetune}
\begin{tabular}{c|c|ccc|ccc|ccc}
\hline
\multicolumn{2}{c|}{} & \multicolumn{9}{c}{{Miss@K*}} \\
\cline{3-11}
\multicolumn{2}{c|}{} & \multicolumn{3}{c|}{1} & \multicolumn{3}{c|}{10} & \multicolumn{3}{c}{100} \\
\cline{3-11}
\multicolumn{2}{c|}{\multirow{-3}{*}{Model}} & Fold 0 & Fold 1 & Fold 2 & Fold 0 & Fold 1 & Fold 2 & Fold 0 & Fold 1 & Fold 2 \\
\hline \hline
 & BERT$_{base}$ & 3.33 & 2.89 & 3.87 & 29.57 & 28.48 & 31.22 & 57.35 & 57.50 & 54.98 \\
\multirow{-2}{*}{Synonym / Antonym} & BERT$_{large}$ & 4.42 & 4.21 & 4.36 & 29.53 & 29.63 & 30.35 & 56.34 & 58.43 & 58.21 \\
\hline
 & BERT$_{base}$ & 17.16 & 13.63 & 13.63 & 37.56 & 35.20 & 35.24 & 58.91 & 56.75 & 57.78 \\
\multirow{-2}{*}{Antonym / Synonym} & BERT$_{large}$ & 8.69 & 8.57 & 8.20 & 31.50 & 27.40 & 28.24 & 55.26 & 49.87 & 52.70 \\
\hline

\end{tabular}
{\footnotesize
\begin{tablenotes}
\item[*] Miss@K = The ratio of words that are considered right when graded with the opposite relation (undesirable) in top K predictions.
\end{tablenotes}
}
% {\small
% \begin{tablenotes}
% \item[a] testing
% \end{tablenotes}
% }
\end{threeparttable}
\end{table}

\begin{table*}[]
\centering
\begin{threeparttable}
% \footnotesize
\caption{\textbf{Total proportion of \textit{has answer} questions with less lexical overlapping (\textit{similarity} < 0.2) and proportions of question types for each model in SQuAD.}}
 \label{tab:question_types_squad}
\begin{tabular}{c|c|c|c|c|c|c|c|c}
\hline
\multirow{4}{*}{Model} & \multirow{4}{*}{Status\tnote{*}} & \multicolumn{6}{c|}{Question type} & \multirow{4}{*}{\begin{tabular}[c]{@{}c@{}}Number of \\examples\end{tabular}} \\ \cline{3-8}
 &  & \multicolumn{2}{c|}{Semantic variation} & \multirow{2}{*}{\begin{tabular}[c]{@{}c@{}}Multiple\\sentence\\reasoning\end{tabular}} & \multirow{2}{*}{\begin{tabular}[c]{@{}c@{}}No\\semantic\\variation\end{tabular}} & \multirow{3}{*}{Typo} & \multirow{3}{*}{Others} &  \\ \cline{3-4}
 &  & Synonymy & \multicolumn{1}{c|}{\begin{tabular}[c]{@{}c@{}}Commonsense \\ Knowledge\end{tabular}} &  &  &  &  &  \\ \hline\hline
BERT$_{large}$&Correct&27.29&22.54&11.82&\textbf{54.14}&1.22&1.10&905\\
BERT$_{large}$&Incorrect&30.71&\textbf{49.46}&19.84&27.45&1.63&2.99&368\\
\hline
ALBERT1$_{xlarge}$&Correct&28.06&24.40&12.46&\textbf{51.52}&1.15&1.15&955\\
ALBERT1$_{xlarge}$&Incorrect&28.93&\textbf{48.11}&19.18&31.13&1.89&3.14&318\\
\hline
ALBERT2$_{xlarge}$&Correct&28.22&24.41&12.56&\textbf{51.80}&1.13&0.93&971\\
ALBERT2$_{xlarge}$&Incorrect&28.48&\textbf{49.34}&19.21&29.14&1.99&3.97&302\\
\hline
RoBERTa$_{large}$&Correct&28.81&25.12&12.36&\textbf{50.55}&1.30&1.10&1003\\
RoBERTa$_{large}$&Incorrect&26.30&\textbf{49.63}&20.74&31.11&1.48&3.70&270\\
\hline
Overall\tnote{\textdagger}&Correct&27.29&17.86&10.97&\textbf{57.81}&1.27&0.84&711\\
Overall&Incorrect&25.69&\textbf{55.96}&22.02&24.77&1.83&4.59&109\\
\hline\hline
\multicolumn{2}{c|}{Total proportion}&28.28&30.32&14.14&46.43&1.34&1.65&1273\\
\hline
\end{tabular}

{\footnotesize
\begin{tablenotes}
\item[] The categories can be tagged with duplicates except for semantic variation and no semantic variation.
\item[*] Correct: Questions correctly predicted by the model,  Incorrect: Questions incorrectly predicted by the model
\item[\textdagger] Questions succeed or failed to predict by all experimental models commonly
% \item[\textdaggerdbl] Symbol 1
% \item[\S] Symbol 1  
% \item[$\|$] Symbol 1
% \item[$\!$\#] Symbol 1          
\end{tablenotes}
}
\end{threeparttable}
\end{table*}

\begin{table*}[]
\centering
\begin{threeparttable}
% \footnotesize
\caption{\textbf{Total proportion of question types for each model in ReCORD.}}
 \label{tab:question_types_record}
\begin{tabular}{c|c|c|c|c|c|c|c|c}
\hline
\multirow{4}{*}{Model} & \multirow{4}{*}{Status\tnote{*}} & \multicolumn{6}{c|}{Question type} & \multirow{4}{*}{\begin{tabular}[c]{@{}c@{}}Number of \\examples\end{tabular}} \\ \cline{3-8}
 &  & \multicolumn{2}{c|}{Semantic variation} & \multirow{2}{*}{\begin{tabular}[c]{@{}c@{}}Multiple\\sentence\\reasoning\end{tabular}} & \multirow{2}{*}{\begin{tabular}[c]{@{}c@{}}No\\semantic\\variation\end{tabular}} & \multirow{3}{*}{Typo} & \multirow{3}{*}{Others} &  \\ \cline{3-4}
 &  & Synonymy & \multicolumn{1}{c|}{\begin{tabular}[c]{@{}c@{}}Commonsense \\ Knowledge\end{tabular}} &  &  &  &  &  \\ \hline\hline
BERT$_{large}$&Correct&13.38&88.73&13.38&1.41&0.00&4.23&142\\
BERT$_{large}$&Incorrect&17.24&72.41&6.90&3.45&3.45&15.52&58\\
\hline
ALBERT1$_{xlarge}$&Correct&13.70&88.36&13.01&1.37&0.68&3.42&146\\
ALBERT1$_{xlarge}$&Incorrect&16.67&72.22&7.41&3.70&1.85&18.52&54\\
\hline
ALBERT2$_{xlarge}$&Correct&12.58&88.74&12.58&1.32&1.32&3.31&151\\
ALBERT2$_{xlarge}$&Incorrect&20.41&69.39&8.16&4.08&0.00&20.41&49\\
\hline
RoBERTa$_{large}$&Correct&11.84&91.45&12.50&1.32&0.66&1.97&152\\
RoBERTa$_{large}$&Incorrect&22.92&60.42&8.33&4.17&2.08&25.00&48\\
\hline
Overall\tnote{\textdagger}&Correct&11.82&90.91&14.55&1.82&0.00&1.82&110\\
Overall&Incorrect&11.82&42.11&5.26&10.53&0.00&42.11&19\\
\hline\hline
\multicolumn{2}{c|}{Total proportion}&14.50&84.00&11.50&2.00&1.00&7.50&200\\
\hline
\end{tabular}

{\footnotesize
\begin{tablenotes}
\item[] The categories can be tagged with duplicates except for semantic variation and no semantic variation.
\item[*] Correct: Questions correctly predicted by the model,  Incorrect: Questions incorrectly predicted by the model
\item[\textdagger] Questions succeed or failed to predict by all experimental models commonly
% \item[\textdaggerdbl] Symbol 1
% \item[\S] Symbol 1  
% \item[$\|$] Symbol 1
% \item[$\!$\#] Symbol 1          
\end{tablenotes}
}
\end{threeparttable}
\end{table*}

\begin{table*}[!ht]
 \centering
 \footnotesize
 \begin{threeparttable}
  \caption{\textbf{Examples of manual external commonsense knowledge integration in SQuAD.}}
  \label{tab:example_of_manually_integrating_in_record}
\begin{tabular}{m{0.2\columnwidth}|m{0.3\columnwidth}|m{0.33\columnwidth}}\hline
\multicolumn{1}{c|}{Required Knowledge\tnote{*}} & \multicolumn{1}{c|}{Question} & \multicolumn{1}{c}{Context} \\\hline\hline
\textbf{\textcolor{blue}{uv} \textcolor{red}{Synonym} \textcolor{teal}{ultraviolet radiation}} & \_\_\_\_\_ Helps the biospher from \textbf{\textcolor{blue}{UV}, \textcolor{red}{which is the same as} \textcolor{teal}{ultraviolet radiation}.} &  ... the high-altitude ozone layer helps protect the biosphere from \textbf{\textcolor{teal}{ultraviolet radiation}}, ... \\\hline
\textbf{\textcolor{blue}{rare} \textcolor{red}{Antonym} \textcolor{teal}{frequent}} & How \textbf{\textcolor{teal}{frequent}} is snow in the Southwest of the state? & ... But snow is very \textbf{\textcolor{blue}{rare}, \textcolor{red}{which is the opposite of} \textcolor{teal}{frequent},} in the Southwest of the state, ...\\\hline
\textbf{\textcolor{blue}{punishment} \textcolor{red}{RelatedTo} \textcolor{teal}{sentence}} & Why would one want to give more \textbf{\textcolor{blue}{punishment}, \textcolor{red}{which is related to} \textcolor{teal}{sentence}}? &... the judge increased her \textbf{\textcolor{teal}{sentence}} from 40 to 60 days. ...\\\hline
% \begin{tabular}[c]{@{}c@{}}Multi-sentence\\reasoning\end{tabular} & Hints for solving questions are shattered in multiple sentences.  & \begin{tabular}{@{}p{6.5cm}@{}}\textbf{Question}: Why did \textcolor{red}{\textbf{France}} choose to give up continental lands?\\\textbf{Context}: ...  \textcolor{blue}{\textbf{France}} chose to cede the former, ... \textcolor{blue}{\textbf{They}} viewed the economic value of the Caribbean islands' sugar cane ...\end{tabular} \\\hline
% Others & The labeled answer is incorrect. &  \begin{tabular}{@{}p{6.5cm}@{}}\textbf{Question}: Who \textcolor{red}{\textbf{won the battle}} of Lake George?\\\textbf{Context}: ...  The \textcolor{blue}{\textbf{battle ended inconclusively}}, with both sides withdrawing from the field.  ...\end{tabular} \\\hline
% Typo & There exist typing errors in the question or context. & \begin{tabular}{@{}p{6.5cm}@{}}\textbf{Question}: What kind of measurements define \textbf{\textcolor{red}{accelerlations}}?\\\textbf{Context}... \textbf{\textcolor{blue}{Accelerations}} can be defined through kinematic measurements. ...\end{tabular}\\\hline
\end{tabular}
{\footnotesize
\begin{tablenotes}
\item[*] \textcolor{blue}{The blue words indicate the subject term of the triple} \newline \textcolor{red}{The red words indicate the relation or the relation's relative pronoun template of the triple} \newline \textcolor{teal}{The turquoise words indicate the object term of the triple} \newline 
% \item[\textdaggerdbl] Symbol 1
% \item[\S] Symbol 1  
% \item[$\|$] Symbol 1
% \item[$\!$\#] Symbol 1          
\end{tablenotes}
}
\end{threeparttable}
\end{table*}

\begin{table*}[!ht]
 \centering
 \footnotesize
 \begin{threeparttable}
  \caption{\textbf{Examples of manual external commonsense knowledge integration in ReCoRD.}}
  \label{tab:example_of_manually_integrating_in_squad}
\begin{tabular}{m{0.2\columnwidth}|m{0.3\columnwidth}|m{0.33\columnwidth}}\hline
\multicolumn{1}{c|}{Required Knowledge\tnote{*}} & \multicolumn{1}{c|}{Question} & \multicolumn{1}{c}{Context} \\\hline\hline
\textbf{\textcolor{blue}{uv} \textcolor{red}{Synonym} \textcolor{teal}{ultraviolet radiation}} & \_\_\_\_\_ Helps the biospher from \textbf{\textcolor{blue}{UV}} &  ... the high-altitude ozone layer helps protect the biosphere from \textbf{\textcolor{teal}{ultraviolet radiation}}, ... \textbf{@highlight \textcolor{blue}{uv} \textcolor{red}{is the same as} \textcolor{teal}{ultraviolet radiation}}. \\\hline
\textbf{\textcolor{blue}{rare} \textcolor{red}{Antonym} \textcolor{teal}{frequent}} & How \textbf{\textcolor{teal}{frequent}} is snow in the Southwest of the state? & ... But snow is very \textbf{\textcolor{blue}{rare}} in the Southwest of the state, ... \textbf{@highlight \textcolor{blue}{rare} \textcolor{red}{is the opposite of} \textcolor{teal}{frequent}}\\\hline
\textbf{\textcolor{blue}{punishment} \textcolor{red}{RelatedTo} \textcolor{teal}{sentence}} & Why would one want to give more \textbf{\textcolor{blue}{punishment}}? &... the judge increased her \textbf{\textcolor{teal}{sentence}} from 40 to 60 days. ...\textbf{@highlight
\textcolor{blue}{punishment}
\textcolor{red}{is related to} \textcolor{teal}{sentence}}\\\hline
% \begin{tabular}[c]{@{}c@{}}Multi-sentence\\reasoning\end{tabular} & Hints for solving questions are shattered in multiple sentences.  & \begin{tabular}{@{}p{6.5cm}@{}}\textbf{Question}: Why did \textcolor{red}{\textbf{France}} choose to give up continental lands?\\\textbf{Context}: ...  \textcolor{blue}{\textbf{France}} chose to cede the former, ... \textcolor{blue}{\textbf{They}} viewed the economic value of the Caribbean islands' sugar cane ...\end{tabular} \\\hline
% Others & The labeled answer is incorrect. &  \begin{tabular}{@{}p{6.5cm}@{}}\textbf{Question}: Who \textcolor{red}{\textbf{won the battle}} of Lake George?\\\textbf{Context}: ...  The \textcolor{blue}{\textbf{battle ended inconclusively}}, with both sides withdrawing from the field.  ...\end{tabular} \\\hline
% Typo & There exist typing errors in the question or context. & \begin{tabular}{@{}p{6.5cm}@{}}\textbf{Question}: What kind of measurements define \textbf{\textcolor{red}{accelerlations}}?\\\textbf{Context}... \textbf{\textcolor{blue}{Accelerations}} can be defined through kinematic measurements. ...\end{tabular}\\\hline
\end{tabular}
{\footnotesize
\begin{tablenotes}
\item[*] \textcolor{blue}{The blue words indicate the subject term of the triple} \newline \textcolor{red}{The red words indicate the relation or the relation's relative pronoun template of the triple} \newline \textcolor{teal}{The turquoise words indicate the object term of the triple} \newline 
% \item[\textdaggerdbl] Symbol 1
% \item[\S] Symbol 1  
% \item[$\|$] Symbol 1
% \item[$\!$\#] Symbol 1          
\end{tablenotes}
}
\end{threeparttable}
\end{table*}

\begin{table}[h!]
\centering
 \begin{threeparttable}
  \caption{\textbf{Experimental results of the knowledge integration test.}}
  \label{tab:result_of_manual_integration}

\begin{tabular}{c|c|c|cc|cc|c}
\hline
\multicolumn{1}{c|}{\multirow{2}{*}{Dataset}}&
\multicolumn{1}{c|}{\multirow{2}{*}{Model}}&
\multicolumn{1}{c|}{\multirow{2}{*}{Status}}& \multicolumn{2}{c|}{Original text} & \multicolumn{2}{c|}{Knowledge integrated text}&
\multicolumn{1}{c}{\multirow{2}{*}{$\Delta$ F1}}\\

\cline{4-7}
&\multicolumn{1}{c|}{} & \multicolumn{1}{c|}{} & EM\tnote{*} & F1\tnote{\textdagger} & \multicolumn{1}{c}{\quad\enspace EM\quad\enspace} & \multicolumn{1}{c|}{F1}&\\
\hline\hline
\multicolumn{1}{c|}{\multirow{8}{*}{SQuAD}}&\multicolumn{1}{c|}{\multirow{2}{*}{BERT$_{large}$}} & Correct & 100.00 & 100.00 & 93.37 & 95.93 & -4.07 \\
&& Incorrect& 0.00 & 28.08 & 11.53 & 37.11&\textbf{9.53}\\
\cline{2-8}
&\multicolumn{1}{c|}{\multirow{2}{*}{ALBERT1$_{xlarge}$}} & Correct& 100.00 & 100.00& 87.22 & 89.73&-10.27\\
&&Incorrect& 0.00 & 29.86& 18.75 & 46.64&\textbf{16.78}\\
\cline{2-8}
&\multicolumn{1}{c|}{\multirow{2}{*}{ALBERT2$_{xlarge}$}} & Correct& 100.00 & 100.00& 85.71 & 90.59&-9.41\\
&&Incorrect& 0.00 & 29.86& 14.49 & 44.37&\textbf{13.02}\\
\cline{2-8}
&\multicolumn{1}{c|}{\multirow{2}{*}{RoBERTa$_{large}$}} & Correct& 100.00 & 100.00& 90.00 & 93.48&-6.52\\
&&Incorrect& 0.00 & 32.94& 14.06 & 43.43&\textbf{10.49}\\
\hline
\multicolumn{1}{c|}{\multirow{8}{*}{ReCORD}}&\multicolumn{1}{c|}{\multirow{2}{*}{BERT$_{large}$}} & Correct& 100.00 & 100.00 & 98.68 & 98.68&-1.32\\
&& Incorrect& 0.00 & 2.81 & 10.71 & 10.71&\textbf{7.90}\\
\cline{2-8}
&\multicolumn{1}{c|}{\multirow{2}{*}{ALBERT1$_{xlarge}$}} & Correct & 100.00 & 100.00& 97.40 & 97.40&-2.60\\
&&Incorrect& 0.00 & 11.77& 14.81 & 27.04&\textbf{15.27}\\
\cline{2-8}
&\multicolumn{1}{c|}{\multirow{2}{*}{ALBERT2$_{xlarge}$}} & Correct &  100.00 & 100.00& 98.77 & 98.77&-1.23\\
&&Incorrect& 0.00 & 15.51& 8.70 & 21.96&\textbf{6.45}\\
\cline{2-8}
&\multicolumn{1}{c|}{\multirow{2}{*}{RoBERTa$_{large}$}} & Correct & 100.00 & 100.00& 98.72 & 98.72&-1.28\\
&&Incorrect& 0.00 & 19.55& 3.85 & 23.40&\textbf{3.85}\\
\hline
\end{tabular}
{\footnotesize
\begin{tablenotes}
%\item[] Data presented as mean $\pm$ standard error of mean
\item[*] F1 score
 \item[\textdagger] Exact match score %$EM$=a\,binary\,measure\,of\,whether\,the\,predicted\,output\,exactly\,matches\,the ground\,truth
% \item[\S] Symbol 1  
% \item[$\|$] Symbol 1
% \item[$\!$\#] Symbol 1          
\end{tablenotes}
}
\end{threeparttable}
\end{table}

\newpage

\newpage
\appendix
\onecolumn
\setcounter{figure}{0}
\setcounter{table}{0}
\setcounter{equation}{0}
\setcounter{page}{1}
\section{Details on the Templates}
\label{apx:details_on_the_templates}
\begin{table*}[!th]
  \centering
  \scriptsize
  \caption{\textbf{This table presents details on the templates of the relations in the ConceptNet used in the experiments. Note that, difference from BERT, ALBERT and RoBERTa models, only original templates where [[OBJ]] comes at the end of the sentence are utilized for GPT models because of the characteristics of GPT models that decode input sentences from left to right.}}
  \label{tab:template}
  %\begin{center}
   %\begin{adjustwidth}{-1cm}{}

  \begin{tabular}{c|m{10.0cm}|c|c}
    \hline
    \textbf{Relation} & \multicolumn{1}{c|}{\textbf{Original Templates}} & \# of samples & \begin{tabular}[c]{@{}c@{}}\# of \\average answers\end{tabular} \\
    \hline\hline
    \multirow{1}{*}{RelatedTo} & `[[SUBJ]] is related to [[OBJ]] .'; `[[SUBJ]] is like [[OBJ]] .'; `[[SUBJ]] is associated to [[OBJ]] .' & \multirow{1}{*}{252,003} &\multirow{1}{*}{2.85}\\
    \hline
    \multirow{1}{*}{HasContext} & `[[SUBJ]] is used in the context of [[OBJ]] .'; `[[SUBJ]] is a kind of [[OBJ]] .'; `[[SUBJ]] is a type of [[OBJ]] .'& \multirow{1}{*}{84,755} &\multirow{1}{*}{1.22}\\
    \hline
    \multirow{1}{*}{IsA} & `[[SUBJ]] is a [[OBJ]] .'; `[[SUBJ]] is used in the region of [[OBJ]] .'; `[[SUBJ]] is utilized in [[OBJ]] .' & \multirow{1}{*}{67,208 } &\multirow{1}{*}{1.22}\\
    \hline
    \multirow{1}{*}{DerivedFrom} & `[[SUBJ]] is derived from [[OBJ]] .'; `[[SUBJ]] is stemmed from [[OBJ]] .'; `[[SUBJ]] is originated from [[OBJ]] .'  & \multirow{1}{*}{52,297} &\multirow{1}{*}{1.05}\\
    \hline
    \multirow{1}{*}{Synonym} & `[[SUBJ]] is same with [[OBJ]] .'; `[[SUBJ]] is same as [[OBJ]] .'; `[[SUBJ]] and [[OBJ]] are the same .'; `[[SUBJ]] and [[OBJ]] are the synonym .'& \multirow{1}{*}{20,065} &\multirow{1}{*}{1.64}\\
    \hline
    \multirow{1}{*}{FormOf} & `[[OBJ]] is the root word of [[SUBJ]] .'; `[[SUBJ]] is come from [[OBJ]] .'; `[[SUBJ]] is originated from [[OBJ]] .' & \multirow{1}{*}{18,405} &\multirow{1}{*}{1.01}\\
    \hline
    \multirow{1}{*}{SimilarTo} & `[[SUBJ]] is similar to [[OBJ]] .'; `[[SUBJ]] is like to [[OBJ]] .'; `[[SUBJ]] are comparable to [[OBJ]]'& \multirow{1}{*}{ 7,266 } &\multirow{1}{*}{1.35}\\
    \hline
    \multirow{1}{*}{EtymologicallyRelatedTo} & `[[SUBJ]] is etymologically related to [[OBJ]] .'; `[[SUBJ]] is etymologically like [[OBJ]] .'; `[[SUBJ]] is etymologically associated to [[OBJ]] .'& \multirow{1}{*}{ 7,081} &\multirow{1}{*}{1.27}\\
    \hline
    \multirow{1}{*}{AtLocation} & `Something you find at [[OBJ]] is [[SUBJ]] .'; `you are likely to find [[OBJ]] in [[SUBJ]] .'; `[[[OBJ]] is located in [[SUBJ]] .& \multirow{1}{*}{ 6,532} &\multirow{1}{*}{1.95}\\
    \hline
    \multirow{1}{*}{MannerOf} & `[[SUBJ]] is a way to [[OBJ]] .'; `[[SUBJ]] is a manner to [[OBJ]] .'; `[[SUBJ]] is a means to [[OBJ]] .'& \multirow{1}{*}{ 5,963} &\multirow{1}{*}{1.56}\\
    \hline
    \multirow{1}{*}{Antonym} & `[[SUBJ]] and [[OBJ]] are opposite .'; `[[SUBJ]] is the opposite of [[OBJ]] .'; `[[SUBJ]] and [[OBJ]] are antonym .' & \multirow{1}{*}{3,457} &\multirow{1}{*}{ 1.98}\\
    \hline
    \multirow{1}{*}{HasProperty} & `[[SUBJ]] can be [[OBJ]] .'; `[[SUBJ]] is [[OBJ]] .'; `[[SUBJ]] has [[OBJ]] .'  & \multirow{1}{*}{2,708} &\multirow{1}{*}{1.25}\\
    \hline
    \multirow{1}{*}{PartOf} & `[[SUBJ]] is part of [[OBJ]] .'; `[[OBJ]] is part of [[SUBJ]] .'; `[[SUBJ]] has [[OBJ]] .'; `[[SUBJ]] contains [[OBJ]] .'& \multirow{1}{*}{2,317} &\multirow{1}{*}{1.22}\\
    \hline
    \multirow{1}{*}{UsedFor} & `[[SUBJ]] may be used for [[OBJ]] .'; `[[SUBJ]] is used for [[OBJ]] .'; `[[SUBJ]] is for [[OBJ]] .'& \multirow{1}{*}{2,090} &\multirow{1}{*}{3.01}\\
    \hline
    \multirow{1}{*}{DistinctFrom} & `[[SUBJ]] is not [[OBJ]] .'; `[[SUBJ]] is different from [[OBJ]] .'; `[[SUBJ]] is distinct from [[OBJ]] .' & \multirow{1}{*}{1,184} &\multirow{1}{*}{2.08}\\
    
    \hline
    \multirow{1}{*}{HasPrerequisite} & `[[SUBJ]] requires [[OBJ]] .'; `Something you need to do before you [[SUBJ]] is [[OBJ]] .'; `If you want to [[SUBJ]] then you should [[OBJ]] .' &
    \multirow{1}{*}{1,127} &\multirow{1}{*}{3.06}\\
    \hline
    \multirow{1}{*}{HasSubevent} & `One of the things you do when you [[SUBJ]] is [[OBJ]] .'; `Something you might do while [[SUBJ]] is [[OBJ]] .' &
    \multirow{1}{*}{1,100} &\multirow{1}{*}{3.10}\\
    \hline
    \multirow{1}{*}{Causes} & `[[SUBJ]] causes [[OBJ]] .'; `The effect of [[SUBJ]] is that [[OBJ]] .'; `Something that might happen as a consequence of [[SUBJ]] is [[OBJ]] .'&
    \multirow{1}{*}{952} &\multirow{1}{*}{3.68}\\
    \hline
    \multirow{1}{*}{HasA} & `[[SUBJ]] contains [[OBJ]] .'; `[[SUBJ]] has [[OBJ]] .'; `[[SUBJ]] includes [[OBJ]] .' &
    \multirow{1}{*}{847} &\multirow{1}{*}{1.3}\\
    \hline
    \multirow{1}{*}{InstanceOf} & `[[SUBJ]] is an instance of [[OBJ]] .'; `[[SUBJ]] is an example of [[OBJ]] .'; `[[SUBJ]] is an illustration of [[OBJ]] .' &
    \multirow{1}{*}{738} &\multirow{1}{*}{1.03}\\
    \hline
    \multirow{1}{*}{CapableOf} & `[[SUBJ]] can [[OBJ]] .'; `[[SUBJ]] may [[OBJ]] .'; `[[SUBJ]] sometimes [[OBJ]] .' &
    \multirow{1}{*}{656} &\multirow{1}{*}{ 1.40}\\

    \hline
    % ReceivesAction & [[SUBJ]] can be [[OBJ]] . & 627 & 1.19\\ 
    \multirow{1}{*}{MotivatedByGoal} &`You would [[OBJ]] because you want to [[SUBJ]] .'; `You would [[OBJ]] because you like to [[SUBJ]] .'; `If you want to [[SUBJ]] then you should [[OBJ]] .'&
    \multirow{1}{*}{593} &\multirow{1}{*}{2.78}\\
    \hline
    % CausesDesire & [[SUBJ]] would make you want to [[OBJ]] . & 530 & 1.1\\ 
    \multirow{1}{*}{MadeOf} &`[[SUBJ]] can be made of [[OBJ]] .'; `[[SUBJ]] is made from [[OBJ]] .'; `[[SUBJ]] consists of [[OBJ]] .'&
    \multirow{1}{*}{303} &\multirow{1}{*}{1.16}\\
    \hline
    % HasLastSubevent & The last thing you do when you [[SUBJ]] is [[OBJ]] . & 297 & 1.42\\ 
    Entails & `[[SUBJ]] entails [[OBJ]] .'; `[[SUBJ]] brings about [[OBJ]] .'; `[[SUBJ]] leads to [[OBJ]] .' & 284 & 1.06\\ 
    \hline
    % HasFirstSubevent & The first thing you do when you [[SUBJ]] is [[OBJ]] . & 274 & 1.39\\ 
    Desires & `[[SUBJ]] wants [[OBJ]] .'; `[[SUBJ]] likes to [[OBJ]] .'; `[[SUBJ]] desires to [[OBJ]] .' & 196 & 2.62\\ 
    \hline
    NotHasProperty & `[[SUBJ]] can not be [[OBJ]] .'; `[[SUBJ]] does not have [[OBJ]] .'; `[[SUBJ]] is not [[OBJ]] .' & 158 & 1.08 \\ 
    \hline
    CreatedBy & `[[SUBJ]] is created by [[OBJ]] .'; `[[SUBJ]] is developed by [[OBJ]] .'; `[[SUBJ]] is invented by [[OBJ]] .' & 102 & 1.22\\
    \hline
    NotDesires & `[[SUBJ]] does not want [[OBJ]] .'; `[[SUBJ]] does not want to [[OBJ]] .'; `[[SUBJ]] dislikes to [[OBJ]] .' & 70 & 4.4 \\
    \hline
    DefinedAs & [[SUBJ]] can be defined as [[OBJ]] ; `[[SUBJ]] is the [[OBJ]] .'; `The definition of [[SUBJ]] is [[OBJ]] .' & 63 & 1.14\\ 
    \hline
    NotCapableOf & `[[SUBJ]] cannot [[OBJ]] .'; `[[SUBJ]] does not [[OBJ]] .'; `[[SUBJ]] may not [[OBJ]] .' & 42 & 1.1 \\ 
    \hline
    LocatedNear & `[[SUBJ]] is typically near [[OBJ]] .'; `[[SUBJ]] is typically close by [[OBJ]] .'; `[[SUBJ]] is typically nearby [[OBJ]] .'' & 32 & 1.03 \\ 
    \hline
    EtymologicallyDerivedFrom & `[[SUBJ]] is etymologically derived from [[OBJ]] .'; `[[SUBJ]] is etymologically stemmed from [[OBJ]] .'; `[[SUBJ]] is etymologically originated from [[OBJ]] .' & 23 & 1\\ 
    % SymbolOf & [[SUBJ]] is an symbol of [[OBJ]] . & 3 & 1 \\ 

    \hline
  \end{tabular}
  %\end{center}
%  \end{adjustwidth}
\end{table*}

\newpage
% \section{Detail Information of the Experimental MNLMs}
% \label{apx:detail_information_of_the_experimental_mnlms}

\section{Details on the Grammar Transformation}
\label{apx:details_on_the_grammar_transformation}
Since injecting a subject and a object directly into a template can induce a grammatically problematic sentence, we apply ensuing regulations yielding candidate sentences. Then, by using a language model, we choose the most grammatical sentence among the grammatically diversified sentences.

\begin{enumerate}
    \item If the first word is a adjective or noun, or if the first word is a verb and the second word is a noun or adjective, prepend a definite or indefinite article. (i.e. opera -> an opera, the opera)
    \item If the first word is infinitive verb, then convert it to a gerund. (i.e. make -> making)
    \item If first word is number, then pluralize the next word. (i.e. two leg -> two legs)
    \item Adding variations of plural verb if it is possible, to keep subject-verb concord. (i.e. skywards is -> skywards are)
\end{enumerate}

In order to analyze the part-of-speech tags, python library spaCy (\url{https://spacy.io}) is applied. Furthermore, pattern (\url{https://github.com/clips/pattern/wiki}) is used for conjugation and pluralization. Finally, GPT model (\url{https://huggingface.co/transformers/model_doc/gpt.html}) is utilized to figure out which candidate sentence is most likely to be grammatically.
\newpage

\section{Qualitative Results of Knowledge Probing Test}
\label{apx:qualitative_results}
\begin{figure*}[!ht]
  \adjustbox{minipage=2em,raise=-\height}{\subcaption{} \label{fig:L}}%
  \raisebox{-\height}{\includegraphics[width=.95\linewidth]{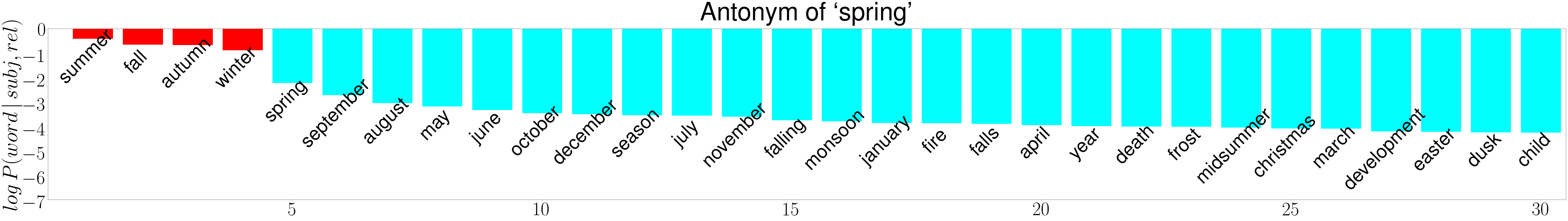}}
  \newline
  \adjustbox{minipage=2em,raise=-\height}{\subcaption{} \label{fig:U}}%
  \raisebox{-\height}{\includegraphics[width=.95\linewidth]{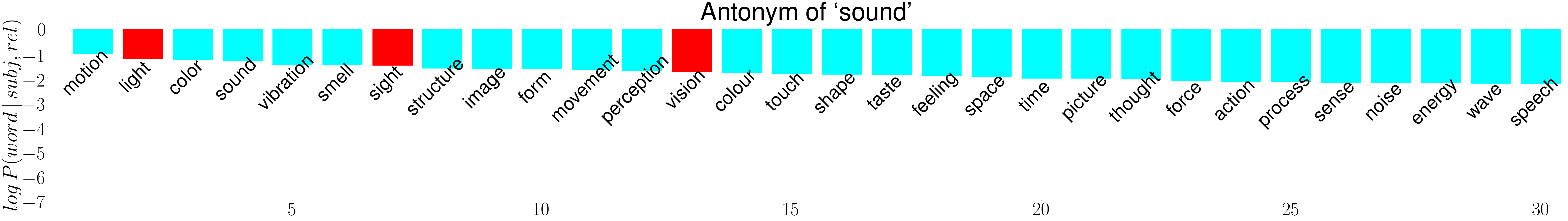}}
  \newline
  \adjustbox{minipage=2em,raise=-\height}{\subcaption{} \label{fig:H}}%
  \raisebox{-\height}{\includegraphics[width=.95\linewidth]{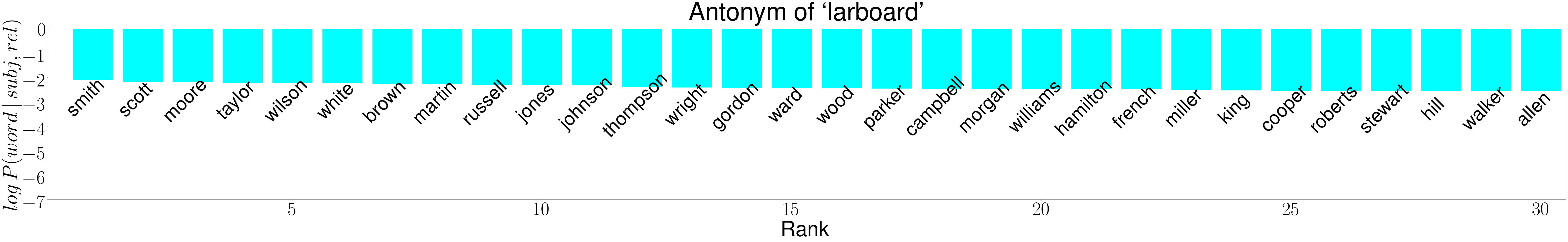}}
  
  \caption{\textbf{Representative probabilistic distributions of the knowledge probing test results on the BERT$_{base}$ model}. (a), (b) and (c) respectively show results of `\textit{antonym} of \textit{sound}', `\textit{antonym} of \textit{spring}' and `\textit{antonym} of \textit{larboard}'. The y-axis indicates log$_{10}$ probability and the x-axis denotes the ranking of the words. Correct answers are marked in red.}
  \label{fig:blank_filling_qulitative}
\end{figure*}

Figure~\ref{fig:blank_filling_qulitative} visualizes three sampled examples of the knowledge probing test results. We demonstrate predicted conditional log likelihood results of Equation~(\ref{DISTRIBUTION}) of the examples.

\begin{equation} \label{DISTRIBUTION} 
\hspace*{3.0cm} 
\text{Conditional\,log\,likelihood\,=\,}logP\text{(word|subject,relation)}
\end{equation}

First, we can see that the `\textit{antonym} of \textit{spring}' case 
not only are the correct answers ranked high, but they show a significantly higher likelihood compared to non-answers. We suppose that the semantic triple of this case is relatively frequently trained on some words as the model is significantly more confident in some words than others.
On the other hand, the `\textit{antonym} of \textit{sound}' case answers are predicted, but there is a non-answer that has a higher rank than the answers and there is little difference between answers and non-answers likelihoods. Finally, in the `\textit{antonym} of \textit{larboard}' case, all answers are failed to be predicted and all candidates have similar likelihoods. We conjecture that the relations are not trained enough in the training as the model is not as confident on any of its predictions. 
% The correct answers were predicted, but the rank was not relatively high.

% Given these empirical observations, we suppose that semantic triples that show `L'-shaped graphs are relatively frequently trained on some words as the model is significantly more confident in some words than others. If the semantic triples are properly trained on the model, the words with high probabilities will be the answers as shown in Fig.~\ref{fig:L}. In contrast, we conjecture that the relations are not trained enough in the training when the results show `--'-shaped graphs as the model is not as confident on any of its predictions. 

% % \section{Additional Examples of Representative Probabilistic Distributions of the Knowledge Probing Test Result}
% \label{apx:additional_probabilistic_distribution}

\newpage
\section{Quantitative Analysis for Probabilistic Distributions}
\label{apx:quantitative_analysis_for_probabilistic_distributions}

\begin{table*}[!ht]
 \centering
 \small
 \caption{Results of the BERT $hits@K$ metric for each relation in ConceptNet. }
 \label{tab:results_on_each_relation}
    \begin{tabular}{@{}c|ccc|ccc@{}}
    \hline
    \multicolumn{1}{c|}{\multirow{3}{*}{\textbf{Relations}}} & \multicolumn{6}{c}{$hits@K$} \\ \cline{2-7} 
    \multicolumn{1}{c|}{} & \multicolumn{3}{c|}{BERT$_{base}$} & \multicolumn{3}{c}{BERT$_{large}$} \\ \cline{2-7} &
    1 & 10 &  \multicolumn{1}{c|}{100} & 1 & 10 &  \multicolumn{1}{c}{100}  \\ \hline\hline
RelatedTo & 3.20$\pm$0.03 & 9.49$\pm$0.05 & 22.68$\pm$0.07 & 2.95$\pm$0.03 & 9.15$\pm$0.05 & 21.89$\pm$0.07\\
HasContext & 7.05$\pm$0.09 & 24.33$\pm$0.15 & 48.07$\pm$0.17 & 7.11$\pm$0.09 & 24.17$\pm$0.15 & 47.68$\pm$0.17\\
IsA & 22.90$\pm$0.16 & 42.06$\pm$0.18 & 62.76$\pm$0.18 & 22.32$\pm$0.16 & 42.51$\pm$0.19 & 63.06$\pm$0.18\\
DerivedFrom & 13.16$\pm$0.15 & 32.03$\pm$0.20 & 52.81$\pm$0.22 & 9.27$\pm$0.13 & 27.74$\pm$0.20 & 49.74$\pm$0.22\\
Synonym & 9.58$\pm$0.20 & 27.81$\pm$0.30 & 47.95$\pm$0.33 & 8.46$\pm$0.19 & 27.69$\pm$0.30 & 48.62$\pm$0.33\\
FormOf & 0.26$\pm$0.04 & 15.73$\pm$0.26 & 31.19$\pm$0.32 & 0.98$\pm$0.07 & 24.05$\pm$0.30 & 36.07$\pm$0.34\\
EtymologicallyRelatedTo & 6.91$\pm$0.28 & 14.98$\pm$0.41 & 26.00$\pm$0.51 & 7.67$\pm$0.30 & 18.43$\pm$0.44 & 32.07$\pm$0.54\\
SimilarTo & 1.03$\pm$0.12 & 4.62$\pm$0.24 & 13.11$\pm$0.39 & 1.94$\pm$0.16 & 8.86$\pm$0.33 & 21.33$\pm$0.47\\
AtLocation & 0.03$\pm$0.02 & 1.21$\pm$0.13 & 13.09$\pm$0.38 & 0.00$\pm$0.00 & 0.59$\pm$0.09 & 9.30$\pm$0.33\\
MannerOf & 2.05$\pm$0.18 & 9.68$\pm$0.36 & 38.37$\pm$0.57 & 2.04$\pm$0.18 & 10.08$\pm$0.37 & 38.00$\pm$0.58\\
PartOf & 0.00$\pm$0.00 & 6.19$\pm$0.50 & 26.56$\pm$0.90 & 0.00$\pm$0.00 & 7.00$\pm$0.53 & 26.30$\pm$0.90\\
Antonym & 7.49$\pm$0.41 & 21.38$\pm$0.63 & 38.39$\pm$0.76 & 10.43$\pm$0.48 & 25.61$\pm$0.68 & 43.11$\pm$0.78\\
HasProperty & 4.81$\pm$0.40 & 18.56$\pm$0.73 & 47.18$\pm$0.95 & 7.01$\pm$0.48 & 25.44$\pm$0.82 & 52.75$\pm$0.94\\
UsedFor & 5.51$\pm$0.41 & 21.57$\pm$0.72 & 48.30$\pm$0.86 & 6.42$\pm$0.46 & 20.44$\pm$0.71 & 48.43$\pm$0.87\\
DistinctFrom & 5.69$\pm$0.58 & 23.69$\pm$1.11 & 47.43$\pm$1.30 & 9.48$\pm$0.76 & 28.55$\pm$1.17 & 53.26$\pm$1.29\\
HasPrerequisite & 0.13$\pm$0.10 & 0.28$\pm$0.14 & 8.01$\pm$0.66 & 0.13$\pm$0.10 & 0.41$\pm$0.15 & 9.16$\pm$0.67\\
HasSubevent & 5.21$\pm$0.53 & 17.90$\pm$0.89 & 47.49$\pm$1.14 & 6.32$\pm$0.58 & 20.55$\pm$0.92 & 50.75$\pm$1.12\\
Causes & 0.12$\pm$0.11 & 0.77$\pm$0.26 & 8.47$\pm$0.80 & 0.53$\pm$0.23 & 0.99$\pm$0.30 & 5.08$\pm$0.62\\
HasA & 5.93$\pm$0.79 & 17.27$\pm$1.28 & 44.98$\pm$1.67 & 7.81$\pm$0.91 & 22.88$\pm$1.42 & 52.11$\pm$1.68\\
InstanceOf & 0.17$\pm$0.17 & 7.93$\pm$1.12 & 27.10$\pm$1.84 & 0.00$\pm$0.00 & 7.93$\pm$1.12 & 34.77$\pm$1.97\\
CapableOf & 9.66$\pm$1.12 & 26.69$\pm$1.67 & 56.86$\pm$1.87 & 11.96$\pm$1.23 & 30.66$\pm$1.74 & 57.30$\pm$1.86\\
MotivatedByGoal & 2.62$\pm$0.51 & 6.77$\pm$0.74 & 24.93$\pm$1.32 & 1.40$\pm$0.37 & 5.92$\pm$0.71 & 25.29$\pm$1.31\\
MadeOf & 18.32$\pm$2.20 & 53.01$\pm$2.85 & 74.23$\pm$2.49 & 19.74$\pm$2.27 & 51.02$\pm$2.85 & 74.57$\pm$2.49\\
Entails & 0.71$\pm$0.50 & 3.90$\pm$1.15 & 16.08$\pm$2.16 & 1.77$\pm$0.79 & 3.90$\pm$1.15 & 13.77$\pm$2.04\\
Desires & 7.34$\pm$1.85 & 18.29$\pm$2.68 & 33.43$\pm$3.30 & 7.79$\pm$1.91 & 18.94$\pm$2.71 & 34.01$\pm$3.31\\
NotHasProperty & 2.55$\pm$1.26 & 16.24$\pm$2.94 & 35.88$\pm$3.77 & 6.37$\pm$1.96 & 23.46$\pm$3.36 & 48.83$\pm$3.94\\
CreatedBy & 6.38$\pm$2.42 & 20.04$\pm$3.88 & 47.25$\pm$5.05 & 4.79$\pm$2.15 & 22.16$\pm$4.18 & 47.34$\pm$4.99\\
DefinedAs & 12.22$\pm$4.19 & 25.28$\pm$5.54 & 50.56$\pm$6.36 & 16.67$\pm$4.85 & 37.50$\pm$6.15 & 55.00$\pm$6.38\\
NotDesires & 9.60$\pm$3.46 & 13.95$\pm$4.11 & 23.60$\pm$5.00 & 9.60$\pm$3.46 & 13.95$\pm$4.11 & 25.05$\pm$5.11\\
NotCapableOf & 23.17$\pm$6.56 & 48.78$\pm$7.90 & 73.78$\pm$6.87 & 28.05$\pm$6.99 & 46.34$\pm$7.88 & 79.88$\pm$6.25\\
LocatedNear & 0.00$\pm$0.00 & 18.75$\pm$7.01 & 46.88$\pm$8.96 & 3.12$\pm$3.12 & 12.50$\pm$5.94 & 37.50$\pm$8.70\\
EtymologicallyDerivedFrom & 8.70$\pm$6.01 & 13.04$\pm$7.18 & 13.04$\pm$7.18 & 13.04$\pm$7.18 & 21.74$\pm$8.79 & 26.09$\pm$9.36\\
\hline
    \end{tabular}
    {\footnotesize
\begin{tablenotes}
\item[] Data presented as mean $\pm$ standard error of mean
\end{tablenotes}
}

\end{table*}

\begin{table*}[!p]
 \centering
 \small
 \caption{Results of the ALBERT1 $hits@K$ metric for each relation in ConceptNet. }
 \label{tab:results_on_each_relation}
 \rotatebox{270}{
    \begin{tabular}{@{}c|ccc|ccc|ccc@{}}
    \hline
    \multicolumn{1}{c|}{\multirow{3}{*}{\textbf{Relations}}} & \multicolumn{9}{c}{$hits@K$} \\ \cline{2-10} 
    \multicolumn{1}{c|}{} & \multicolumn{3}{c|}{ALBERT1$_{base}$} & \multicolumn{3}{c|}{ALBERT1$_{large}$} & \multicolumn{3}{c}{ALBERT1$_{xlarge}$} \\ \cline{2-10} &
    1 & 10 &  \multicolumn{1}{c|}{100} & 1 & 10 &  \multicolumn{1}{c|}{100} & 1 & 10 &  \multicolumn{1}{c}{100} \\ \hline\hline
RelatedTo & 4.65$\pm$0.04 & 11.29$\pm$0.06 & 24.52$\pm$0.08 & 2.56$\pm$0.03 & 10.13$\pm$0.05 & 24.59$\pm$0.08 & 1.83$\pm$0.02 & 6.12$\pm$0.04 & 19.03$\pm$0.07\\
HasContext & 9.32$\pm$0.10 & 25.25$\pm$0.15 & 47.35$\pm$0.17 & 1.49$\pm$0.04 & 9.51$\pm$0.10 & 27.83$\pm$0.15 & 3.13$\pm$0.06 & 8.69$\pm$0.10 & 20.75$\pm$0.14\\
IsA & 22.18$\pm$0.16 & 37.45$\pm$0.18 & 57.98$\pm$0.18 & 22.22$\pm$0.16 & 40.43$\pm$0.18 & 60.20$\pm$0.18 & 19.44$\pm$0.15 & 35.94$\pm$0.18 & 54.93$\pm$0.18\\
DerivedFrom & 13.70$\pm$0.15 & 32.91$\pm$0.21 & 57.67$\pm$0.22 & 5.57$\pm$0.10 & 22.41$\pm$0.18 & 42.11$\pm$0.22 & 1.68$\pm$0.06 & 4.87$\pm$0.09 & 12.95$\pm$0.15\\
Synonym & 11.78$\pm$0.22 & 23.47$\pm$0.28 & 38.65$\pm$0.33 & 12.04$\pm$0.22 & 25.25$\pm$0.29 & 40.78$\pm$0.33 & 7.07$\pm$0.17 & 22.22$\pm$0.28 & 40.68$\pm$0.33\\
FormOf & 0.14$\pm$0.03 & 15.52$\pm$0.26 & 33.06$\pm$0.33 & 1.27$\pm$0.08 & 10.31$\pm$0.21 & 25.13$\pm$0.30 & 0.66$\pm$0.06 & 15.47$\pm$0.26 & 28.60$\pm$0.32\\
EtymologicallyRelatedTo & 7.32$\pm$0.29 & 15.82$\pm$0.42 & 27.20$\pm$0.52 & 5.44$\pm$0.25 & 14.71$\pm$0.40 & 27.83$\pm$0.52 & 5.58$\pm$0.26 & 13.04$\pm$0.38 & 25.07$\pm$0.50\\
SimilarTo & 0.73$\pm$0.10 & 5.50$\pm$0.26 & 16.43$\pm$0.42 & 0.44$\pm$0.08 & 4.08$\pm$0.23 & 14.82$\pm$0.41 & 0.41$\pm$0.07 & 3.27$\pm$0.20 & 15.89$\pm$0.41\\
AtLocation & 0.01$\pm$0.01 & 1.21$\pm$0.12 & 16.04$\pm$0.41 & 0.05$\pm$0.03 & 1.86$\pm$0.16 & 16.59$\pm$0.42 & 0.02$\pm$0.02 & 2.87$\pm$0.19 & 16.76$\pm$0.42\\
MannerOf & 2.28$\pm$0.18 & 11.46$\pm$0.38 & 39.09$\pm$0.57 & 4.05$\pm$0.24 & 15.38$\pm$0.43 & 42.01$\pm$0.58 & 1.05$\pm$0.12 & 6.67$\pm$0.29 & 32.23$\pm$0.55\\
PartOf & 0.00$\pm$0.00 & 4.07$\pm$0.41 & 22.78$\pm$0.86 & 0.00$\pm$0.00 & 5.82$\pm$0.49 & 23.34$\pm$0.87 & 0.00$\pm$0.00 & 7.13$\pm$0.53 & 28.53$\pm$0.92\\
Antonym & 1.42$\pm$0.19 & 13.42$\pm$0.53 & 30.14$\pm$0.71 & 2.30$\pm$0.23 & 12.42$\pm$0.51 & 27.60$\pm$0.69 & 6.19$\pm$0.37 & 16.38$\pm$0.58 & 33.17$\pm$0.73\\
HasProperty & 0.81$\pm$0.17 & 6.07$\pm$0.45 & 24.05$\pm$0.81 & 1.31$\pm$0.21 & 7.97$\pm$0.52 & 26.88$\pm$0.84 & 1.83$\pm$0.25 & 8.62$\pm$0.53 & 23.60$\pm$0.81\\
UsedFor & 2.48$\pm$0.28 & 11.35$\pm$0.58 & 34.85$\pm$0.84 & 3.13$\pm$0.32 & 13.92$\pm$0.63 & 36.26$\pm$0.86 & 4.04$\pm$0.37 & 14.60$\pm$0.65 & 35.66$\pm$0.86\\
DistinctFrom & 0.69$\pm$0.21 & 16.56$\pm$0.96 & 43.51$\pm$1.28 & 1.24$\pm$0.29 & 15.74$\pm$0.95 & 40.61$\pm$1.27 & 4.18$\pm$0.51 & 18.80$\pm$1.02 & 40.65$\pm$1.29\\
HasPrerequisite & 0.00$\pm$0.00 & 1.23$\pm$0.29 & 9.89$\pm$0.75 & 0.05$\pm$0.05 & 0.41$\pm$0.15 & 16.20$\pm$0.88 & 0.14$\pm$0.10 & 0.35$\pm$0.17 & 3.38$\pm$0.41\\
HasSubevent & 0.00$\pm$0.00 & 0.15$\pm$0.10 & 5.90$\pm$0.52 & 0.00$\pm$0.00 & 2.18$\pm$0.38 & 19.58$\pm$0.94 & 0.03$\pm$0.03 & 1.50$\pm$0.31 & 13.68$\pm$0.82\\
Causes & 0.00$\pm$0.00 & 0.04$\pm$0.02 & 1.95$\pm$0.40 & 0.00$\pm$0.00 & 0.29$\pm$0.15 & 4.72$\pm$0.58 & 0.11$\pm$0.11 & 0.27$\pm$0.15 & 2.62$\pm$0.44\\
HasA & 1.34$\pm$0.40 & 3.98$\pm$0.67 & 13.36$\pm$1.15 & 1.94$\pm$0.48 & 7.44$\pm$0.88 & 26.51$\pm$1.49 & 3.68$\pm$0.63 & 13.47$\pm$1.17 & 30.18$\pm$1.56\\
InstanceOf & 0.69$\pm$0.34 & 10.34$\pm$1.25 & 21.95$\pm$1.71 & 5.86$\pm$0.97 & 15.52$\pm$1.49 & 32.99$\pm$1.95 & 3.02$\pm$0.71 & 12.82$\pm$1.38 & 20.03$\pm$1.65\\
CapableOf & 3.24$\pm$0.68 & 14.88$\pm$1.36 & 43.98$\pm$1.87 & 1.59$\pm$0.48 & 11.95$\pm$1.25 & 37.13$\pm$1.83 & 1.91$\pm$0.53 & 9.46$\pm$1.12 & 35.12$\pm$1.81\\
MotivatedByGoal & 0.00$\pm$0.00 & 4.81$\pm$0.67 & 20.80$\pm$1.21 & 1.62$\pm$0.39 & 6.50$\pm$0.74 & 24.20$\pm$1.29 & 0.68$\pm$0.28 & 3.00$\pm$0.50 & 14.76$\pm$1.07\\
MadeOf & 11.49$\pm$1.80 & 43.69$\pm$2.84 & 73.27$\pm$2.51 & 13.54$\pm$1.96 & 41.07$\pm$2.80 & 70.14$\pm$2.62 & 15.19$\pm$2.05 & 50.23$\pm$2.84 & 74.57$\pm$2.49\\
Entails & 0.35$\pm$0.35 & 2.96$\pm$1.00 & 13.00$\pm$1.95 & 0.71$\pm$0.50 & 2.30$\pm$0.88 & 11.17$\pm$1.85 & 0.71$\pm$0.50 & 2.84$\pm$0.99 & 9.40$\pm$1.71\\
Desires & 2.05$\pm$1.02 & 9.05$\pm$2.00 & 26.54$\pm$3.07 & 2.62$\pm$1.14 & 11.11$\pm$2.19 & 26.67$\pm$3.07 & 2.67$\pm$1.08 & 11.87$\pm$2.25 & 24.61$\pm$3.02\\
NotHasProperty & 1.27$\pm$0.90 & 12.10$\pm$2.61 & 35.99$\pm$3.80 & 2.55$\pm$1.26 & 10.19$\pm$2.42 & 30.57$\pm$3.63 & 3.18$\pm$1.41 & 12.10$\pm$2.61 & 30.25$\pm$3.66\\
CreatedBy & 0.00$\pm$0.00 & 13.62$\pm$3.42 & 35.66$\pm$4.79 & 0.00$\pm$0.00 & 13.80$\pm$3.52 & 32.35$\pm$4.71 & 2.69$\pm$1.60 & 12.19$\pm$3.28 & 27.06$\pm$4.51\\
DefinedAs & 5.00$\pm$2.84 & 21.39$\pm$5.21 & 36.11$\pm$6.20 & 1.67$\pm$1.67 & 15.83$\pm$4.47 & 31.11$\pm$5.87 & 6.67$\pm$3.25 & 22.50$\pm$5.37 & 28.33$\pm$5.87\\
NotDesires & 0.00$\pm$0.00 & 11.05$\pm$3.70 & 18.50$\pm$4.61 & 5.25$\pm$2.56 & 11.05$\pm$3.70 & 23.77$\pm$5.00 & 1.45$\pm$1.45 & 11.78$\pm$3.88 & 20.14$\pm$4.75\\
NotCapableOf & 7.32$\pm$4.12 & 25.00$\pm$6.76 & 60.37$\pm$7.66 & 10.98$\pm$4.79 & 17.07$\pm$5.95 & 48.78$\pm$7.90 & 10.98$\pm$4.79 & 26.83$\pm$7.01 & 48.78$\pm$7.90\\
LocatedNear & 0.00$\pm$0.00 & 9.38$\pm$5.24 & 29.69$\pm$8.05 & 0.00$\pm$0.00 & 0.00$\pm$0.00 & 35.94$\pm$8.47 & 0.00$\pm$0.00 & 3.12$\pm$3.12 & 15.62$\pm$6.52\\
EtymologicallyDerivedFrom & 4.35$\pm$4.35 & 17.39$\pm$8.08 & 26.09$\pm$9.36 & 4.35$\pm$4.35 & 4.35$\pm$4.35 & 26.09$\pm$9.36 & 4.35$\pm$4.35 & 4.35$\pm$4.35 & 13.04$\pm$7.18\\
\hline
    \end{tabular}
    }
    {\footnotesize
\begin{tablenotes}
\item[] Data presented as mean $\pm$ standard error of mean
\end{tablenotes}
}

\end{table*}

\begin{table*}[!p]
 \centering
 \small
 \caption{Results of the ALBERT2 $hits@K$ metric for each relation in ConceptNet. }
 \label{tab:results_on_each_relation}
 \rotatebox{270}{
    \begin{tabular}{@{}c|ccc|ccc|ccc@{}}
    \hline
    \multicolumn{1}{c|}{\multirow{3}{*}{\textbf{Relations}}} & \multicolumn{9}{c}{$hits@K$} \\ \cline{2-10} 
    \multicolumn{1}{c|}{} & \multicolumn{3}{c|}{ALBERT2$_{base}$} & \multicolumn{3}{c|}{ALBERT2$_{large}$} & \multicolumn{3}{c}{ALBERT2$_{xlarge}$} \\ \cline{2-10} &
    1 & 10 &  \multicolumn{1}{c|}{100} & 1 & 10 &  \multicolumn{1}{c|}{100} & 1 & 10 &  \multicolumn{1}{c}{100} \\ \hline\hline
RelatedTo & 3.18$\pm$0.03 & 10.31$\pm$0.05 & 23.21$\pm$0.07 & 6.31$\pm$0.04 & 14.09$\pm$0.06 & 27.33$\pm$0.08 & 5.64$\pm$0.04 & 14.78$\pm$0.06 & 28.44$\pm$0.08\\
HasContext & 2.78$\pm$0.06 & 14.75$\pm$0.12 & 41.11$\pm$0.17 & 4.26$\pm$0.07 & 17.22$\pm$0.13 & 40.99$\pm$0.17 & 3.07$\pm$0.06 & 12.78$\pm$0.12 & 36.00$\pm$0.16\\
IsA & 17.00$\pm$0.14 & 35.16$\pm$0.18 & 55.08$\pm$0.18 & 19.06$\pm$0.15 & 38.85$\pm$0.18 & 60.36$\pm$0.18 & 20.08$\pm$0.15 & 40.30$\pm$0.18 & 60.45$\pm$0.18\\
DerivedFrom & 13.52$\pm$0.15 & 31.26$\pm$0.20 & 53.90$\pm$0.22 & 16.75$\pm$0.16 & 38.13$\pm$0.21 & 62.08$\pm$0.21 & 24.59$\pm$0.19 & 43.48$\pm$0.22 & 62.24$\pm$0.21\\
Synonym & 9.84$\pm$0.20 & 23.45$\pm$0.28 & 40.96$\pm$0.33 & 12.13$\pm$0.22 & 28.77$\pm$0.30 & 46.78$\pm$0.33 & 9.12$\pm$0.19 & 27.52$\pm$0.30 & 47.78$\pm$0.33\\
FormOf & 1.60$\pm$0.09 & 21.98$\pm$0.29 & 40.94$\pm$0.34 & 2.15$\pm$0.10 & 35.20$\pm$0.33 & 59.06$\pm$0.32 & 2.95$\pm$0.12 & 41.57$\pm$0.34 & 57.45$\pm$0.33\\
EtymologicallyRelatedTo & 5.86$\pm$0.26 & 12.76$\pm$0.38 & 23.42$\pm$0.49 & 9.92$\pm$0.33 & 19.89$\pm$0.46 & 31.49$\pm$0.54 & 4.80$\pm$0.23 & 14.20$\pm$0.40 & 26.69$\pm$0.51\\
SimilarTo & 0.87$\pm$0.11 & 7.45$\pm$0.30 & 21.32$\pm$0.46 & 1.48$\pm$0.14 & 9.12$\pm$0.33 & 22.62$\pm$0.48 & 2.60$\pm$0.18 & 12.98$\pm$0.38 & 31.78$\pm$0.53\\
AtLocation & 0.01$\pm$0.01 & 1.84$\pm$0.15 & 16.04$\pm$0.41 & 0.01$\pm$0.01 & 3.60$\pm$0.21 & 19.95$\pm$0.45 & 0.02$\pm$0.02 & 3.03$\pm$0.19 & 16.50$\pm$0.42\\
MannerOf & 1.04$\pm$0.12 & 6.98$\pm$0.30 & 34.97$\pm$0.56 & 4.46$\pm$0.26 & 17.26$\pm$0.45 & 47.28$\pm$0.59 & 2.27$\pm$0.18 & 13.33$\pm$0.41 & 38.74$\pm$0.57\\
PartOf & 0.00$\pm$0.00 & 7.89$\pm$0.56 & 28.19$\pm$0.92 & 0.22$\pm$0.10 & 13.16$\pm$0.70 & 40.33$\pm$1.00 & 0.00$\pm$0.00 & 10.92$\pm$0.65 & 34.44$\pm$0.97\\
Antonym & 5.52$\pm$0.36 & 19.48$\pm$0.62 & 38.11$\pm$0.75 & 9.50$\pm$0.47 & 28.22$\pm$0.70 & 50.02$\pm$0.77 & 13.86$\pm$0.55 & 34.19$\pm$0.74 & 55.43$\pm$0.76\\
HasProperty & 3.47$\pm$0.34 & 13.91$\pm$0.65 & 39.10$\pm$0.93 & 4.04$\pm$0.37 & 16.14$\pm$0.70 & 41.79$\pm$0.94 & 7.62$\pm$0.50 & 23.50$\pm$0.80 & 52.10$\pm$0.95\\
UsedFor & 3.71$\pm$0.35 & 16.27$\pm$0.67 & 42.65$\pm$0.88 & 6.89$\pm$0.47 & 21.89$\pm$0.74 & 48.65$\pm$0.87 & 6.20$\pm$0.44 & 22.30$\pm$0.76 & 49.05$\pm$0.90\\
DistinctFrom & 4.75$\pm$0.55 & 25.92$\pm$1.14 & 48.26$\pm$1.30 & 7.61$\pm$0.68 & 32.65$\pm$1.21 & 58.80$\pm$1.27 & 8.59$\pm$0.73 & 33.27$\pm$1.23 & 56.90$\pm$1.29\\
HasPrerequisite & 0.66$\pm$0.16 & 4.10$\pm$0.45 & 19.58$\pm$0.90 & 0.16$\pm$0.08 & 3.51$\pm$0.40 & 15.99$\pm$0.82 & 1.12$\pm$0.22 & 6.22$\pm$0.57 & 22.79$\pm$0.98\\
HasSubevent & 2.96$\pm$0.43 & 15.58$\pm$0.85 & 39.63$\pm$1.11 & 4.45$\pm$0.52 & 16.34$\pm$0.88 & 42.73$\pm$1.14 & 5.45$\pm$0.55 & 20.84$\pm$0.95 & 47.68$\pm$1.14\\
Causes & 0.29$\pm$0.16 & 1.71$\pm$0.39 & 12.56$\pm$0.92 & 0.23$\pm$0.15 & 1.50$\pm$0.37 & 8.38$\pm$0.80 & 1.67$\pm$0.35 & 11.63$\pm$0.85 & 39.18$\pm$1.27\\
HasA & 2.45$\pm$0.54 & 7.88$\pm$0.93 & 25.52$\pm$1.48 & 2.59$\pm$0.55 & 7.51$\pm$0.91 & 27.49$\pm$1.52 & 5.24$\pm$0.76 & 16.48$\pm$1.27 & 37.64$\pm$1.64\\
InstanceOf & 0.00$\pm$0.00 & 8.02$\pm$1.12 & 23.10$\pm$1.74 & 0.34$\pm$0.24 & 12.67$\pm$1.37 & 19.89$\pm$1.65 & 0.17$\pm$0.17 & 3.79$\pm$0.79 & 22.73$\pm$1.73\\
CapableOf & 5.45$\pm$0.88 & 24.64$\pm$1.63 & 52.47$\pm$1.87 & 2.33$\pm$0.59 & 13.32$\pm$1.30 & 41.30$\pm$1.86 & 11.25$\pm$1.20 & 29.62$\pm$1.73 & 56.21$\pm$1.86\\
MotivatedByGoal & 0.46$\pm$0.21 & 3.76$\pm$0.58 & 19.89$\pm$1.19 & 0.94$\pm$0.30 & 6.36$\pm$0.73 & 22.56$\pm$1.29 & 1.17$\pm$0.32 & 6.48$\pm$0.73 & 24.70$\pm$1.29\\
MadeOf & 13.42$\pm$1.96 & 43.74$\pm$2.81 & 73.72$\pm$2.52 & 19.34$\pm$2.25 & 49.77$\pm$2.82 & 77.65$\pm$2.38 & 23.15$\pm$2.40 & 55.63$\pm$2.81 & 78.56$\pm$2.35\\
Entails & 1.06$\pm$0.61 & 4.61$\pm$1.25 & 20.04$\pm$2.35 & 1.06$\pm$0.61 & 7.80$\pm$1.58 & 26.65$\pm$2.61 & 0.35$\pm$0.35 & 2.84$\pm$0.99 & 17.14$\pm$2.23\\
Desires & 7.08$\pm$1.77 & 15.55$\pm$2.51 & 31.59$\pm$3.25 & 7.25$\pm$1.81 & 19.82$\pm$2.80 & 33.59$\pm$3.31 & 9.98$\pm$2.10 & 19.97$\pm$2.80 & 35.18$\pm$3.34\\
NotHasProperty & 1.91$\pm$1.10 & 10.83$\pm$2.49 & 29.19$\pm$3.61 & 3.18$\pm$1.41 & 19.43$\pm$3.15 & 33.97$\pm$3.73 & 4.46$\pm$1.65 & 21.34$\pm$3.26 & 36.20$\pm$3.80\\
CreatedBy & 6.45$\pm$2.56 & 22.58$\pm$4.22 & 52.87$\pm$5.04 & 5.38$\pm$2.35 & 22.76$\pm$4.24 & 52.69$\pm$4.92 & 3.76$\pm$1.91 & 13.44$\pm$3.43 & 36.38$\pm$4.84\\
DefinedAs & 7.50$\pm$3.32 & 31.39$\pm$5.93 & 64.44$\pm$6.18 & 8.33$\pm$3.60 & 17.50$\pm$4.87 & 42.22$\pm$6.38 & 17.50$\pm$4.87 & 32.22$\pm$6.03 & 53.89$\pm$6.44\\
NotDesires & 3.80$\pm$2.15 & 13.95$\pm$4.11 & 23.05$\pm$4.89 & 6.52$\pm$2.90 & 15.40$\pm$4.29 & 27.22$\pm$5.30 & 1.45$\pm$1.45 & 16.85$\pm$4.45 & 26.96$\pm$5.19\\
NotCapableOf & 15.85$\pm$5.64 & 37.20$\pm$7.57 & 76.83$\pm$6.56 & 12.20$\pm$5.17 & 31.71$\pm$7.36 & 67.68$\pm$7.32 & 23.17$\pm$6.56 & 43.90$\pm$7.85 & 68.90$\pm$7.24\\
LocatedNear & 3.12$\pm$3.12 & 14.06$\pm$6.04 & 40.62$\pm$8.82 & 3.12$\pm$3.12 & 10.94$\pm$5.38 & 32.81$\pm$8.28 & 3.12$\pm$3.12 & 10.94$\pm$5.38 & 46.88$\pm$8.96\\
EtymologicallyDerivedFrom & 8.70$\pm$6.01 & 8.70$\pm$6.01 & 26.09$\pm$9.36 & 13.04$\pm$7.18 & 21.74$\pm$8.79 & 30.43$\pm$9.81 & 8.70$\pm$6.01 & 21.74$\pm$8.79 & 30.43$\pm$9.81\\
\hline
    \end{tabular}
    }
    {\footnotesize
\begin{tablenotes}
\item[] Data presented as mean $\pm$ standard error of mean
\end{tablenotes}
}

\end{table*}

\begin{table*}[!p]
 \centering
 \small
 \caption{Results of the RoBERTa $hits@K$ metric for each relation in ConceptNet. }
 \label{tab:results_on_each_relation}
    \begin{tabular}{@{}c|ccc|ccc@{}}
    \hline
    \multicolumn{1}{c|}{\multirow{3}{*}{\textbf{Relations}}} & \multicolumn{6}{c}{$hits@K$} \\ \cline{2-7} 
    \multicolumn{1}{c|}{} & \multicolumn{3}{c|}{RoBERTa$_{base}$} & \multicolumn{3}{c}{RoBERTa$_{large}$} \\ \cline{2-7} &
    1 & 10 &  \multicolumn{1}{c|}{100} & 1 & 10 &  \multicolumn{1}{c}{100}  \\ \hline\hline
RelatedTo & 1.15$\pm$0.02 & 6.28$\pm$0.04 & 19.44$\pm$0.07 & 0.51$\pm$0.01 & 7.64$\pm$0.05 & 23.89$\pm$0.07\\
HasContext & 0.37$\pm$0.02 & 3.41$\pm$0.06 & 20.20$\pm$0.14 & 0.93$\pm$0.03 & 6.38$\pm$0.08 & 26.53$\pm$0.15\\
IsA & 14.70$\pm$0.13 & 34.14$\pm$0.18 & 56.97$\pm$0.18 & 17.77$\pm$0.14 & 40.89$\pm$0.18 & 63.92$\pm$0.18\\
DerivedFrom & 1.21$\pm$0.05 & 10.63$\pm$0.14 & 36.08$\pm$0.21 & 2.74$\pm$0.07 & 23.92$\pm$0.19 & 56.30$\pm$0.22\\
Synonym & 4.10$\pm$0.13 & 15.76$\pm$0.25 & 32.80$\pm$0.31 & 17.50$\pm$0.25 & 37.76$\pm$0.32 & 55.84$\pm$0.33\\
FormOf & 0.22$\pm$0.03 & 2.41$\pm$0.11 & 11.21$\pm$0.22 & 2.46$\pm$0.11 & 9.29$\pm$0.20 & 19.13$\pm$0.28\\
EtymologicallyRelatedTo & 1.92$\pm$0.15 & 8.22$\pm$0.31 & 23.23$\pm$0.48 & 2.76$\pm$0.18 & 16.57$\pm$0.42 & 34.59$\pm$0.55\\
SimilarTo & 0.21$\pm$0.05 & 2.90$\pm$0.19 & 15.53$\pm$0.41 & 0.91$\pm$0.11 & 7.77$\pm$0.31 & 28.80$\pm$0.51\\
AtLocation & 0.00$\pm$0.00 & 0.47$\pm$0.08 & 6.65$\pm$0.28 & 0.02$\pm$0.02 & 0.90$\pm$0.11 & 7.60$\pm$0.30\\
MannerOf & 2.51$\pm$0.19 & 10.25$\pm$0.37 & 38.50$\pm$0.57 & 2.08$\pm$0.17 & 10.41$\pm$0.37 & 38.40$\pm$0.58\\
PartOf & 0.09$\pm$0.06 & 0.90$\pm$0.19 & 11.52$\pm$0.65 & 0.00$\pm$0.00 & 2.00$\pm$0.29 & 15.38$\pm$0.74\\
Antonym & 9.96$\pm$0.47 & 24.49$\pm$0.67 & 44.17$\pm$0.76 & 10.59$\pm$0.49 & 28.16$\pm$0.70 & 49.73$\pm$0.77\\
HasProperty & 5.57$\pm$0.44 & 17.29$\pm$0.72 & 39.92$\pm$0.93 & 6.58$\pm$0.47 & 20.93$\pm$0.77 & 47.42$\pm$0.95\\
UsedFor & 2.74$\pm$0.30 & 11.89$\pm$0.57 & 36.28$\pm$0.85 & 4.46$\pm$0.39 & 17.93$\pm$0.69 & 45.14$\pm$0.87\\
DistinctFrom & 7.59$\pm$0.66 & 23.87$\pm$1.10 & 47.70$\pm$1.29 & 8.37$\pm$0.70 & 28.81$\pm$1.18 & 55.04$\pm$1.29\\
HasPrerequisite & 0.63$\pm$0.17 & 7.17$\pm$0.63 & 23.49$\pm$1.02 & 0.21$\pm$0.09 & 6.50$\pm$0.60 & 29.33$\pm$1.08\\
HasSubevent & 6.66$\pm$0.61 & 23.09$\pm$0.98 & 49.01$\pm$1.09 & 6.87$\pm$0.61 & 23.61$\pm$0.98 & 52.49$\pm$1.13\\
Causes & 0.14$\pm$0.11 & 4.94$\pm$0.64 & 24.94$\pm$1.15 & 0.26$\pm$0.15 & 5.47$\pm$0.66 & 30.88$\pm$1.22\\
HasA & 2.79$\pm$0.57 & 9.98$\pm$1.00 & 28.68$\pm$1.52 & 4.38$\pm$0.69 & 15.41$\pm$1.22 & 38.77$\pm$1.65\\
InstanceOf & 0.69$\pm$0.34 & 7.99$\pm$1.11 & 18.56$\pm$1.61 & 1.21$\pm$0.45 & 9.71$\pm$1.22 & 26.95$\pm$1.83\\
CapableOf & 8.96$\pm$1.09 & 23.53$\pm$1.61 & 48.26$\pm$1.88 & 9.01$\pm$1.09 & 27.28$\pm$1.70 & 56.25$\pm$1.87\\
MotivatedByGoal & 1.01$\pm$0.31 & 3.74$\pm$0.55 & 15.77$\pm$1.10 & 1.57$\pm$0.41 & 5.14$\pm$0.68 & 20.64$\pm$1.23\\
MadeOf & 13.59$\pm$1.96 & 45.11$\pm$2.83 & 76.00$\pm$2.41 & 22.01$\pm$2.34 & 61.04$\pm$2.77 & 83.33$\pm$2.14\\
Entails & 0.35$\pm$0.35 & 1.60$\pm$0.73 & 12.00$\pm$1.89 & 1.06$\pm$0.61 & 3.19$\pm$1.05 & 14.30$\pm$2.07\\
Desires & 7.60$\pm$1.87 & 15.98$\pm$2.58 & 33.47$\pm$3.31 & 7.34$\pm$1.82 & 19.36$\pm$2.76 & 35.39$\pm$3.35\\
NotHasProperty & 3.82$\pm$1.53 & 14.76$\pm$2.81 & 34.18$\pm$3.79 & 3.82$\pm$1.53 & 15.71$\pm$2.90 & 35.77$\pm$3.79\\
CreatedBy & 1.06$\pm$1.06 & 12.77$\pm$3.38 & 32.98$\pm$4.76 & 0.00$\pm$0.00 & 10.64$\pm$3.01 & 42.38$\pm$4.90\\
DefinedAs & 13.33$\pm$4.43 & 31.67$\pm$6.06 & 58.89$\pm$6.36 & 13.33$\pm$4.43 & 31.67$\pm$6.06 & 50.00$\pm$6.51\\
NotDesires & 6.52$\pm$2.90 & 13.95$\pm$4.11 & 26.66$\pm$5.21 & 13.95$\pm$4.11 & 19.75$\pm$4.75 & 30.98$\pm$5.47\\
NotCapableOf & 25.61$\pm$6.79 & 49.39$\pm$7.83 & 82.32$\pm$5.94 & 23.17$\pm$6.56 & 39.02$\pm$7.71 & 78.05$\pm$6.54\\
LocatedNear & 0.00$\pm$0.00 & 9.38$\pm$5.24 & 31.25$\pm$8.32 & 0.00$\pm$0.00 & 12.50$\pm$5.94 & 50.00$\pm$8.98\\
EtymologicallyDerivedFrom & 0.00$\pm$0.00 & 4.35$\pm$4.35 & 17.39$\pm$8.08 & 4.35$\pm$4.35 & 17.39$\pm$8.08 & 26.09$\pm$9.36\\

\hline
    \end{tabular}
    {\footnotesize
\begin{tablenotes}
\item[] Data presented as mean $\pm$ standard error of mean
\end{tablenotes}
}

\end{table*}

\begin{table*}[!p]
 \centering
 \small
 \caption{Results of the GPT1,2, and 3 $hits@K$ metric for each relation in ConceptNet. }
 \label{tab:results_on_each_relation}
 \rotatebox{270}{
    \begin{tabular}{@{}c|ccc|ccc|ccc@{}}
    \hline
    \multicolumn{1}{c|}{\multirow{3}{*}{\textbf{Relations}}} & \multicolumn{9}{c}{$hits@K$} \\ \cline{2-10} 
    \multicolumn{1}{c|}{} & \multicolumn{3}{c|}{GPT1} & \multicolumn{3}{c|}{GPT2} & \multicolumn{3}{c}{GPT3} \\ \cline{2-10} &
    1 & 10 &  \multicolumn{1}{c|}{100} & 1 & 10 &  \multicolumn{1}{c|}{100} & 1 & 10 &  \multicolumn{1}{c}{100} \\ \hline\hline
RelatedTo & 0.03$\pm$0.00 & 3.70$\pm$0.03 & 14.82$\pm$0.06 & 1.50$\pm$0.02 & 15.02$\pm$0.06 & 33.51$\pm$0.08 & 0.19$\pm$0.01 & 15.60$\pm$0.07 & 42.42$\pm$0.09\\
HasContext & 0.01$\pm$0.00 & 3.02$\pm$0.06 & 18.36$\pm$0.13 & 0.32$\pm$0.02 & 12.22$\pm$0.11 & 40.18$\pm$0.17 & 1.75$\pm$0.04 & 16.39$\pm$0.13 & 50.24$\pm$0.17\\
IsA & 0.03$\pm$0.01 & 0.27$\pm$0.02 & 3.77$\pm$0.07 & 19.68$\pm$0.15 & 45.53$\pm$0.19 & 65.28$\pm$0.17 & 22.66$\pm$0.16 & 54.75$\pm$0.18 & 74.73$\pm$0.16\\
DerivedFrom & 0.04$\pm$0.01 & 1.53$\pm$0.05 & 19.43$\pm$0.17 & 0.56$\pm$0.03 & 45.66$\pm$0.22 & 76.88$\pm$0.18 & 0.34$\pm$0.03 & 48.53$\pm$0.22 & 86.00$\pm$0.15\\
Synonym & 5.46$\pm$0.15 & 21.38$\pm$0.27 & 39.83$\pm$0.32 & 5.57$\pm$0.15 & 25.20$\pm$0.29 & 47.76$\pm$0.33 & 2.04$\pm$0.10 & 21.57$\pm$0.28 & 51.34$\pm$0.33\\
FormOf & 0.00$\pm$0.00 & 0.10$\pm$0.02 & 0.91$\pm$0.07 & 0.03$\pm$0.01 & 8.50$\pm$0.20 & 31.09$\pm$0.32 & 0.04$\pm$0.02 & 10.87$\pm$0.23 & 38.98$\pm$0.34\\
EtymologicallyRelatedTo & 0.23$\pm$0.04 & 6.45$\pm$0.28 & 19.35$\pm$0.45 & 0.91$\pm$0.11 & 28.94$\pm$0.52 & 50.48$\pm$0.58 & 2.76$\pm$0.19 & 40.54$\pm$0.56 & 71.66$\pm$0.52\\
SimilarTo & 0.31$\pm$0.06 & 5.91$\pm$0.27 & 24.67$\pm$0.49 & 2.27$\pm$0.17 & 14.74$\pm$0.41 & 40.31$\pm$0.56 & 0.80$\pm$0.11 & 13.35$\pm$0.39 & 42.18$\pm$0.57\\
AtLocation & 0.01$\pm$0.00 & 1.20$\pm$0.12 & 7.72$\pm$0.29 & 0.00$\pm$0.00 & 0.04$\pm$0.02 & 1.02$\pm$0.11 & 0.00$\pm$0.00 & 0.10$\pm$0.03 & 5.62$\pm$0.27\\
MannerOf & 0.91$\pm$0.11 & 7.25$\pm$0.31 & 38.32$\pm$0.57 & 4.24$\pm$0.25 & 17.98$\pm$0.47 & 48.99$\pm$0.59 & 3.69$\pm$0.23 & 17.51$\pm$0.46 & 53.65$\pm$0.59\\
PartOf & 0.01$\pm$0.01 & 1.43$\pm$0.24 & 19.83$\pm$0.81 & 0.00$\pm$0.00 & 0.11$\pm$0.07 & 4.80$\pm$0.44 & 0.00$\pm$0.00 & 0.24$\pm$0.10 & 4.54$\pm$0.43\\
Antonym & 3.66$\pm$0.29 & 15.43$\pm$0.56 & 36.10$\pm$0.74 & 15.11$\pm$0.58 & 33.61$\pm$0.75 & 56.13$\pm$0.76 & 13.98$\pm$0.57 & 37.37$\pm$0.77 & 61.95$\pm$0.73\\
HasProperty & 0.58$\pm$0.14 & 9.34$\pm$0.55 & 34.03$\pm$0.90 & 1.24$\pm$0.21 & 12.81$\pm$0.63 & 42.26$\pm$0.94 & 3.56$\pm$0.36 & 17.37$\pm$0.72 & 48.66$\pm$0.95\\
UsedFor & 0.89$\pm$0.18 & 9.19$\pm$0.53 & 34.65$\pm$0.84 & 2.77$\pm$0.32 & 15.11$\pm$0.67 & 43.36$\pm$0.86 & 3.13$\pm$0.35 & 17.06$\pm$0.69 & 48.42$\pm$0.87\\
DistinctFrom & 0.09$\pm$0.09 & 2.39$\pm$0.40 & 18.38$\pm$1.00 & 2.76$\pm$0.42 & 18.48$\pm$1.01 & 49.02$\pm$1.29 & 2.30$\pm$0.39 & 14.81$\pm$0.90 & 49.25$\pm$1.29\\
HasPrerequisite & 0.00$\pm$0.00 & 5.04$\pm$0.47 & 27.38$\pm$1.01 & 0.00$\pm$0.00 & 2.62$\pm$0.39 & 18.87$\pm$0.95 & 0.00$\pm$0.00 & 2.76$\pm$0.39 & 22.30$\pm$1.00\\
HasSubevent & 0.00$\pm$0.00 & 0.08$\pm$0.04 & 6.35$\pm$0.57 & 0.00$\pm$0.00 & 6.96$\pm$0.63 & 31.70$\pm$1.08 & 0.07$\pm$0.05 & 7.10$\pm$0.62 & 34.72$\pm$1.11\\
Causes & 0.01$\pm$0.01 & 2.44$\pm$0.46 & 19.53$\pm$1.07 & 0.32$\pm$0.18 & 1.13$\pm$0.32 & 11.22$\pm$0.92 & 0.46$\pm$0.21 & 7.14$\pm$0.75 & 30.13$\pm$1.19\\
HasA & 0.12$\pm$0.12 & 6.14$\pm$0.81 & 28.70$\pm$1.53 & 0.60$\pm$0.27 & 6.25$\pm$0.83 & 22.60$\pm$1.41 & 1.44$\pm$0.41 & 11.31$\pm$1.09 & 32.11$\pm$1.58\\
InstanceOf & 0.00$\pm$0.00 & 0.00$\pm$0.00 & 9.14$\pm$1.20 & 0.00$\pm$0.00 & 6.44$\pm$1.01 & 25.89$\pm$1.81 & 0.00$\pm$0.00 & 4.89$\pm$0.89 & 51.47$\pm$2.07\\
CapableOf & 1.38$\pm$0.46 & 15.33$\pm$1.38 & 49.13$\pm$1.88 & 2.04$\pm$0.55 & 18.84$\pm$1.49 & 51.70$\pm$1.88 & 2.75$\pm$0.64 & 21.86$\pm$1.58 & 55.76$\pm$1.86\\
MotivatedByGoal & 0.00$\pm$0.00 & 0.00$\pm$0.00 & 9.47$\pm$0.92 & 0.33$\pm$0.13 & 3.48$\pm$0.52 & 18.30$\pm$1.15 & 0.61$\pm$0.23 & 3.49$\pm$0.55 & 20.42$\pm$1.23\\
MadeOf & 11.60$\pm$1.83 & 39.25$\pm$2.79 & 69.62$\pm$2.61 & 12.40$\pm$1.91 & 51.42$\pm$2.84 & 82.25$\pm$2.18 & 21.05$\pm$2.35 & 56.43$\pm$2.81 & 82.31$\pm$2.18\\
Entails & 0.00$\pm$0.00 & 2.13$\pm$0.86 & 8.69$\pm$1.67 & 0.35$\pm$0.35 & 1.06$\pm$0.61 & 15.01$\pm$2.09 & 0.00$\pm$0.00 & 1.77$\pm$0.79 & 15.96$\pm$2.17\\
Desires & 0.51$\pm$0.51 & 3.59$\pm$1.34 & 33.88$\pm$3.33 & 2.62$\pm$1.14 & 12.88$\pm$2.31 & 32.23$\pm$3.28 & 4.10$\pm$1.42 & 13.32$\pm$2.38 & 32.56$\pm$3.29\\
NotHasProperty & 0.64$\pm$0.64 & 8.92$\pm$2.28 & 22.51$\pm$3.30 & 1.27$\pm$0.90 & 13.59$\pm$2.73 & 32.91$\pm$3.72 & 2.55$\pm$1.26 & 12.95$\pm$2.67 & 38.64$\pm$3.89\\
CreatedBy & 0.00$\pm$0.00 & 13.83$\pm$3.42 & 41.76$\pm$4.98 & 0.00$\pm$0.00 & 21.81$\pm$4.11 & 49.91$\pm$4.97 & 1.06$\pm$1.06 & 24.82$\pm$4.32 & 48.23$\pm$4.93\\
DefinedAs & 0.00$\pm$0.00 & 2.22$\pm$1.75 & 22.78$\pm$5.40 & 3.33$\pm$2.34 & 18.89$\pm$5.03 & 45.28$\pm$6.38 & 8.33$\pm$3.60 & 32.50$\pm$6.04 & 54.17$\pm$6.43\\
NotDesires & 0.01$\pm$0.01 & 2.91$\pm$2.03 & 13.25$\pm$3.95 & 1.45$\pm$1.45 & 12.50$\pm$3.91 & 22.85$\pm$5.00 & 0.00$\pm$0.00 & 6.70$\pm$2.90 & 26.47$\pm$5.21\\
NotCapableOf & 4.88$\pm$3.41 & 24.39$\pm$6.79 & 71.95$\pm$6.99 & 14.63$\pm$5.59 & 36.59$\pm$7.62 & 62.20$\pm$7.57 & 18.29$\pm$5.99 & 42.07$\pm$7.73 & 70.12$\pm$7.16\\
LocatedNear & 0.00$\pm$0.00 & 0.00$\pm$0.00 & 20.31$\pm$7.05 & 0.00$\pm$0.00 & 0.00$\pm$0.00 & 15.62$\pm$6.52 & 0.00$\pm$0.00 & 3.12$\pm$3.12 & 28.12$\pm$8.08\\
EtymologicallyDerivedFrom & 0.00$\pm$0.00 & 4.35$\pm$4.35 & 8.70$\pm$6.01 & 0.00$\pm$0.00 & 17.39$\pm$8.08 & 30.43$\pm$9.81 & 0.00$\pm$0.00 & 26.09$\pm$9.36 & 60.87$\pm$10.41\\
\hline
    \end{tabular}
    }
    {\footnotesize
\begin{tablenotes}
\item[] Data presented as mean $\pm$ standard error of mean
\end{tablenotes}
}

\end{table*}

\newpage

\section{Detailed results on the MadeOf relations}
\label{apx:MadeOf}

\begin{figure*}[!ht]
\centering
\begin{subfigure}[b]{0.25\textwidth}
\includegraphics[width=\linewidth]{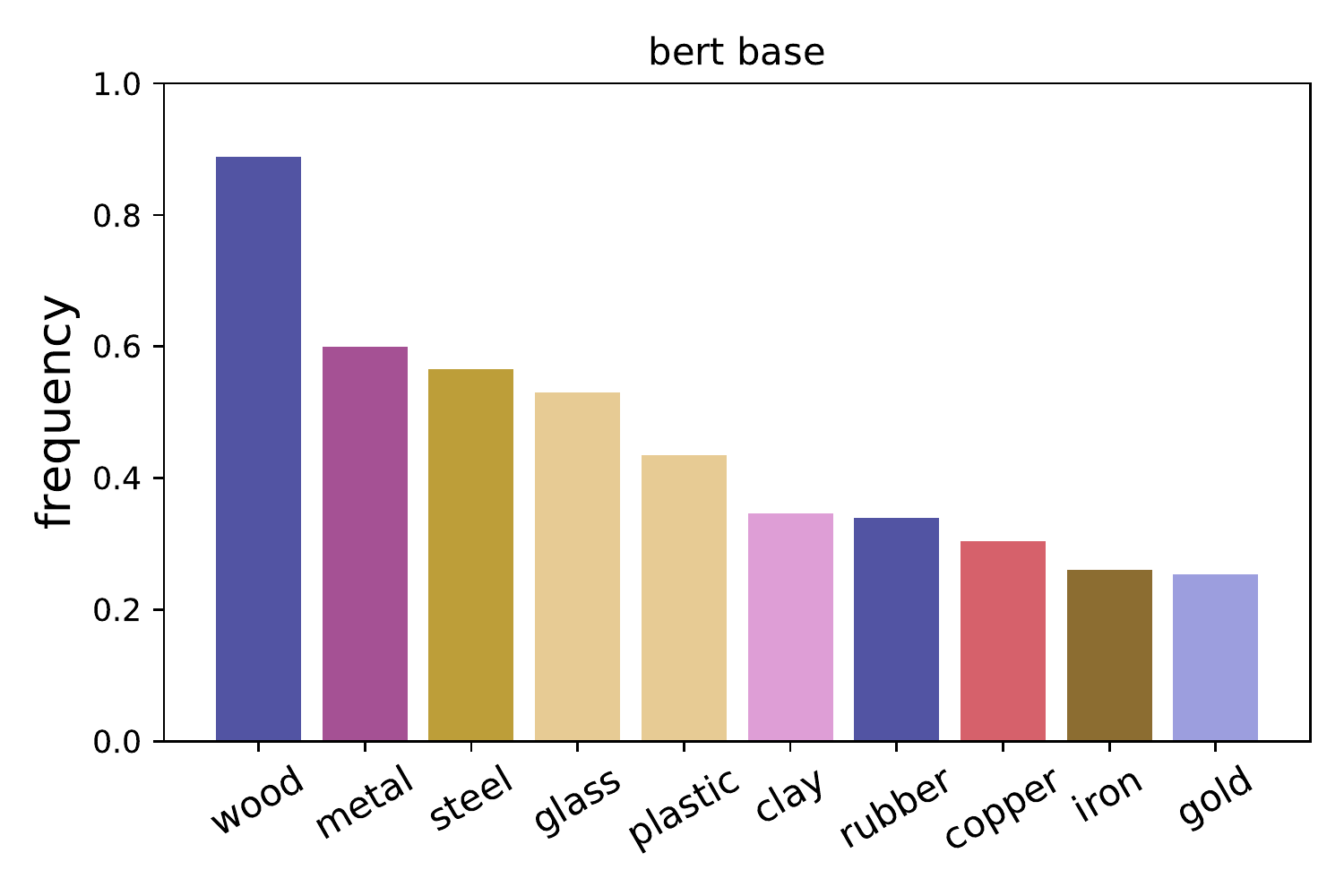}
\caption{BERT$_{base}$}
\label{fig:madeof_bert_base}
\end{subfigure}
\begin{subfigure}[b]{0.25\textwidth}
\includegraphics[width=\linewidth]{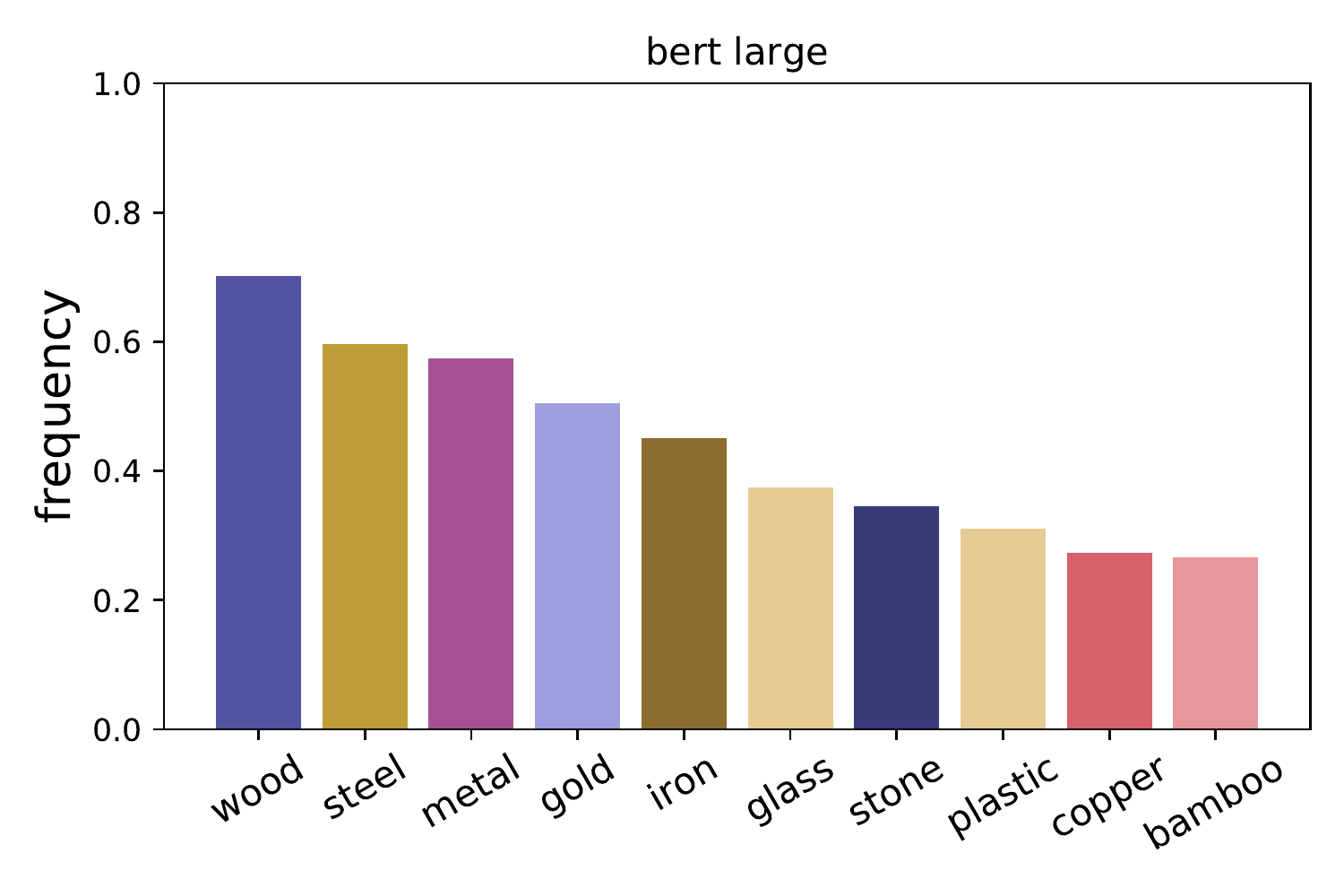}
\caption{BERT$_{large}$}
\label{fig:madeof_bert_base}
\end{subfigure}
\begin{subfigure}[b]{0.25\textwidth}
\includegraphics[width=\linewidth]{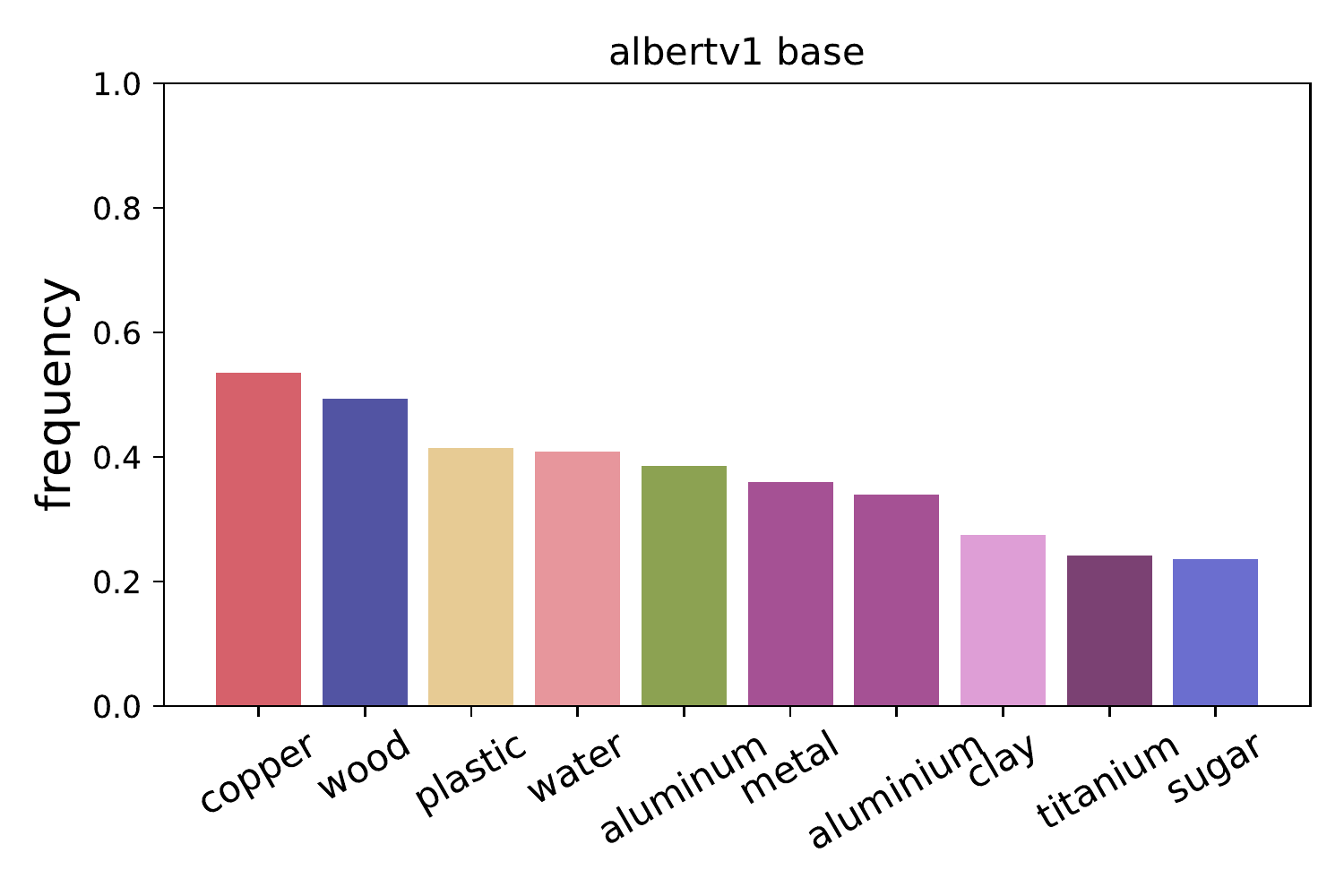}
\caption{ALBERT.v1$_{base}$}
\label{fig:madeof_bert_base}
\end{subfigure}
\begin{subfigure}[b]{0.25\textwidth}
\includegraphics[width=\linewidth]{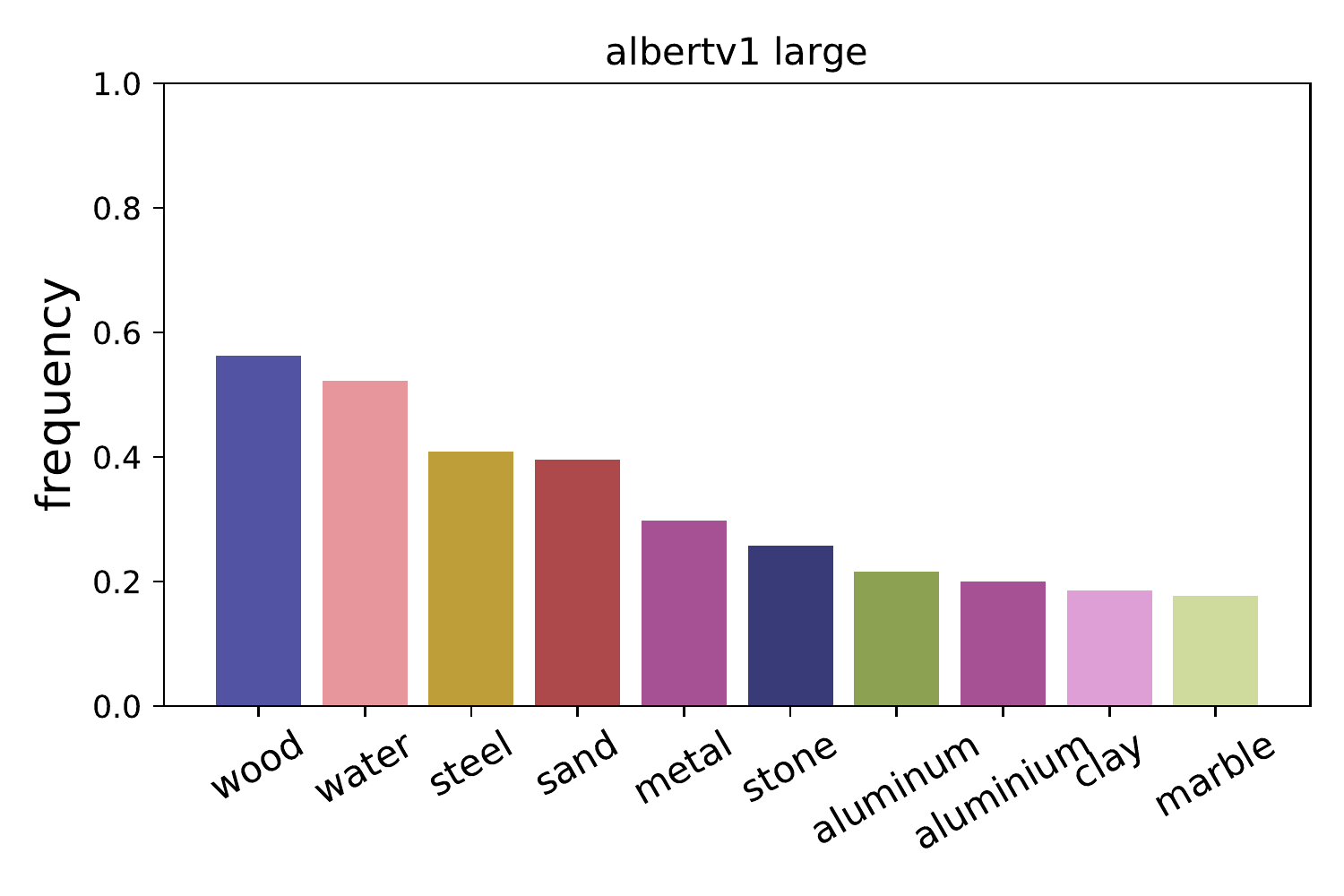}
\caption{ALBERT.v1$_{large}$}
\label{fig:madeof_bert_base}
\end{subfigure}
\begin{subfigure}[b]{0.25\textwidth}
\includegraphics[width=\linewidth]{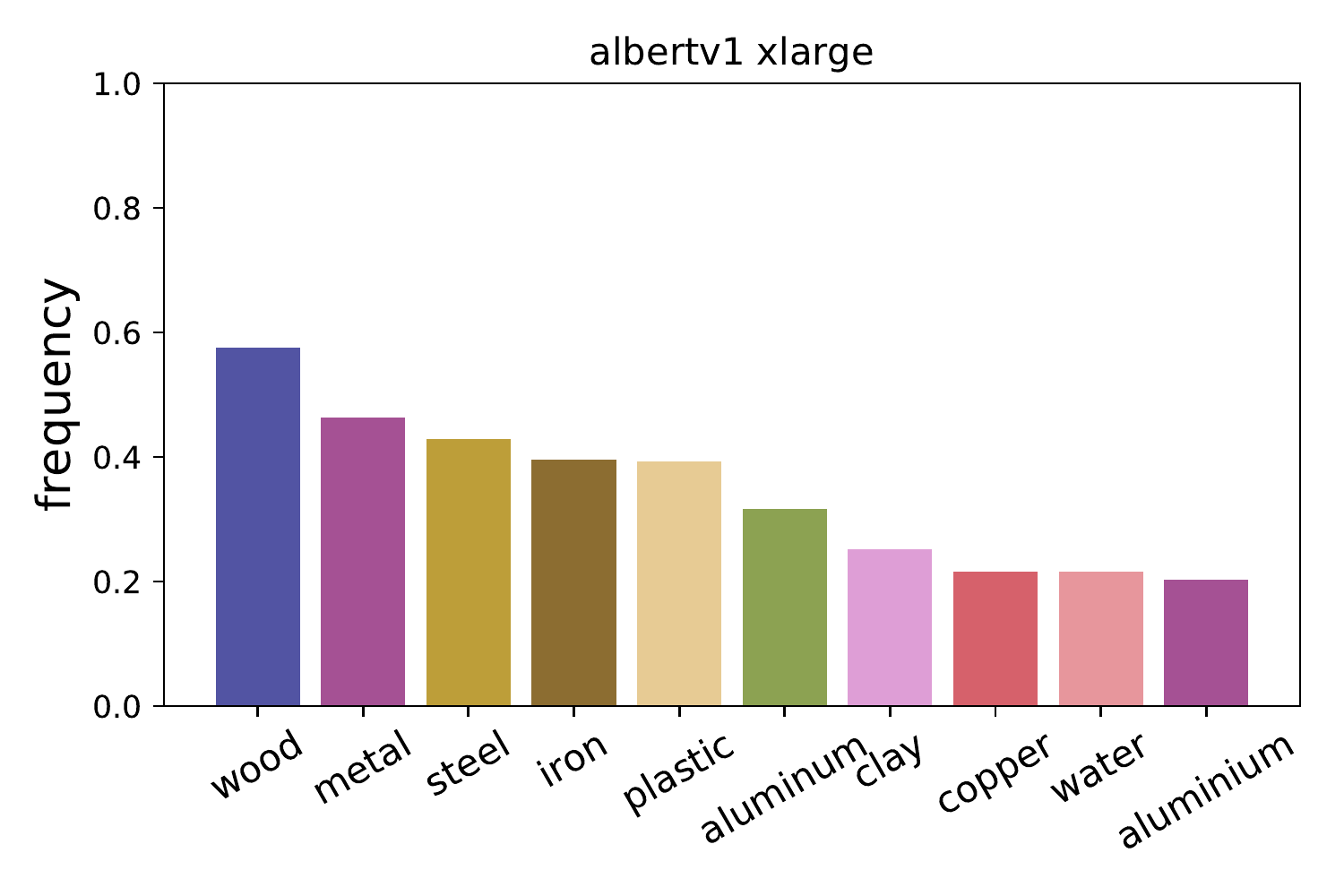}
\caption{ALBERT.v1$_{xlarge}$}
\label{fig:madeof_bert_base}
\end{subfigure}
\begin{subfigure}[b]{0.25\textwidth}
\includegraphics[width=\linewidth]{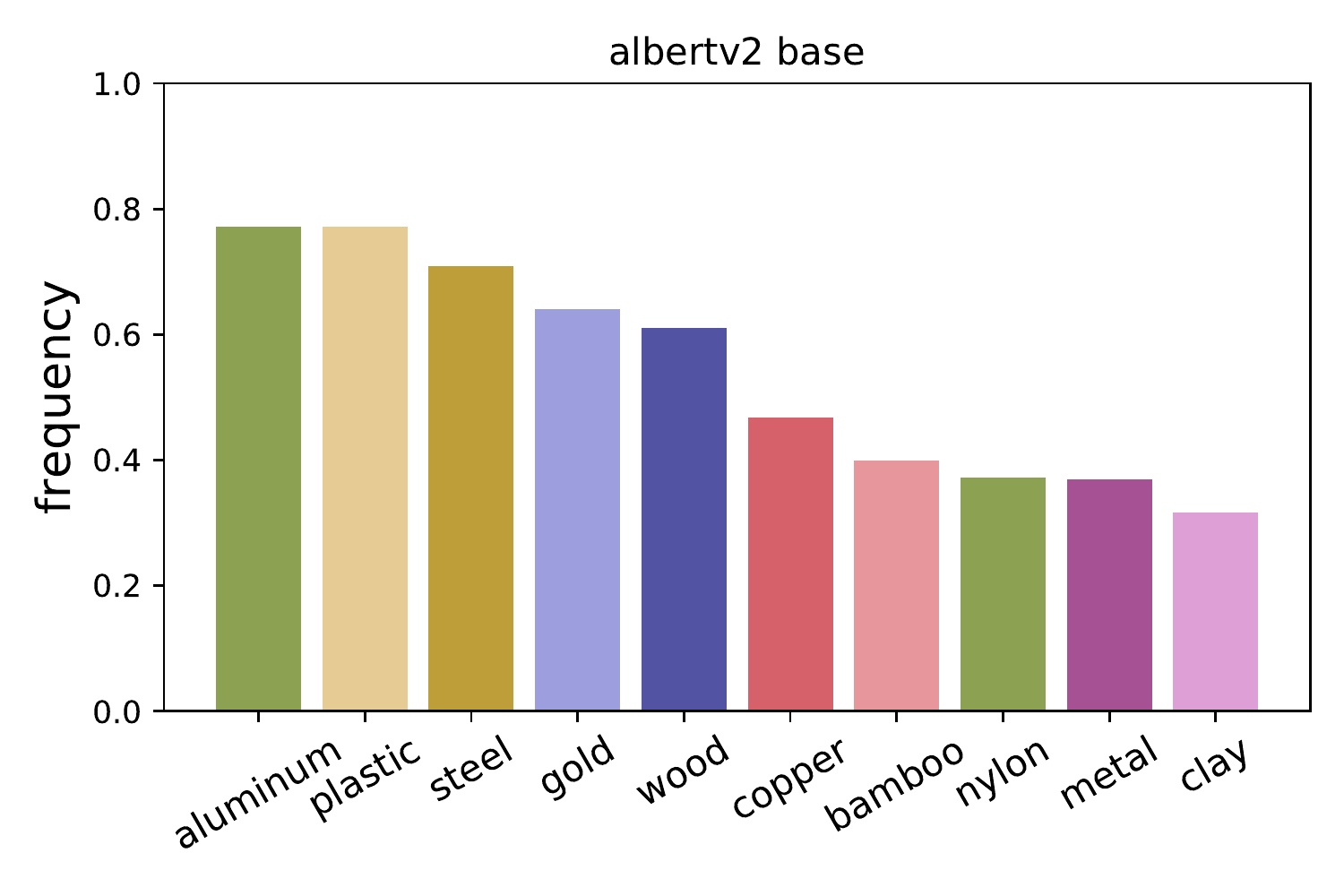}
\caption{ALBERT.v2$_{base}$}
\label{fig:madeof_bert_base}
\end{subfigure}
\begin{subfigure}[b]{0.25\textwidth}
\includegraphics[width=\linewidth]{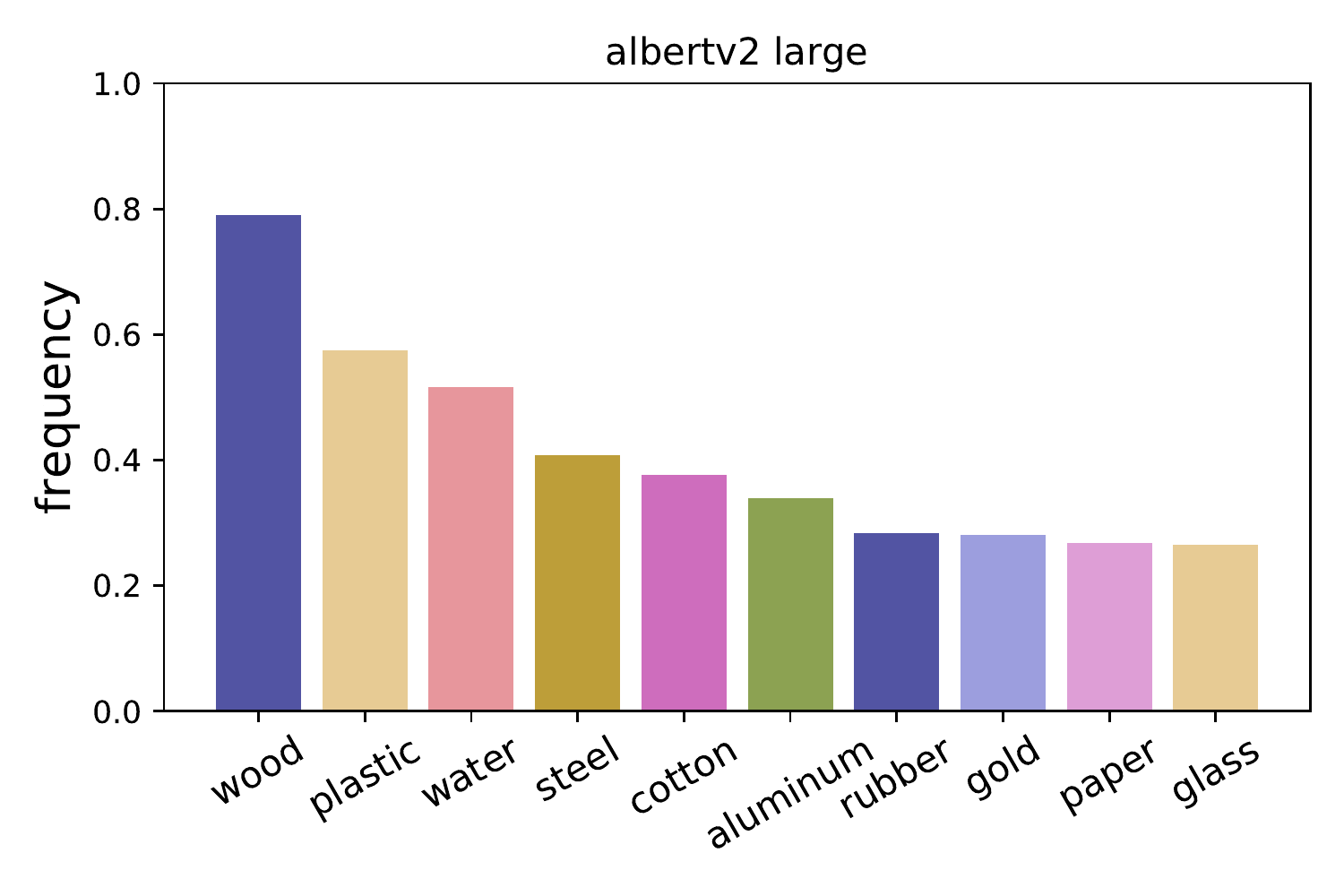}
\caption{ALBERT.v2$_{large}$}
\label{fig:madeof_bert_base}
\end{subfigure}
\begin{subfigure}[b]{0.25\textwidth}
\includegraphics[width=\linewidth]{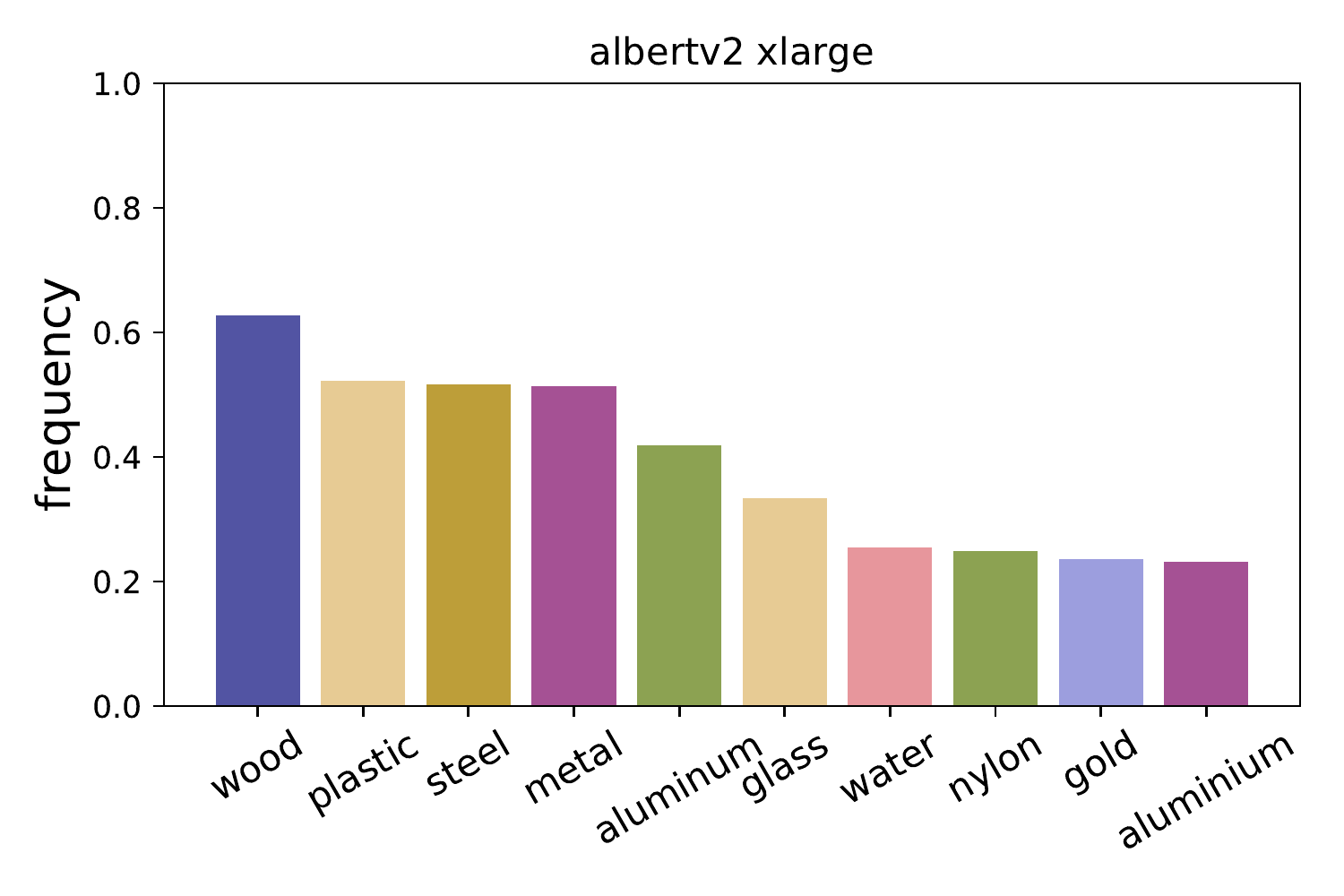}
\caption{ALBERT.v2$_{xlarge}$}
\label{fig:madeof_bert_base}
\end{subfigure}
\begin{subfigure}[b]{0.25\textwidth}
\includegraphics[width=\linewidth]{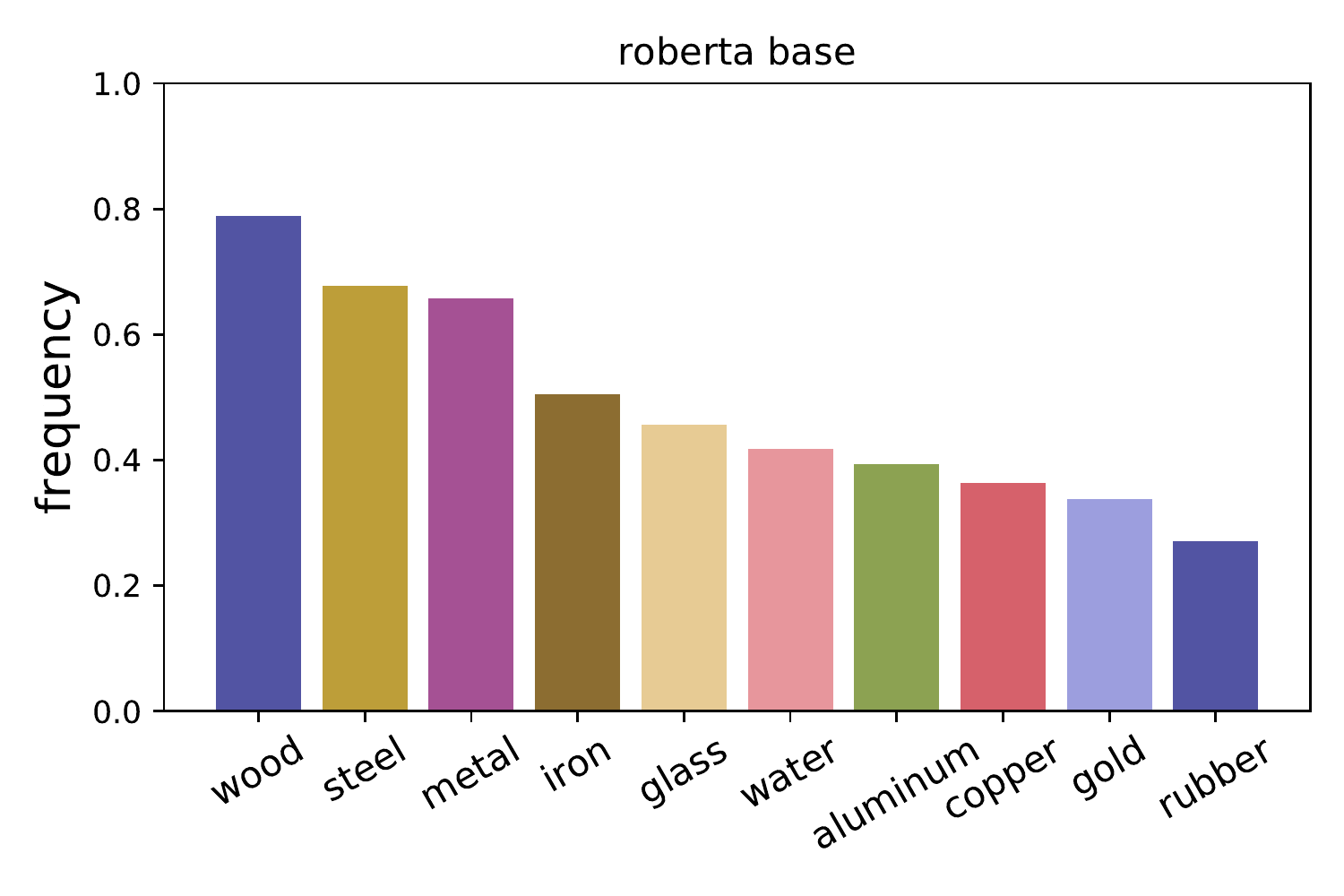}
\caption{RoBERTa$_{base}$}
\label{fig:madeof_bert_base}
\end{subfigure}
\begin{subfigure}[b]{0.25\textwidth}
\includegraphics[width=\linewidth]{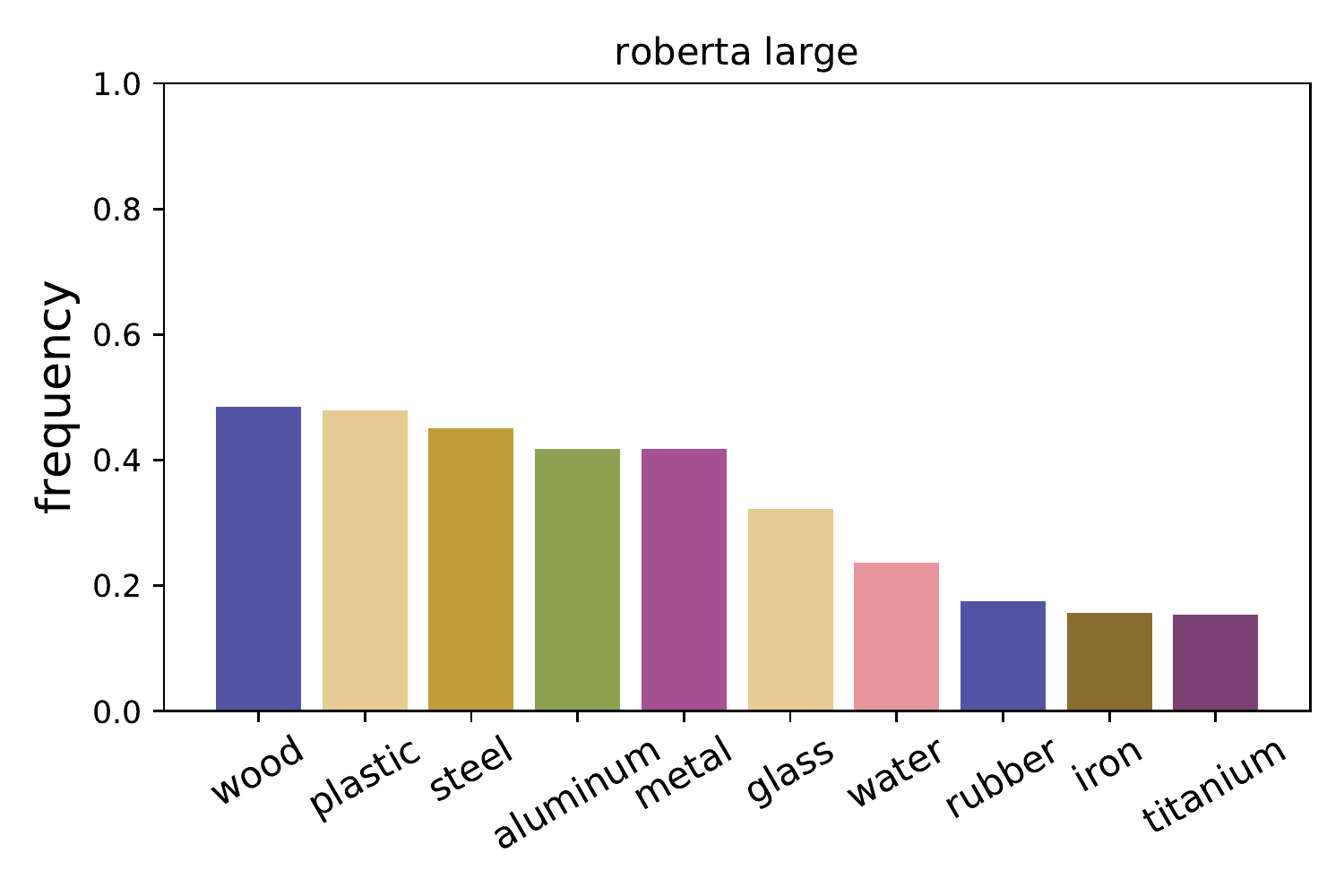}
\caption{RoBERTa$_{large}$}
\label{fig:madeof_bert_base}
\end{subfigure}
\begin{subfigure}[b]{0.25\textwidth}
\includegraphics[width=\linewidth]{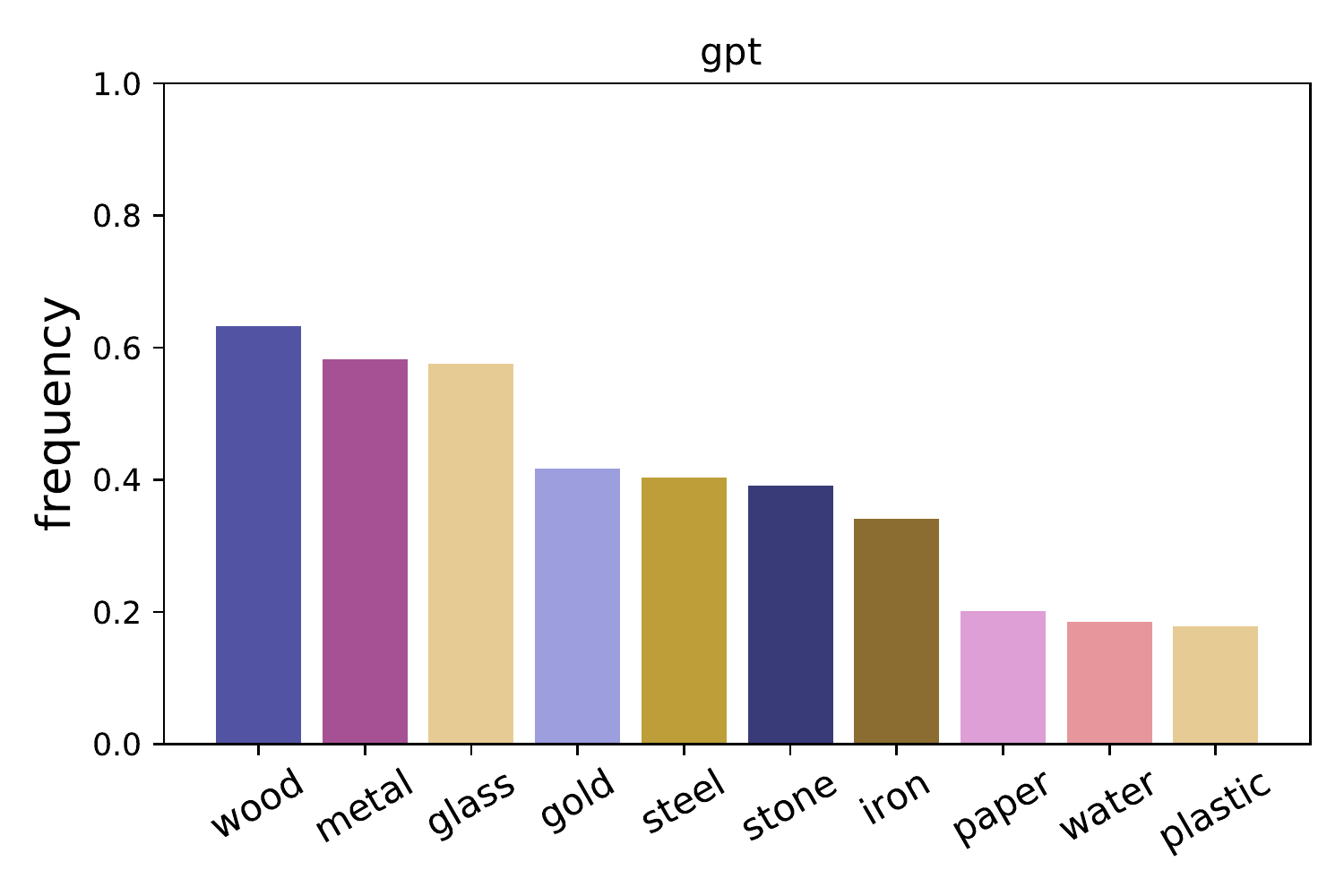}
\caption{GPT}
\label{fig:madeof_bert_base}
\end{subfigure}
\begin{subfigure}[b]{0.25\textwidth}
\includegraphics[width=\linewidth]{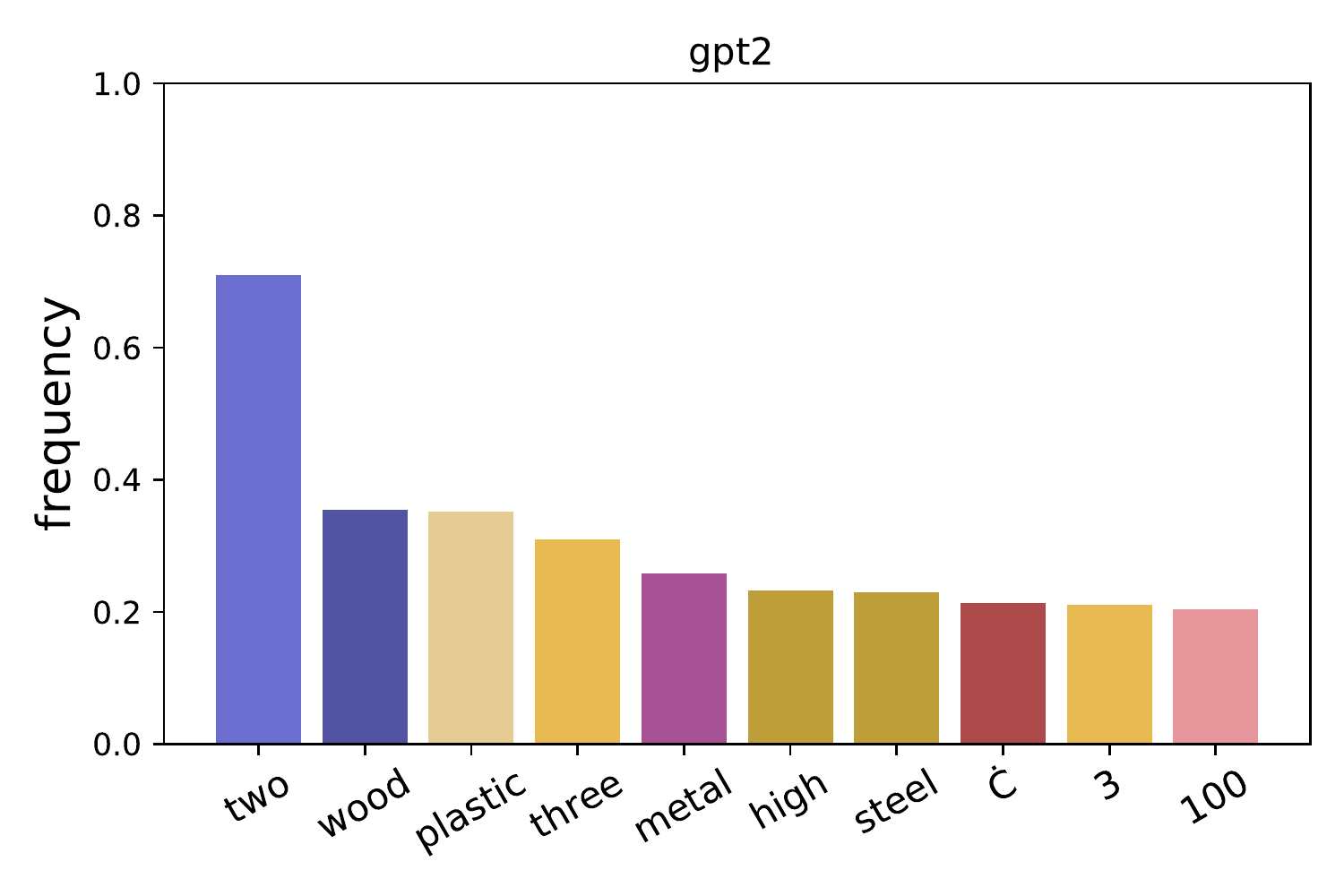}
\caption{GPT2}
\label{fig:madeof_bert_base}
\end{subfigure}
% \begin{subfigure}[b]{0.3\textwidth}
% \includegraphics[width=\linewidth]{images/rebuttal_madeof_gpt3_ada_.pdf}
% \caption{GPT3 Ada}
% \label{fig:madeof_bert_base}
% \end{subfigure}
% \begin{subfigure}[b]{0.3\textwidth}
% \includegraphics[width=\linewidth]{images/rebuttal_madeof_gpt3_curie_.pdf}
% \caption{GPT3 Curie}
% \label{fig:madeof_bert_base}
% \end{subfigure}
\begin{subfigure}[b]{0.25\textwidth}
\includegraphics[width=\linewidth]{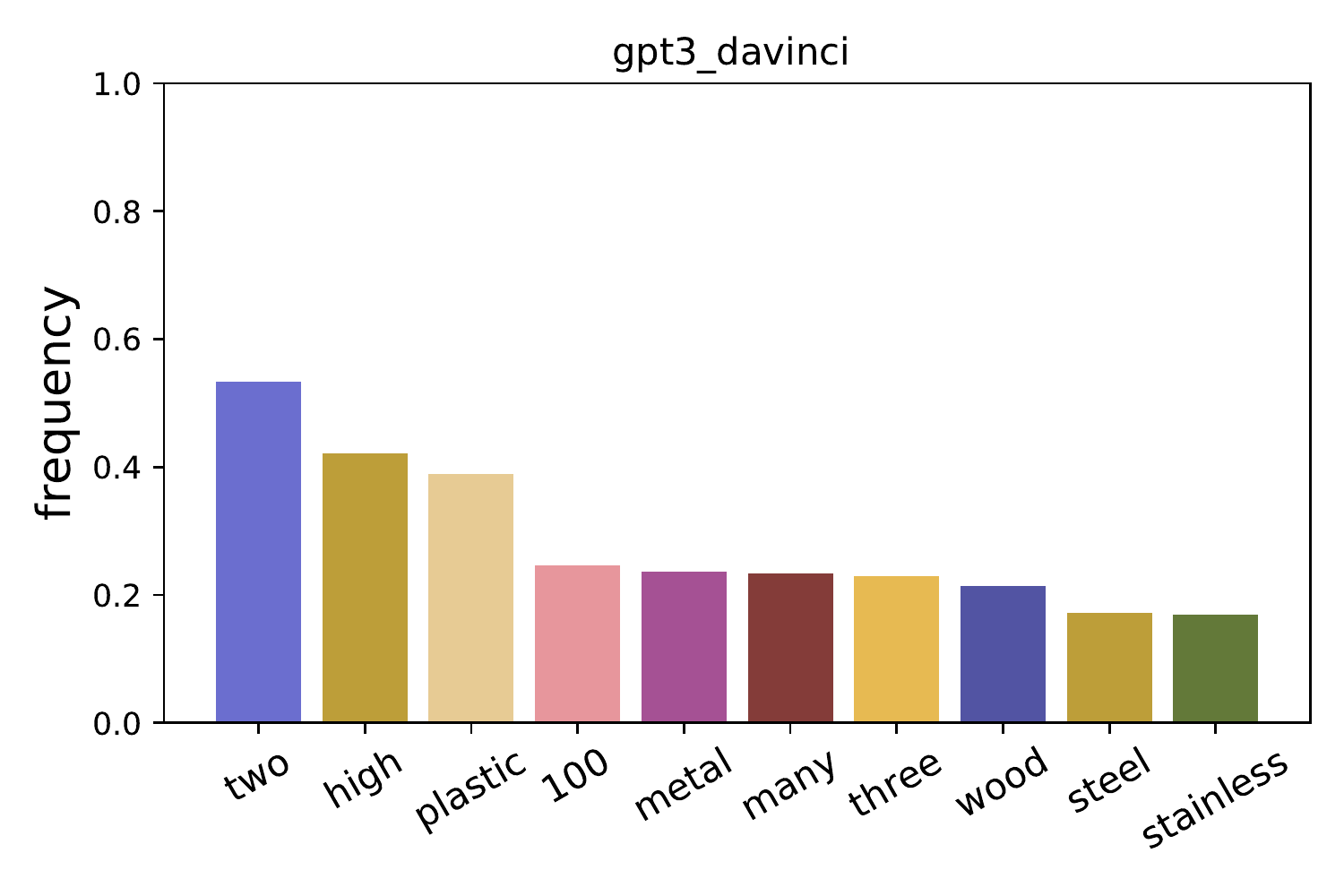}
\caption{GPT3 Davinci}
\label{fig:madeof_bert_base}
\end{subfigure}
\caption{
The 10 most frequent prediction words and its frequencies in the top 10 predictions of `MadeOf' relation. Each figure demonstrates the frequencies of the 10 most frequent words for each model. The frequency of a prediction word indicates the ratio of the number of cases in which the word appears in top 10 predictions over the total number of samples. Each word is colored distinctively throughout figures. }
\end{figure*}

\newpage

% \section{Detailed Results on the Switched-grading Between Synonym and Antonym}
% \label{apx:synonym antonym grading}
% \begin{table}[!ht]
% \centering
% \caption{Experimental results on the incorrect rate between `Synonym' and `Antonym' relations. The incorrect rate is calculated by grading predictions of the `Model' over the semantic triples of the `Relation' with the `Answer' from the opposite relation. }
% \label{tab:intergrade_results_all}
% \begin{tabular}{c|c|c|cc}
% \hline
% \multirow{2}{*}{Model}& \multirow{2}{*}{Relation} & \multirow{2}{*}{Answer} & \multicolumn{2}{c}{Hits@K} \\
% \cline{4-5}
%  & & & 10  & 100 \\
% \hline
% \hline
% \multirow{2}{*}{BERT$_{base}$} & Synonym& Antonym & 30.58 &	54.16 \\
% &Antonym & Synonym & 26.25 &	47.18\\\hline
% \multirow{2}{*}{BERT$_{large}$} &Synonym& Antonym& 39.36 &	60.71 \\
%  & Antonym & Synonym & 25.02 &	48.34 \\
% \hline
% \multirow{2}{*}{ALBERT$_{base}$} & Synonym& Antonym & 26.71 &	47.16 \\
% &Antonym & Synonym & 22.60 &	41.09 \\\hline
% \multirow{2}{*}{ALBERT$_{large}$} &Synonym& Antonym& 31.85 &	55.17 \\
%  & Antonym & Synonym & 24.00 &	42.77 \\
% \hline
% \multirow{2}{*}{ALBERT$_{xlarge}$} &Synonym& Antonym& 35.45 &	56.78 \\
%  & Antonym & Synonym & 25.64 &	47.79 \\
% \hline
% \end{tabular}
% \end{table}
\newpage
\section{Examples of Overlapping Synonym and Antonym}
\label{apx:additional_examples_on_overlapping}
\begin{figure*}[!ht]
\centering
\begin{subfigure}[b]{0.45\textwidth}
\includegraphics[width=\linewidth]{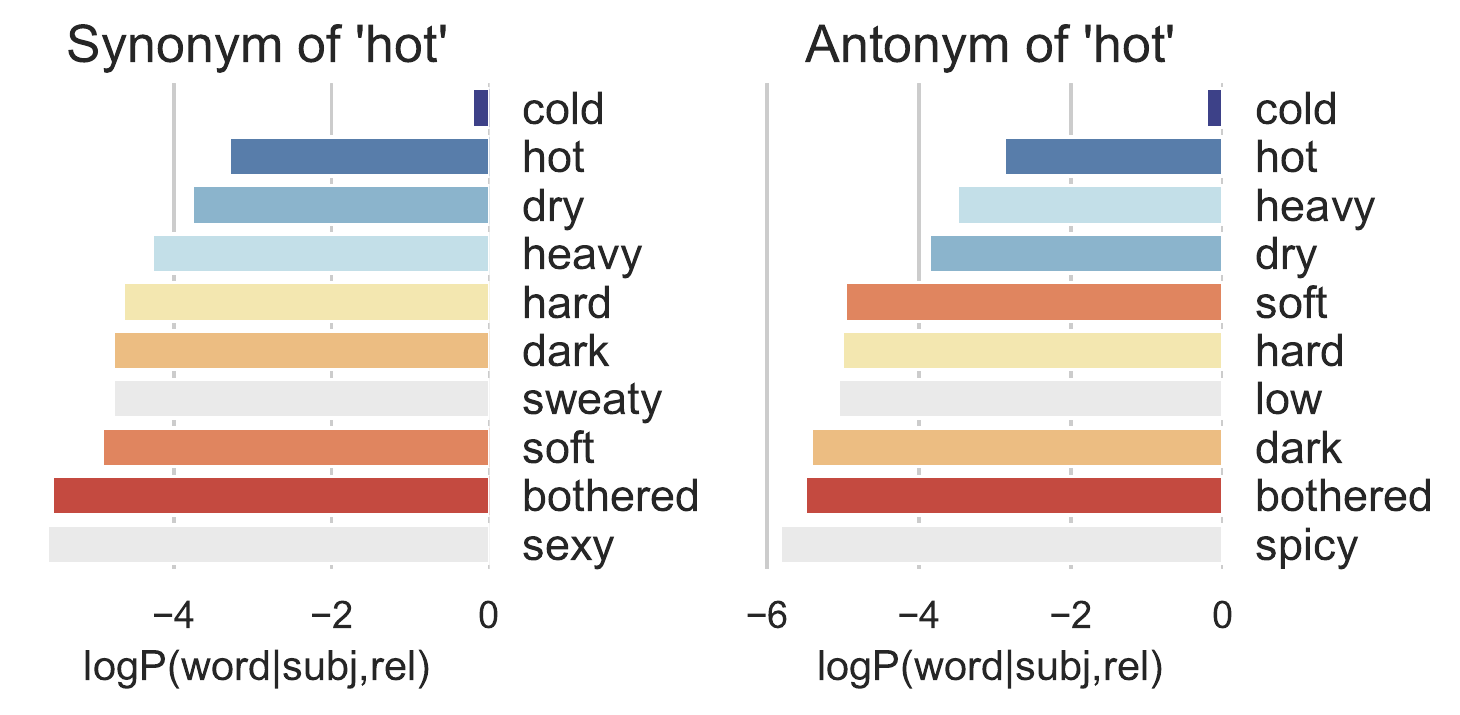}
\caption{`Hot'}
\end{subfigure}
\begin{subfigure}[b]{0.45\textwidth}
\includegraphics[width=\linewidth]{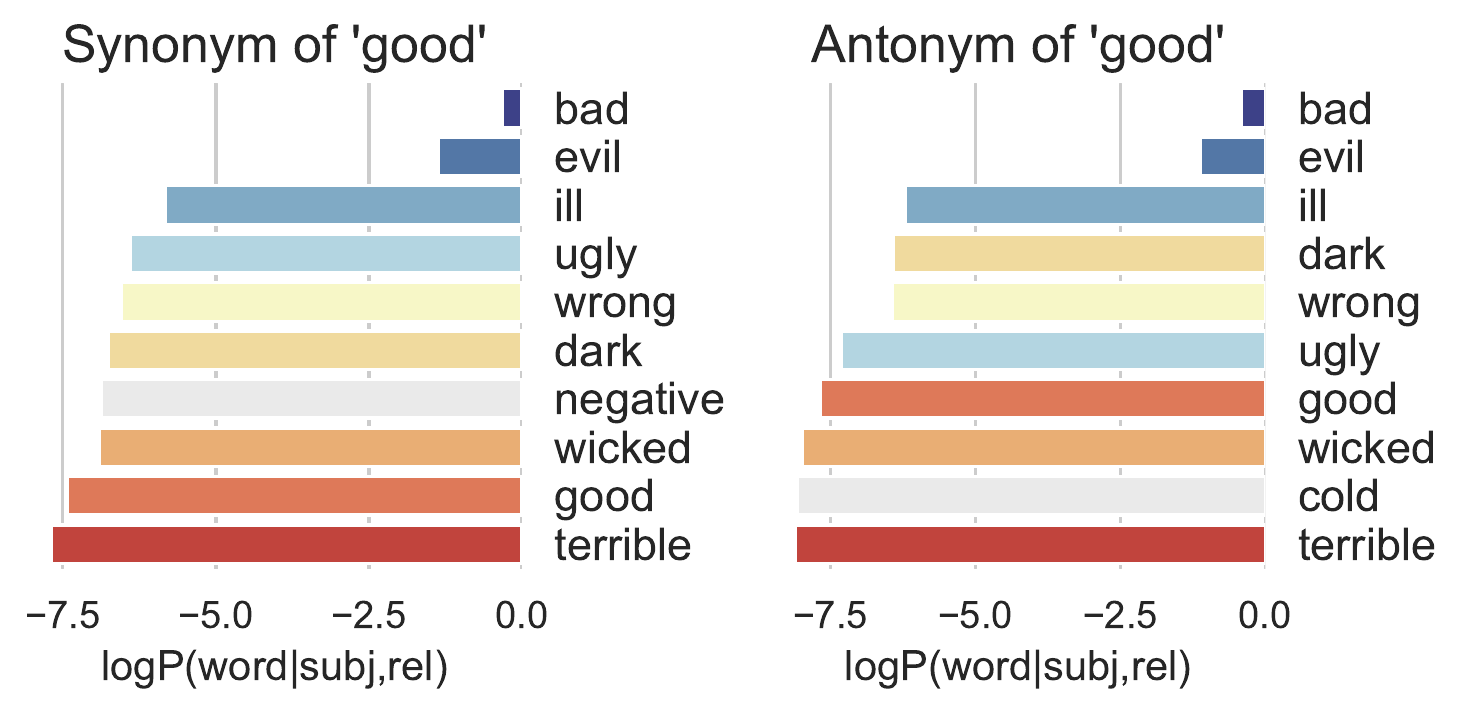}
\caption{`Good'}
\end{subfigure}
% \begin{subfigure}[b]{0.45\textwidth}
% \includegraphics[width=\linewidth]{images/high_diff_color.pdf}
% \caption{`High'}
% \end{subfigure}
\begin{subfigure}[b]{0.45\textwidth}
\includegraphics[width=\linewidth]{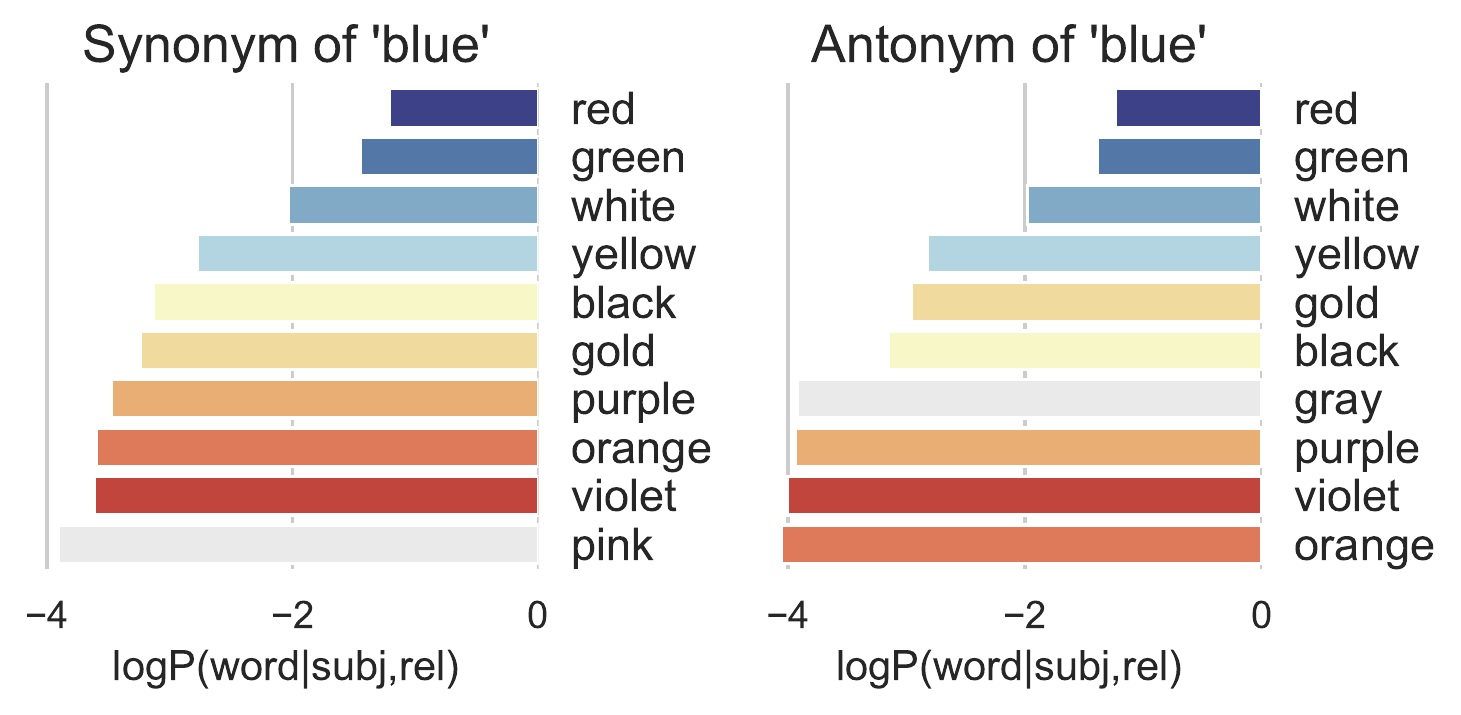}
\caption{`Blue'}
\end{subfigure}
\begin{subfigure}[b]{0.45\textwidth}
\includegraphics[width=\linewidth]{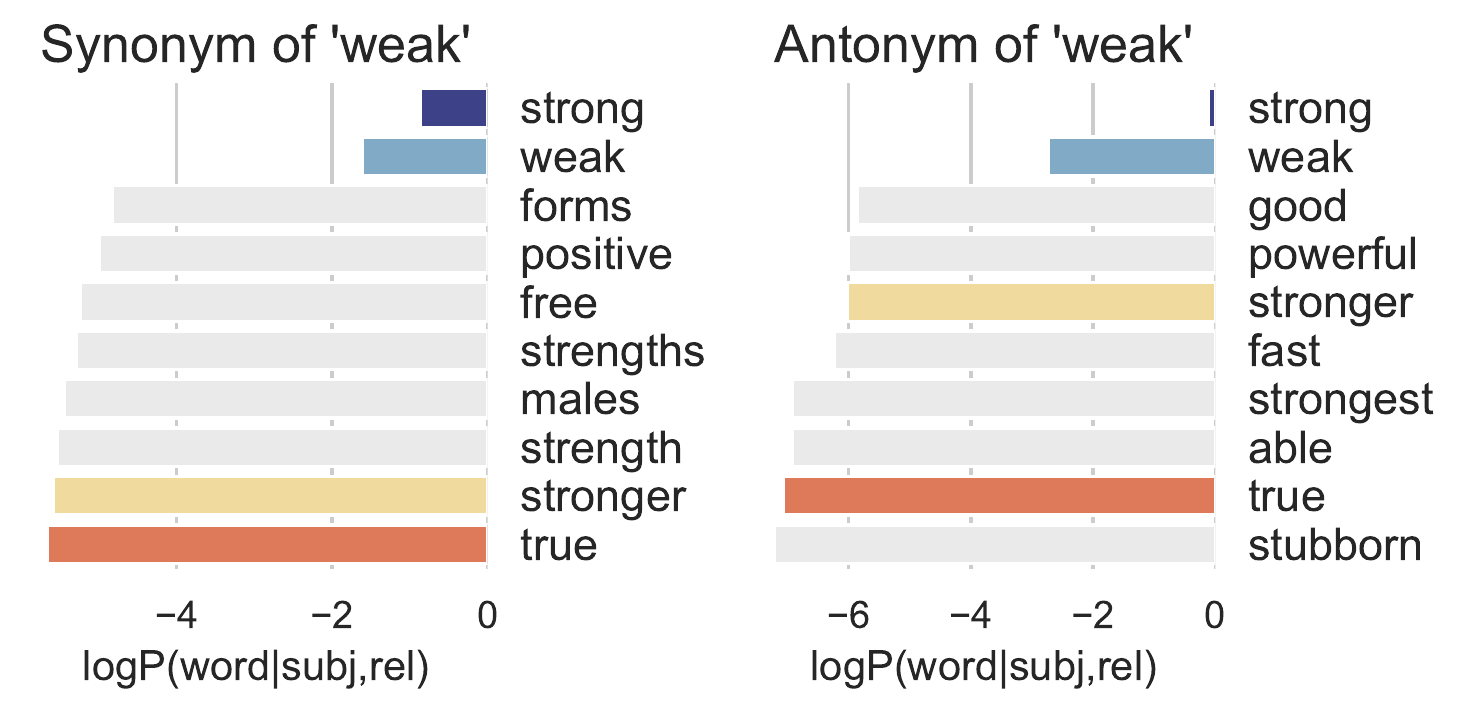}
\caption{`Weak'}
\end{subfigure}
\begin{subfigure}[b]{0.45\textwidth}
\includegraphics[width=\linewidth]{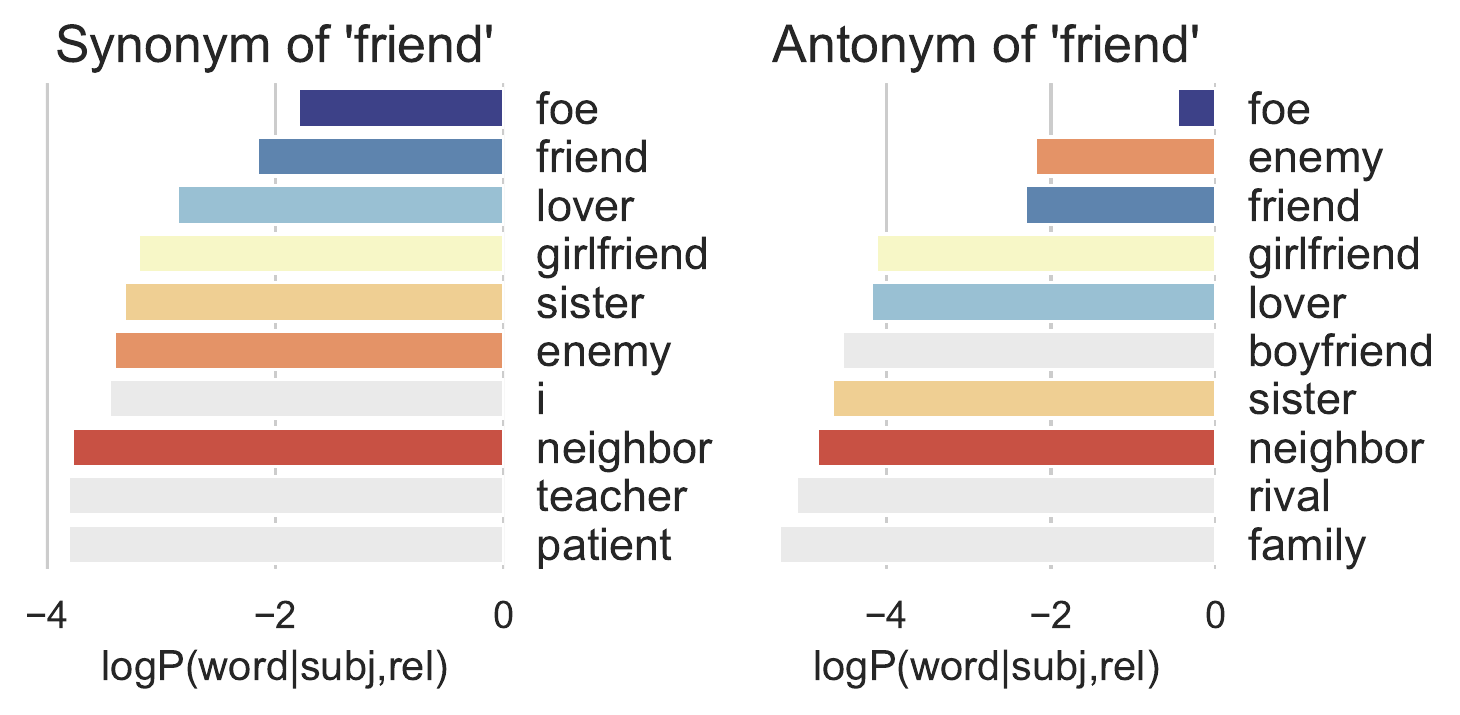}
\caption{`Friend'}
\end{subfigure}
\begin{subfigure}[b]{0.45\textwidth}
\includegraphics[width=\linewidth]{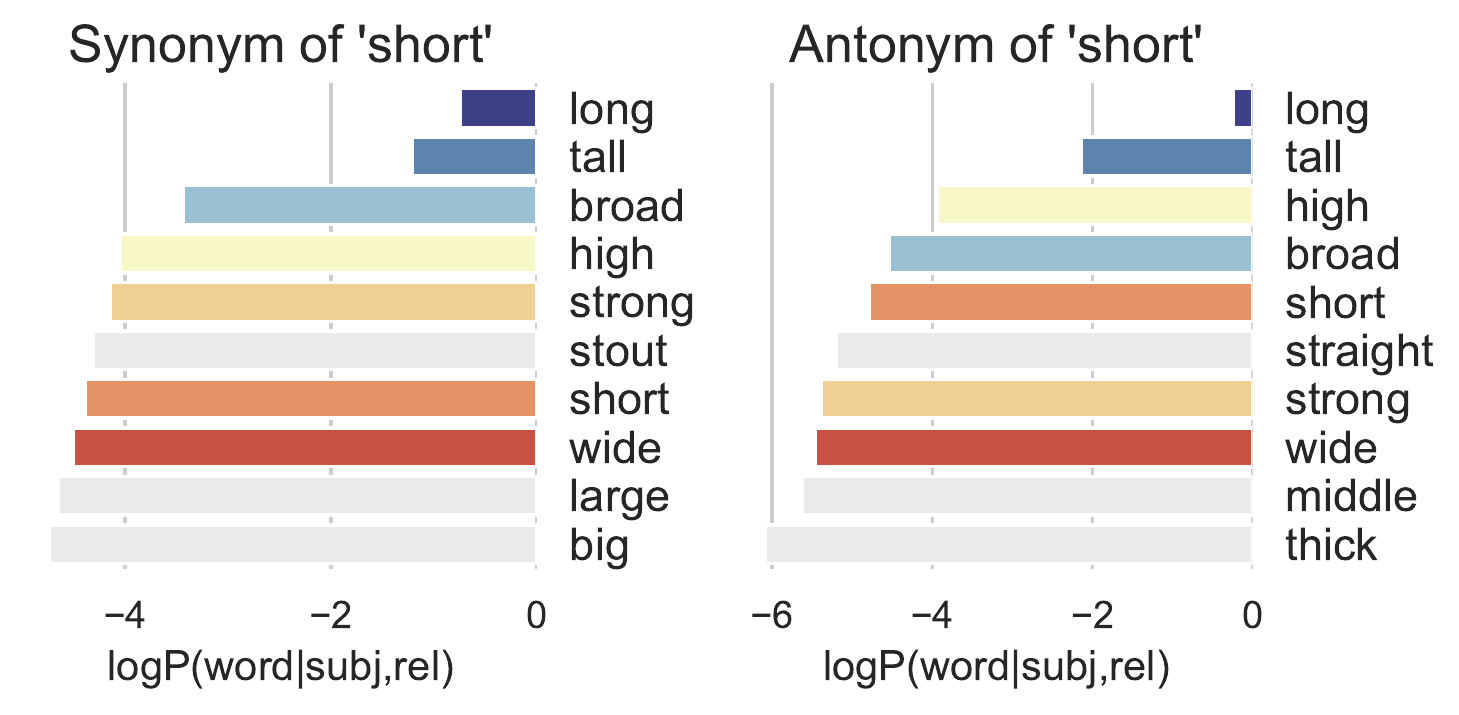}
\caption{`Short'}
\end{subfigure}
\begin{subfigure}[b]{0.45\textwidth}
\includegraphics[width=\linewidth]{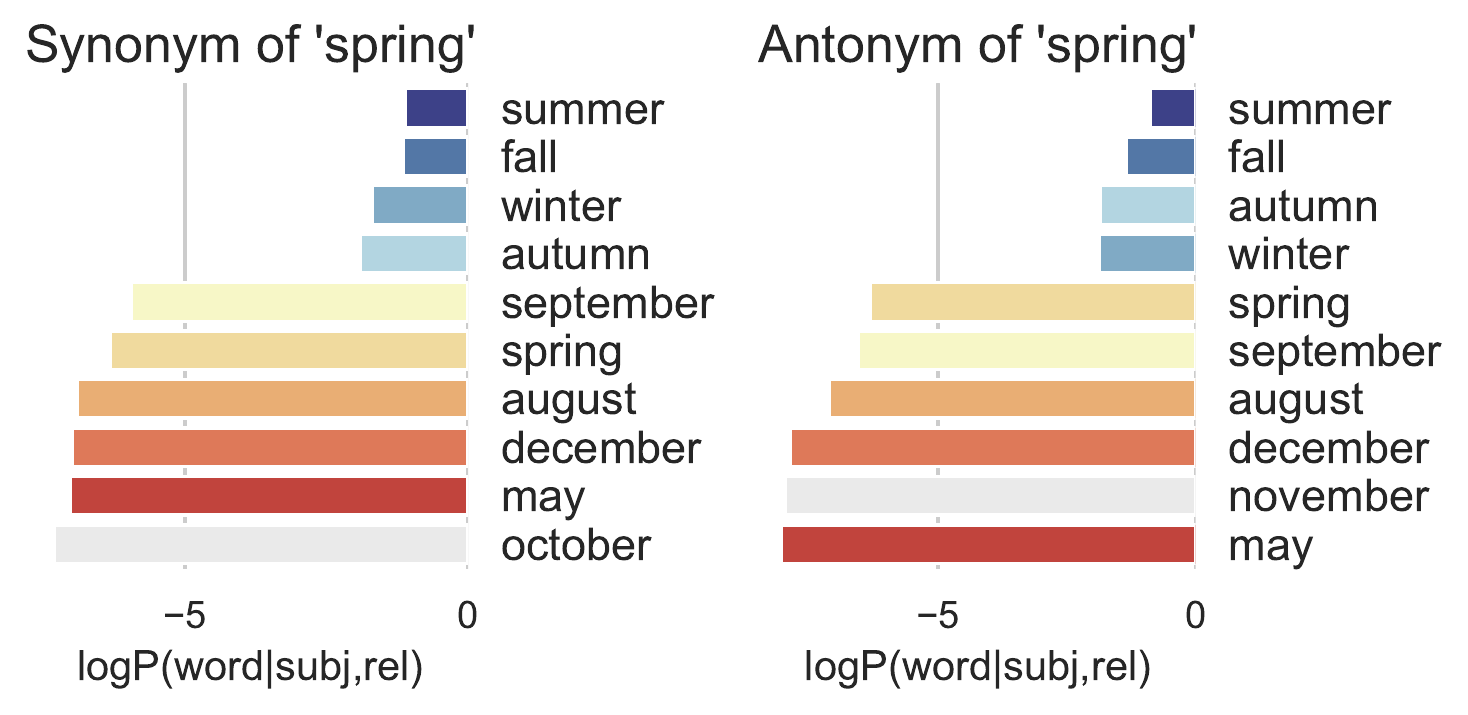}
\caption{`Spring'}
\end{subfigure}
\begin{subfigure}[b]{0.45\textwidth}
\includegraphics[width=\linewidth]{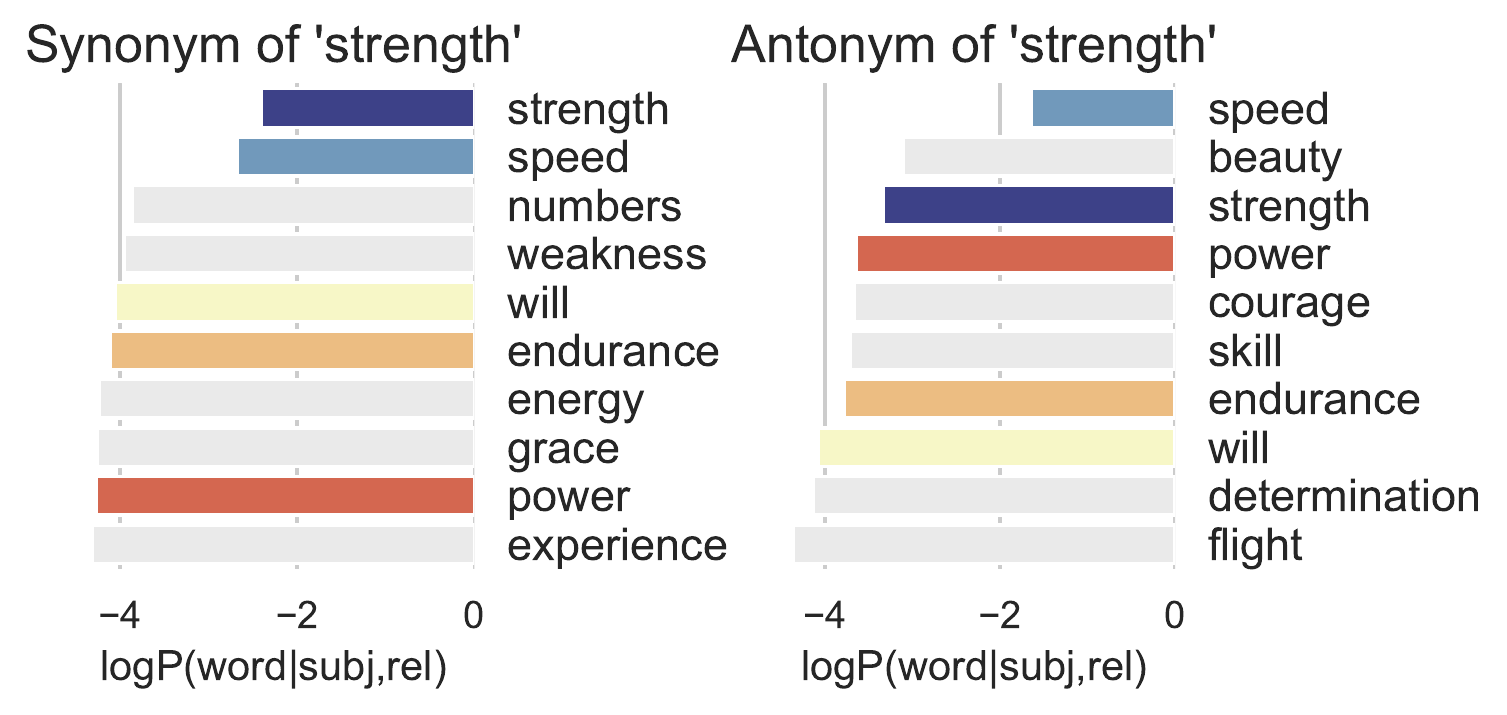}
\caption{`Strength'}
\end{subfigure}

\end{figure*}
\newpage
\section{Impact of the Finetuning on ConceptNet}
\label{apx:finetuning_on_KB}
\begin{table*}[!h]
 \centering
 \small
 \caption{Results of the fine-tuned BERT$_{base}$ $hits@K$ metric on the folds for each relation in ConceptNet. }
 \label{tab:results_on_each_relation}
 \rotatebox{0}{
    \begin{tabular}{@{}c|ccc|ccc|ccc@{}}
    \hline
    \multicolumn{1}{c|}{\multirow{3}{*}{\textbf{Relations}}} & \multicolumn{9}{c}{$hits@K$} \\ \cline{2-10} 
    \multicolumn{1}{c|}{} & \multicolumn{3}{c|}{Fold 0} & \multicolumn{3}{c|}{Fold 1} & \multicolumn{3}{c}{Fold 1} \\ \cline{2-10} &
    1 & 10 &  \multicolumn{1}{c|}{100} & 1 & 10 &  \multicolumn{1}{c|}{100} & 1 & 10 &  \multicolumn{1}{c}{100} \\ \hline\hline
\multirow{-1}{*}{RelatedTo} & 18.30 & 34.70 & 54.20 & 18.50 & 34.90 & 54.70 & 18.60 & 35.00 & 54.50 \\
HasContext & 37.60 & 74.30 & 95.40 & 37.50 & 74.40 & 95.40 & 38.20 & 74.50 & 95.50 \\
IsA & 39.60 & 65.40 & 83.60 & 39.10 & 64.70 & 83.20 & 39.70 & 64.80 & 83.20 \\
DerivedFrom & 57.00 & 80.70 & 88.70 & 58.40 & 81.60 & 89.10 & 57.80 & 81.00 & 88.60 \\
Synonym & 23.60 & 48.00 & 68.60 & 23.90 & 49.00 & 68.80 & 23.00 & 47.70 & 68.00 \\
FormOf & 77.20 & 88.70 & 92.00 & 76.70 & 88.60 & 91.80 & 75.90 & 88.10 & 91.90 \\
EtymologicallyRelatedTo & 20.90 & 35.40 & 48.90 & 19.70 & 33.00 & 44.90 & 19.10 & 32.80 & 45.60 \\
SimilarTo & 8.20 & 33.40 & 60.90 & 7.60 & 33.20 & 61.90 & 7.40 & 32.70 & 62.80 \\
AtLocation & 18.20 & 50.20 & 85.10 & 15.70 & 48.50 & 84.40 & 17.00 & 48.60 & 83.90 \\
MannerOf & 12.20 & 38.00 & 69.20 & 12.90 & 36.80 & 67.10 & 13.10 & 36.40 & 67.00 \\
PartOf & 10.70 & 32.70 & 62.20 & 9.30 & 33.70 & 64.40 & 9.90 & 35.20 & 63.80 \\
Antonym & 8.70 & 44.10 & 71.50 & 8.10 & 43.70 & 69.90 & 8.90 & 38.90 & 67.10 \\
HasProperty & 11.80 & 40.00 & 72.20 & 10.70 & 38.80 & 72.70 & 9.20 & 37.30 & 73.80 \\
UsedFor & 8.70 & 32.70 & 65.00 & 6.20 & 29.80 & 60.80 & 9.40 & 31.20 & 62.30 \\
DistinctFrom & 7.40 & 48.10 & 76.80 & 6.80 & 54.30 & 76.00 & 7.20 & 54.20 & 79.50 \\
HasPrerequisite & 4.90 & 21.60 & 54.60 & 3.90 & 18.80 & 51.50 & 2.30 & 14.40 & 46.70 \\
HasSubevent & 6.20 & 25.30 & 52.00 & 5.50 & 23.00 & 54.20 & 6.60 & 26.90 & 55.20 \\
Causes & 9.30 & 29.90 & 61.90 & 7.70 & 25.00 & 56.30 & 10.00 & 31.40 & 60.30 \\
HasA & 11.50 & 40.60 & 72.40 & 10.20 & 40.70 & 70.90 & 8.00 & 44.40 & 73.40 \\
InstanceOf & 22.40 & 38.80 & 61.20 & 19.10 & 36.40 & 57.90 & 19.90 & 38.90 & 58.80 \\
CapableOf & 15.10 & 43.90 & 68.30 & 15.60 & 42.40 & 69.80 & 15.40 & 42.80 & 68.80 \\
MotivatedByGoal & 3.40 & 18.60 & 41.20 & 6.20 & 21.90 & 43.30 & 3.80 & 17.70 & 51.60 \\
MadeOf & 26.50 & 60.20 & 84.70 & 19.20 & 60.60 & 83.80 & 17.80 & 52.50 & 73.30 \\
Entails & 9.50 & 28.60 & 58.30 & 4.70 & 25.60 & 54.70 & 3.60 & 21.70 & 54.20 \\
Desires & 12.70 & 30.20 & 49.20 & 3.10 & 15.60 & 35.90 & 8.20 & 27.90 & 42.60 \\
NotHasProperty & 6.50 & 26.10 & 54.30 & 9.80 & 25.50 & 47.10 & 8.00 & 32.00 & 58.00 \\
CreatedBy & 0.00 & 37.10 & 68.60 & 0.00 & 36.80 & 71.10 & 9.10 & 39.40 & 69.70 \\
DefinedAs & 12.50 & 45.80 & 70.80 & 5.60 & 44.40 & 61.10 & 4.80 & 19.00 & 57.10 \\
NotDesires & 0.00 & 9.10 & 27.30 & 14.30 & 23.80 & 28.60 & 0.00 & 27.30 & 31.80 \\
NotCapableOf & 33.30 & 58.30 & 83.30 & 23.10 & 53.80 & 76.90 & 16.70 & 50.00 & 66.70 \\
LocatedNear & 0.00 & 50.00 & 87.50 & 9.10 & 45.50 & 45.50 & 9.10 & 45.50 & 63.60 \\
EtymologicallyDerivedFrom & 33.30 & 50.00 & 50.00 & 16.70 & 16.70 & 16.70 & 22.20 & 22.20 & 22.20\\ 
\hline
    \end{tabular}
    }
%     {\footnotesize
% \begin{tablenotes}
% \item[] Data presented as mean $\pm$ standard error of mean
% \end{tablenotes}
% }

\end{table*}

\newpage
\begin{table*}[!h]
 \centering
 \small
 \caption{Results of the fine-tuned BERT$_{large}$ $hits@K$ metric on the folds for each relation in ConceptNet. }
 \label{tab:results_on_each_relation}
 \rotatebox{0}{
    \begin{tabular}{@{}c|ccc|ccc|ccc@{}}
    \hline
    \multicolumn{1}{c|}{\multirow{3}{*}{\textbf{Relations}}} & \multicolumn{9}{c}{$hits@K$} \\ \cline{2-10} 
    \multicolumn{1}{c|}{} & \multicolumn{3}{c|}{Fold 0} & \multicolumn{3}{c|}{Fold 1} & \multicolumn{3}{c}{Fold 2} \\ \cline{2-10} &
    1 & 10 &  \multicolumn{1}{c|}{100} & 1 & 10 &  \multicolumn{1}{c|}{100} & 1 & 10 &  \multicolumn{1}{c}{100} \\ \hline\hline
\multirow{-1}{*}{RelatedTo} & 20.70 & 38.70 & 59.00 & 20.80 & 39.00 & 59.30 & 21.00 & 39.20 & 59.00 \\
HasContext & 40.80 & 76.00 & 95.70 & 40.80 & 76.40 & 95.90 & 41.10 & 76.50 & 95.90 \\
IsA & 42.60 & 69.00 & 86.10 & 42.00 & 68.20 & 85.30 & 42.30 & 68.60 & 85.60 \\
DerivedFrom & 60.80 & 84.70 & 91.30 & 61.60 & 84.60 & 91.30 & 62.30 & 85.00 & 91.60 \\
Synonym & 24.80 & 53.30 & 73.30 & 25.20 & 53.40 & 73.20 & 24.30 & 52.60 & 72.90 \\
FormOf & 80.90 & 90.40 & 93.40 & 80.40 & 90.20 & 93.60 & 80.10 & 90.20 & 93.50 \\
EtymologicallyRelatedTo & 21.70 & 38.80 & 52.90 & 20.20 & 35.80 & 50.00 & 19.70 & 35.10 & 50.10 \\
SimilarTo & 10.90 & 37.90 & 66.00 & 10.50 & 37.00 & 65.50 & 10.00 & 38.40 & 67.10 \\
AtLocation & 21.50 & 58.60 & 87.80 & 20.10 & 57.20 & 88.00 & 20.80 & 55.50 & 86.30 \\
MannerOf & 14.90 & 43.70 & 73.10 & 14.80 & 42.20 & 72.50 & 13.20 & 42.10 & 71.80 \\
PartOf & 9.70 & 37.40 & 67.00 & 10.20 & 38.10 & 66.80 & 10.50 & 40.40 & 68.70 \\
Antonym & 25.60 & 59.60 & 80.50 & 22.60 & 56.00 & 80.60 & 23.10 & 53.60 & 76.70 \\
HasProperty & 14.40 & 46.20 & 76.70 & 15.00 & 42.60 & 76.30 & 13.20 & 40.80 & 77.80 \\
UsedFor & 11.80 & 36.40 & 67.80 & 8.20 & 31.90 & 63.20 & 10.20 & 35.30 & 65.60 \\
DistinctFrom & 18.30 & 57.30 & 82.50 & 16.90 & 59.90 & 80.40 & 19.30 & 61.10 & 81.30 \\
HasPrerequisite & 6.90 & 26.70 & 59.20 & 6.00 & 23.50 & 56.50 & 1.70 & 19.60 & 55.60 \\
HasSubevent & 8.00 & 25.00 & 57.40 & 6.70 & 25.80 & 60.60 & 9.50 & 28.30 & 63.30 \\
Causes & 10.30 & 38.40 & 68.30 & 13.70 & 33.10 & 64.80 & 12.40 & 37.20 & 65.20 \\
HasA & 12.20 & 38.80 & 74.10 & 9.10 & 43.50 & 73.30 & 8.70 & 46.50 & 74.10 \\
InstanceOf & 22.00 & 41.10 & 66.40 & 19.60 & 41.60 & 60.30 & 21.80 & 42.20 & 62.60 \\
CapableOf & 19.50 & 46.80 & 66.30 & 18.50 & 43.40 & 73.70 & 19.70 & 45.20 & 71.60 \\
MotivatedByGoal & 3.40 & 17.50 & 52.50 & 5.10 & 21.30 & 57.90 & 3.20 & 24.70 & 59.70 \\
MadeOf & 28.60 & 62.20 & 87.80 & 25.30 & 66.70 & 88.90 & 20.80 & 51.50 & 78.20 \\
Entails & 8.30 & 26.20 & 59.50 & 3.50 & 23.30 & 62.80 & 4.80 & 25.30 & 60.20 \\
Desires & 20.60 & 33.30 & 65.10 & 3.10 & 20.30 & 39.10 & 13.10 & 24.60 & 49.20 \\
NotHasProperty & 19.60 & 37.00 & 80.40 & 13.70 & 31.40 & 54.90 & 12.00 & 40.00 & 70.00 \\
CreatedBy & 5.70 & 42.90 & 68.60 & 2.60 & 36.80 & 73.70 & 12.10 & 39.40 & 78.80 \\
DefinedAs & 25.00 & 50.00 & 79.20 & 5.60 & 50.00 & 83.30 & 4.80 & 47.60 & 90.50 \\
NotDesires & 0.00 & 4.50 & 31.80 & 14.30 & 23.80 & 33.30 & 13.60 & 27.30 & 45.50 \\
NotCapableOf & 25.00 & 58.30 & 91.70 & 30.80 & 46.20 & 76.90 & 25.00 & 33.30 & 75.00 \\
LocatedNear & 0.00 & 75.00 & 100.00 & 9.10 & 36.40 & 63.60 & 9.10 & 36.40 & 81.80 \\
EtymologicallyDerivedFrom & 50.00 & 50.00 & 50.00 & 16.70 & 16.70 & 16.70 & 22.20 & 22.20 & 22.20\\

\hline
    \end{tabular}
    }
%     {\footnotesize
% \begin{tablenotes}
% \item[] Data presented as mean $\pm$ standard error of mean
% \end{tablenotes}
% }

\end{table*}

\begin{table}[]
\centering
\footnotesize
\begin{threeparttable}
% \small
\caption{\textbf{Comparison of macro average hits@K on pretrained and fune-tuned BERT models. Herin, Avg. indicates average.}}
%The macro avg. equally average the results of all relations, while the micro avg. weighted average the results of the relations according to their portion.}
\label{apx:hits@K_finetune}
\begin{tabular}{c|c|c|cccccccccccc}
\hline
\multirow{3}{*}{\begin{tabular}[c]{@{}c@{}}Train \\ Status\end{tabular}} & \multirow{3}{*}{Model} & \multirow{3}{*}{\begin{tabular}[c]{@{}c@{}}Avg. \\ Type\end{tabular}} & \multicolumn{12}{c}{Hits@K} \\ \cline{4-15} 
 & & & \multicolumn{4}{c|}{1} & \multicolumn{4}{c|}{10} & \multicolumn{4}{c}{100} \\ \cline{4-15} 
 & & & Fold 0 & Fold 1 & Fold 2 & \multicolumn{1}{c|}{Avg.*} & Fold 0 & Fold 1 & Fold 2 & \multicolumn{1}{c|}{Avg.} & Fold 0 & Fold 1 & Fold 2 & Avg. \\ \hline\hline
\multirow{4}{*}{Pretrained} & \multirow{2}{*}{BERT$_{base}$} & Micro & 6.67 & 5.05 & 6.19 & \multicolumn{1}{c|}{5.97} & 17.72 & 16.31 & 17.38 & \multicolumn{1}{c|}{17.14} & 36.13 & 33.78 & 34.80 & 34.90 \\
 & & Macro & 7.93 & 7.74 & 7.79 & \multicolumn{1}{c|}{7.82} & 19.69 & 19.65 & 19.62 & \multicolumn{1}{c|}{19.65} & 37.29 & 37.25 & 37.21 & 37.25 \\ \cline{2-15} 
 & \multirow{2}{*}{BERT$_{large}$} & Micro & 8.00 & 5.51 & 7.31 & \multicolumn{1}{c|}{6.94} & 20.03 & 17.65 & 18.27 & \multicolumn{1}{c|}{18.65} & 38.17 &36.31 & 36.07 & 36.85 \\
 & & Macro & 7.27 & 7.16 & 7.24 & \multicolumn{1}{c|}{7.22} & 19.31 & 19.27 & 19.19 & \multicolumn{1}{c|}{19.26} & 37.09 & 36.90 & 36.88 & 36.96\\ \hline
\multirow{4}{*}{Fine-tuned} & \multirow{2}{*}{BERT$_{base}$} & Micro & 17.73 & 16.40 & 16.25 & \multicolumn{1}{c|}{16.79} & 42.52 & 40.67 & 40.39 & \multicolumn{1}{c|}{41.19} & 66.87 & 62.82 & 63.80 & 64.50 \\
 & & Macro & 29.58 & 29.60 & 29.77 & \multicolumn{1}{c|}{29.65} & 52.20 & 52.22 & 52.16 & \multicolumn{1}{c|}{52.19} & 70.88 & 70.97 & 70.89 & 70.91 \\ \cline{2-15} 
 & \multirow{2}{*}{BERT$_{large}$} & Micro & 21.39 & 19.14 & 19.55 & \multicolumn{1}{c|}{20.03} & 46.93 & 43.63 & 43.93 & \multicolumn{1}{c|}{44.83} & 72.23 & 68.20 & 70.23 & 70.22 \\
 & & Macro & 32.41 & 32.35 & 32.53 & \multicolumn{1}{c|}{32.43} & 55.86 & 55.79 & 55.96 & \multicolumn{1}{c|}{55.87} & 74.28 & 74.26 & 74.23 & 74.26 \\ \hline
\end{tabular}
{\footnotesize
\begin{tablenotes}
\item[*] Average of the performances of Fold 1, 2 and 3
%\item[\textdagger] Questions succeed or failed to predict by all experimental models commonly
% \item[\textdaggerdbl] Symbol 1
% \item[\S] Symbol 1  
% \item[$\|$] Symbol 1
% \item[$\!$\#] Symbol 1          
\end{tablenotes}
}
% {\small
% \begin{tablenotes}
% \item[a] testing
% \end{tablenotes}
% }
\end{threeparttable}
\end{table}

\newpage

\begin{table}[h!]
\centering
\begin{threeparttable}
% \small
\caption{\textbf{Results of missed prediction ratio at top K predictions of fine-tuned BERT models between the opposite relations.}}
%The macro avg. equally average the results of all relations, while the micro avg. weighted average the results of the relations according to their portion.}
\label{apx:miss@K_finetune}
\begin{tabular}{c|c|ccc|ccc|ccc}
\hline
\multicolumn{2}{c|}{} & \multicolumn{9}{c}{{Miss@K*}} \\
\cline{3-11}
\multicolumn{2}{c|}{} & \multicolumn{3}{c|}{1} & \multicolumn{3}{c|}{10} & \multicolumn{3}{c}{100} \\
\cline{3-11}
\multicolumn{2}{c|}{\multirow{-3}{*}{Model}} & Fold 0 & Fold 1 & Fold 2 & Fold 0 & Fold 1 & Fold 2 & Fold 0 & Fold 1 & Fold 2 \\
\hline \hline
 & BERT$_{base}$ & 3.33 & 2.89 & 3.87 & 29.57 & 28.48 & 31.22 & 57.35 & 57.50 & 54.98 \\
\multirow{-2}{*}{Synonym / Antonym} & BERT$_{large}$ & 4.42 & 4.21 & 4.36 & 29.53 & 29.63 & 30.35 & 56.34 & 58.43 & 58.21 \\
\hline
 & BERT$_{base}$ & 17.16 & 13.63 & 13.63 & 37.56 & 35.20 & 35.24 & 58.91 & 56.75 & 57.78 \\
\multirow{-2}{*}{Antonym / Synonym} & BERT$_{large}$ & 8.69 & 8.57 & 8.20 & 31.50 & 27.40 & 28.24 & 55.26 & 49.87 & 52.70 \\
\hline

\end{tabular}
{\footnotesize
\begin{tablenotes}
\item[*] Miss@K = The ratio of words that are considered right when graded with the opposite relation (undesirable) in top K predictions.
\end{tablenotes}
}
% {\small
% \begin{tablenotes}
% \item[a] testing
% \end{tablenotes}
% }
\end{threeparttable}
\end{table}

\newpage

\section{Comparative Studies with respect to the Lexical Similarities of a Question and a Context} 
\label{apx:difficulty_word_overlap}
\begin{figure}[!ht]
  \adjustbox{minipage=2em,raise=-\height}{\subcaption{} \label{fig:has_answer}}%
  %\raisebox{-\height}{\includegraphics[width=.45\linewidth]{images/has_ans_yaxis.pdf}}
  \raisebox{-\height}{\includegraphics[width=.45\linewidth]{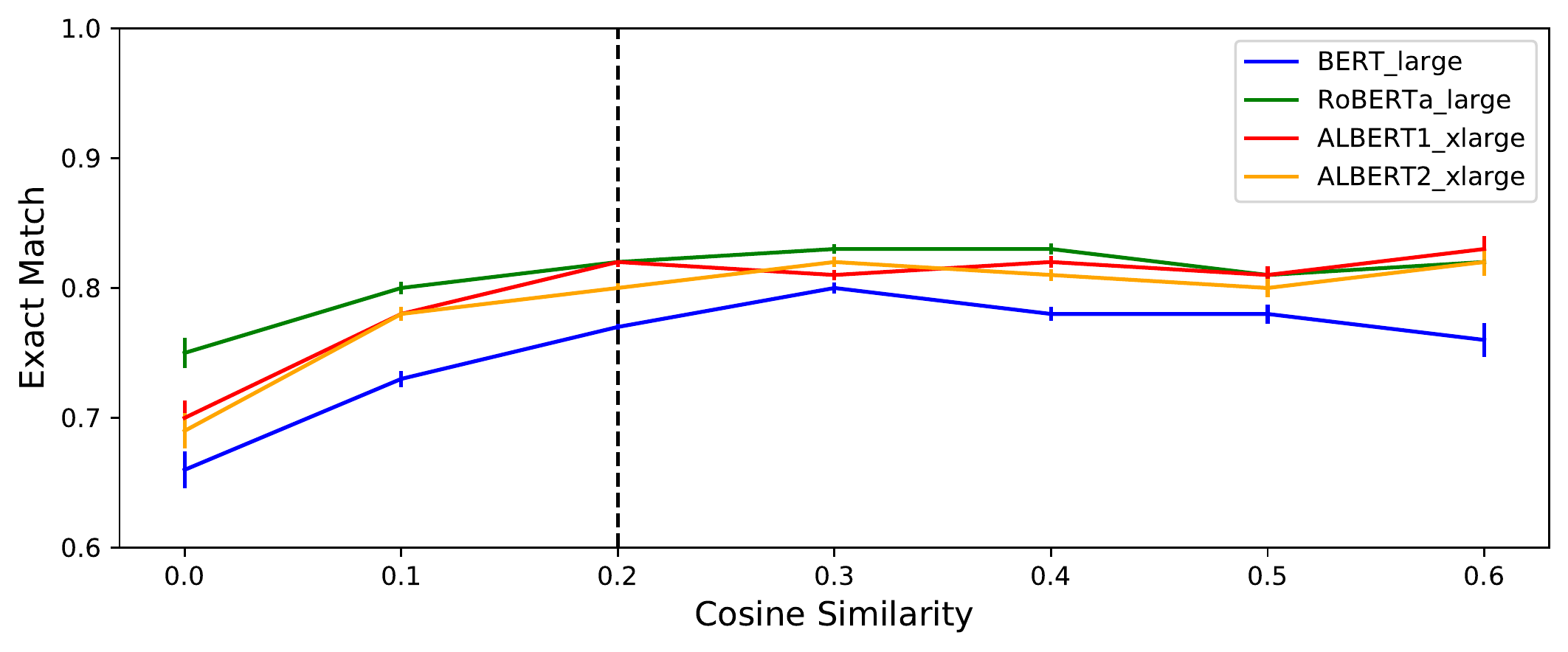}}
% 
%   \newline
  %\adjustbox{minipage=2em,raise=-\height}{\subcaption{} \label{fig:no_answer}}%
  %\raisebox{-\height}{\includegraphics[width=.45\linewidth]{images/no_ans_yaxis.pdf}}
  %\raisebox{-\height}{\includegraphics[width=.45\linewidth]{images/no_ans_full.pdf}}
  \adjustbox{minipage=2em,raise=-\height}{\subcaption{} \label{fig:record}}%
  %\raisebox{-\height}{\includegraphics[width=.45\linewidth]{images/no_ans_yaxis.pdf}}
  \raisebox{-\height}{\includegraphics[width=.45\linewidth]{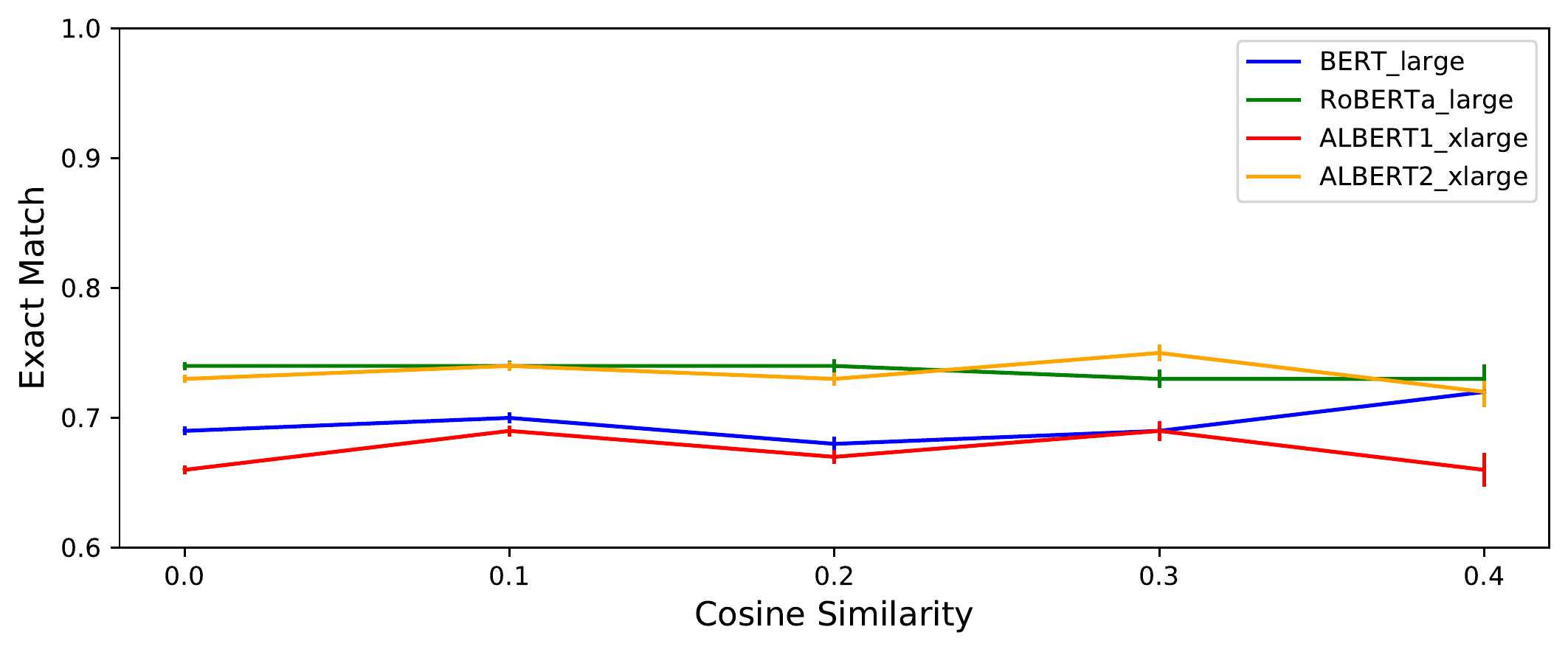}}
  
  \caption{\textbf{Results on the word overlapping rate and the performance of RC models}: \textbf{(a)} shows results of the \textit{has answer} questions of SQuAD 2.0 and \textbf{(b)} shows results of ReCoRD. 
  %X-axis indicates the cosine similarity of context and question. Y-axis denotes the performances of models. 
  \revised{X-axis indicates the cosine similarity of a context and a question that is a measure of normalized lexical affinity between the context and the question. The questions whose similarities over than specific thesholds (0.6 for SQuAD and 0.4 for ReCoRD) are ignored since they comprise a trivial portion (less than 1\%). Error bars of the data indicate the standard deviation.}}  %EM, accuracy
  \label{fig:word_overlapping_test}
\end{figure}

This section analyzes the correlation between the lexical overlap of a question and a context, and performances on the question. First, we measure the degree of the lexical variation based on the similarity between a question and a context. Specifically, a low similarity between a question and a context indicates a high lexical variation, and vice versa. For this, we use cosine similarity between term frequency-inverse document frequency (TF-IDF) term-weighted uni-gram bag-of-words vectors \cite{manning2010introduction}. 
Then, we measure the difficulty level of a question based on the performance of the MNLM-based RC models. \revised{Herein, to control the experimental environment, we only consider \textit{has answer} questions in SQuAD as analysis targets. This is because there is no \textit{no answer} question in the case of ReCoRD.
 Figure~\ref{fig:word_overlapping_test} shows that \textit{has answer} questions tend to be more difficult with high lexical variations in SQuAD. Moreover, examples of \textit{has answer} questions are shown in Table~\ref{tab:example_of_questions}. In the first example, the question and the context are similar, thus the clues are explicitly provided in the context. On the other hand, the second example requires additional clues. For example, since a keyword in the question `country' does not appear in the context, it is necessary to infer that Russia is a country. Also, it is necessary to know that `spread the disease' can be paraphrased to`receive the disease'. Our analysis indicates that the lexical discrepancy between a question and a context requires implicit clues, such as commonsense knowledge, affecting the difficulty level of the question. Otherwise, in ReCoRD, it is difficult to observe the relatedness between lexical overlapping and the performance. From these results, we can see that in the case of SQuAD, lexical discrepancy greatly affects the performance, so that the location of the correct answer can be inferred through the mapping of key words. Herein, we analyze \textit{has answer} questions which have a low lexical similarity ($similarity < 0.2$) since these are relatively difficult and account for approximately 20\% of the entire \textit{has answer} questions (21.47\% = 1,273 out of 5,928).}
 
 \revised{On the other hand, in ReCoRD, the factors determining the difficulty of the problem act more complexly, which means that it is difficult to infer the correct answer simply by mapping the keywords of key words. Therefore, we randomly sample 200 questions from the development set for further analysis.}
 
\begin{table*}[!ht]
 \centering
 \begin{threeparttable}
  \caption{\textbf{Examples of easy questions and hard questions on the \textit{has answer} questions of SQuAD.}}
  \label{tab:example_of_questions}
\begin{tabular}{c|m{0.225\columnwidth}|m{0.10\columnwidth}|m{0.475\columnwidth}}\hline
\begin{tabular}[c]{c}Similarity\tnote{*}\end{tabular} &  \multicolumn{1}{c|}{Question} & \multicolumn{1}{c|}{Answer} & \multicolumn{1}{c}{Context} \\\hline\hline
$>0.6$ & In what year did Savery patent his steam pump?  & 1698  & 	... In 1698 Thomas Savery patented a steam pump that used steam in direct contact with the water being pumped. ... \\\hline
$<0.1$& Which country was the last to receive the disease?  &northwestern Russia & From Italy, the disease spread northwest across Europe, ... Finally it spread to northwestern Russia in 1351. ...\\\hline
\end{tabular}
{\footnotesize
\begin{tablenotes}
\item[*] Cosine similarity between TF-IDF term weighted uni-gram vectors of the question and the context
%\item[\textdagger] Questions succeed or failed to predict by all experimental models commonly
% \item[\textdaggerdbl] Symbol 1
% \item[\S] Symbol 1  
% \item[$\|$] Symbol 1
% \item[$\!$\#] Symbol 1          
\end{tablenotes}
}
\end{threeparttable}
\end{table*}

\newpage
\section{Details on the Reading Comprehension Question Types}
\label{apx:details_on_RC_question_types}
% \begin{table*}[!ht]
% \centering
%   \caption{Results of the $hits@K$ metric for each relations of ConceptNet. }
%   \label{tab:results_on_each_relation}
% \begin{tabular}{c|c|c}

% \hlinee
\begin{table*}[!ht]
 \centering
  \caption{Examples and descriptions for the question type of the \textit{has answer} questions. The main clues for the categorization of the questions are colored. }
  \label{tab:results_on_each_relation}
\begin{tabular}{c|m{0.30\columnwidth}|m{0.40\columnwidth}}\hline
Question Types &  \multicolumn{1}{c|}{Description} & \multicolumn{1}{c}{Example} \\\hline\hline
Synonymy & There is a clear correspondence between question and context.& \begin{tabular}{@{}p{6.5cm}@{}}\textbf{Question}: Which entity is the \textbf{\textcolor{red}{secondary}} legislative body?\\\textbf{Context}: ... The \textbf{\textcolor{blue}{second main}} legislative body is the Council, which is composed of different ministers of the member states. ...\end{tabular}  \\\hline
\begin{tabular}[c]{@{}c@{}}Commonsense\\ knowledge\end{tabular} & Commonsense knowledge is required to solve the question. & \begin{tabular}{@{}p{6.5cm}@{}}\textbf{Question}: Where is the \textcolor{red}{\textbf{Asian}} influence strongest in Victoria?\\\textbf{Context}: ... Many \textcolor{blue}{\textbf{Chinese}} miners worked in Victoria, and their legacy is particularly strong in Bendigo and its environs. ...\end{tabular}\\\hline
No semantic variation & There is no semantic variation such as synonymy or commonsense knowledge. & \begin{tabular}{@{}p{6.5cm}@{}}\textbf{Question}: Who are the \textbf{\textcolor{red}{un-elected subordinates of member state governments}}?\\\textbf{Context}: ... This means Commissioners are, through the appointment process, the \textbf{\textcolor{blue}{unelected subordinates of member state governments}}. ...\end{tabular} \\\hline
Multi-sentence reasoning & Hints for solving questions are shattered in multiple sentences.  & \begin{tabular}{@{}p{6.5cm}@{}}\textbf{Question}: Why did \textcolor{red}{\textbf{France}} choose to give up continental lands?\\\textbf{Context}: ...  \textcolor{blue}{\textbf{France}} chose to cede the former, ... \textcolor{blue}{\textbf{They}} viewed the economic value of the Caribbean islands' sugar cane ...\end{tabular} \\\hline

Typo & There exist typing errors in the question or context. & \begin{tabular}{@{}p{6.5cm}@{}}\textbf{Question}: What kind of measurements define \textbf{\textcolor{red}{accelerlations}}?\\\textbf{Context}... \textbf{\textcolor{blue}{Accelerations}} can be defined through kinematic measurements. ...\end{tabular}\\\hline
Others & The labeled answer is incorrect. &  \begin{tabular}{@{}p{6.5cm}@{}}\textbf{Question}: Who \textcolor{red}{\textbf{won the battle}} of Lake George?\\\textbf{Context}: ...  The \textcolor{blue}{\textbf{battle ended inconclusively}}, with both sides withdrawing from the field.  ...\end{tabular} \\\hline
\end{tabular}
\end{table*}

\newpage
\section{Supplementary Results on Section~\ref{sec:frequency}}
\label{apx:frequency_performance}
\begin{figure*}[!ht]

\centering
\begin{subfigure}[b]{0.45\textwidth}
\includegraphics[width=\linewidth]{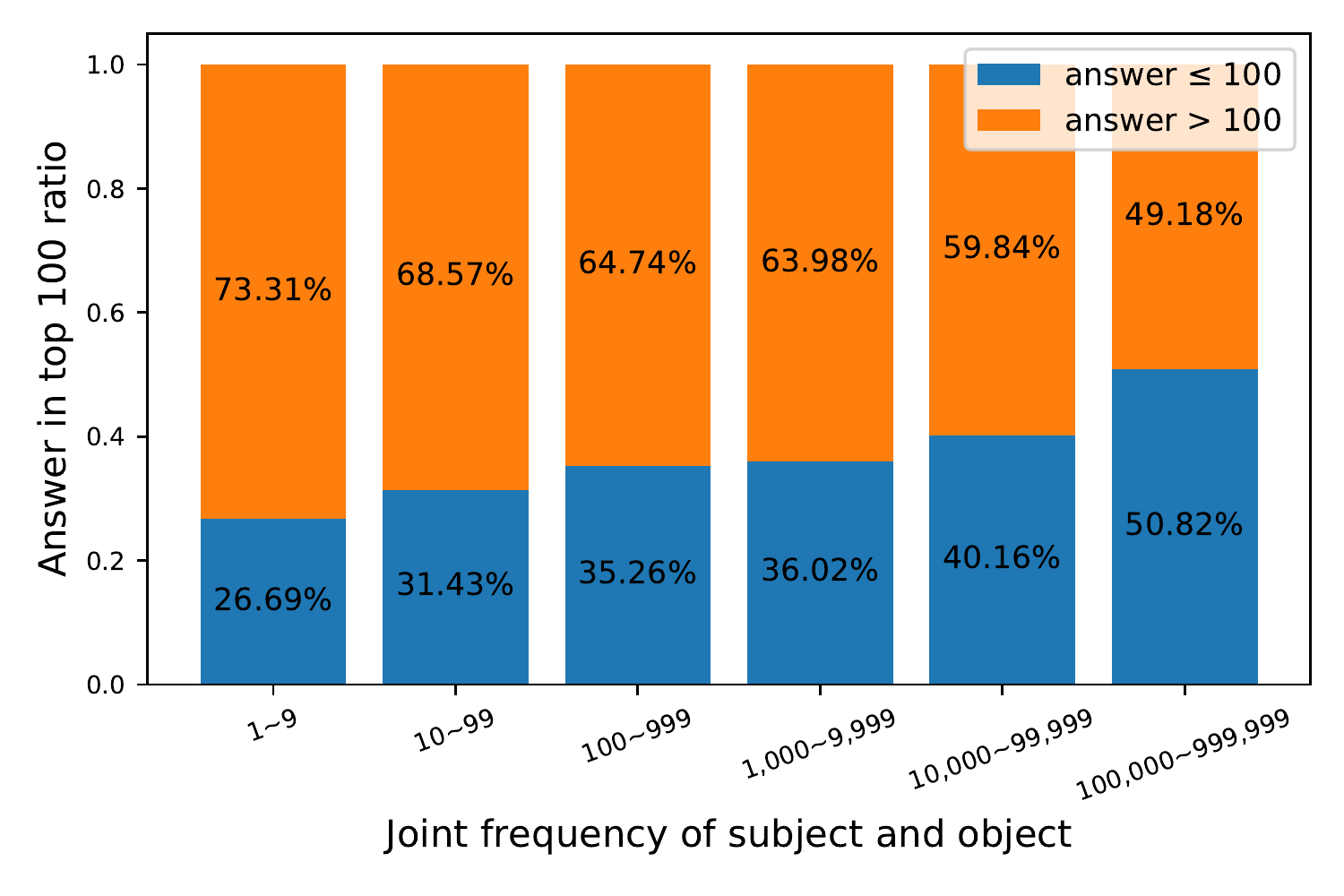}
\caption{BERT$_{base}$}
\end{subfigure}
\begin{subfigure}[b]{0.45\textwidth}
\includegraphics[width=\linewidth]{images/BERT_large_entity_pair_answer_ratio_templates.pdf}
\caption{BERT$_{large}$}
\end{subfigure}
% \begin{subfigure}[b]{0.45\textwidth}
% \includegraphics[width=\linewidth]{images/high_diff_color.pdf}
% \caption{`High'}
% \end{subfigure}
% \newline
\begin{subfigure}[b]{0.45\textwidth}
\includegraphics[width=\linewidth]{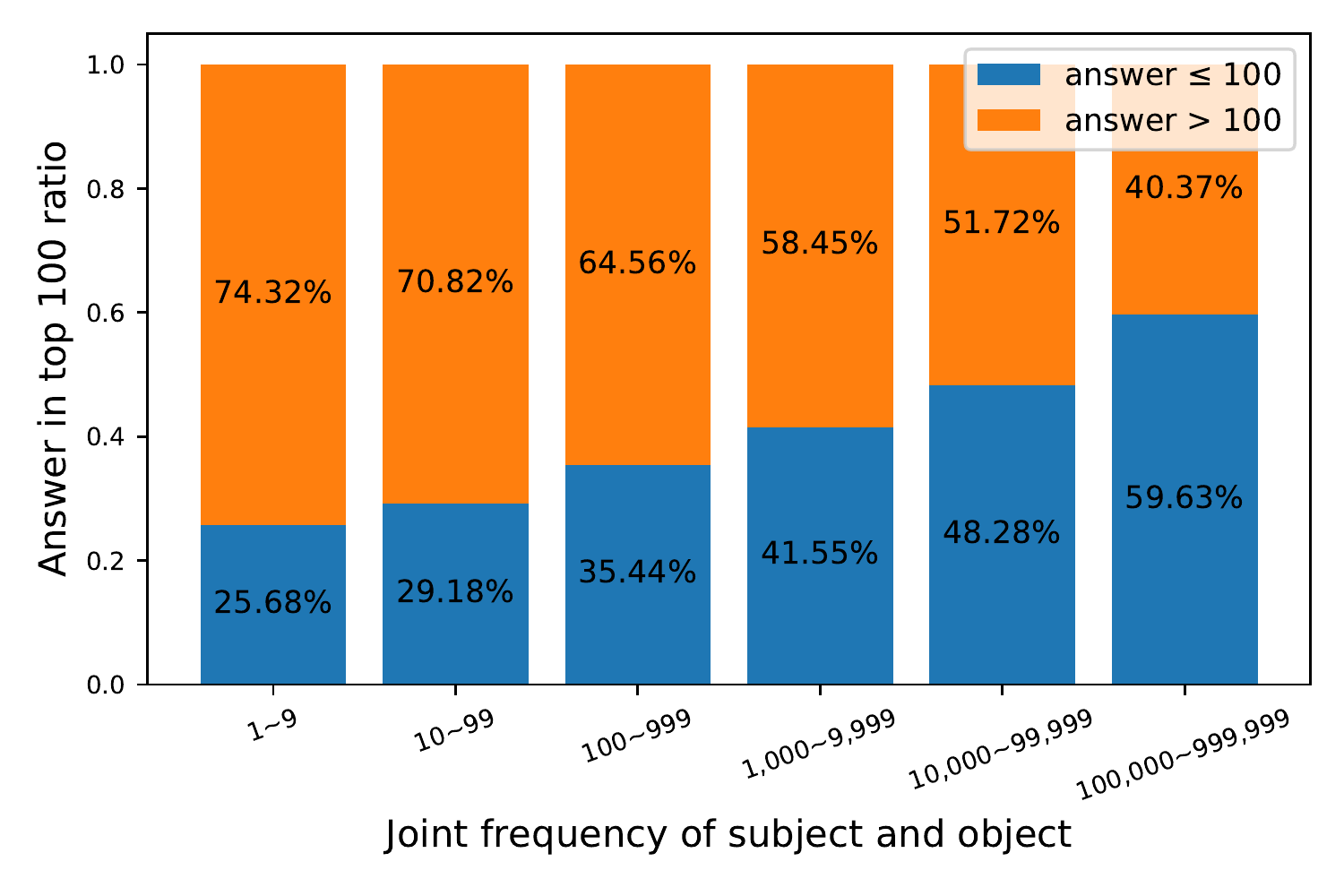}
\caption{ALBERT1$_{base}$}
\end{subfigure}
\begin{subfigure}[b]{0.45\textwidth}
\includegraphics[width=\linewidth]{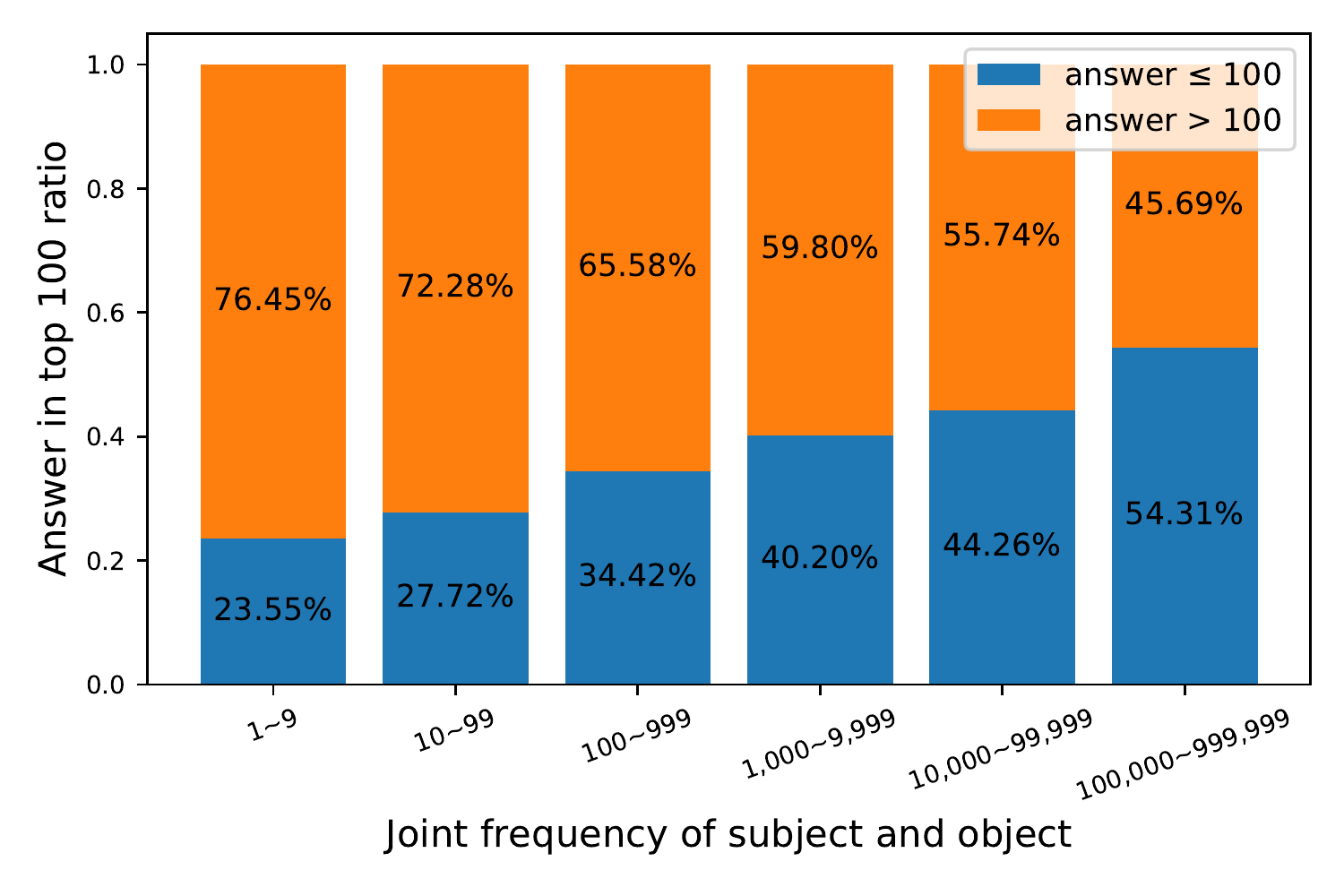}
\caption{ALBERT1$_{large}$}
\end{subfigure}
\begin{subfigure}[b]{0.45\textwidth}
\includegraphics[width=\linewidth]{images/ALBERT_xlarge_entity_pair_answer_ratio_templates.pdf}
\caption{ALBERT1$_{xlarge}$}
\end{subfigure}
\caption{\textbf{The proportion of predicted objects are in top 100 predictions for each model:} \textbf{(a)} BERT$_{base}$, \textbf{(b)} BERT$_{large}$, \textbf{(c)} ALBERT1$_{base}$, \textbf{(d)} ALBERT1$_{large}$, \textbf{(e)} ALBERT1$_{xlarge}$. X-axis indicates frequency with which subject and object entities are observed together. Y-axis is the proportion of which the answer object can be found in top 100 model's prediction for each frequency section. Blue bars indicate that the answer is in the top 100 predictions, while orange bars mean that the answer is not in the top 100 predictions. The results show that joint frequency of subject and object apparently affects the knowledge probing performance.}
\label{fig:bar_plot_of_joint_freq_and_preformance_full}
\end{figure*}

\newpage

\begin{figure*}[!ht]
\centering
\begin{subfigure}[b]{0.45\textwidth}
\includegraphics[width=\linewidth]{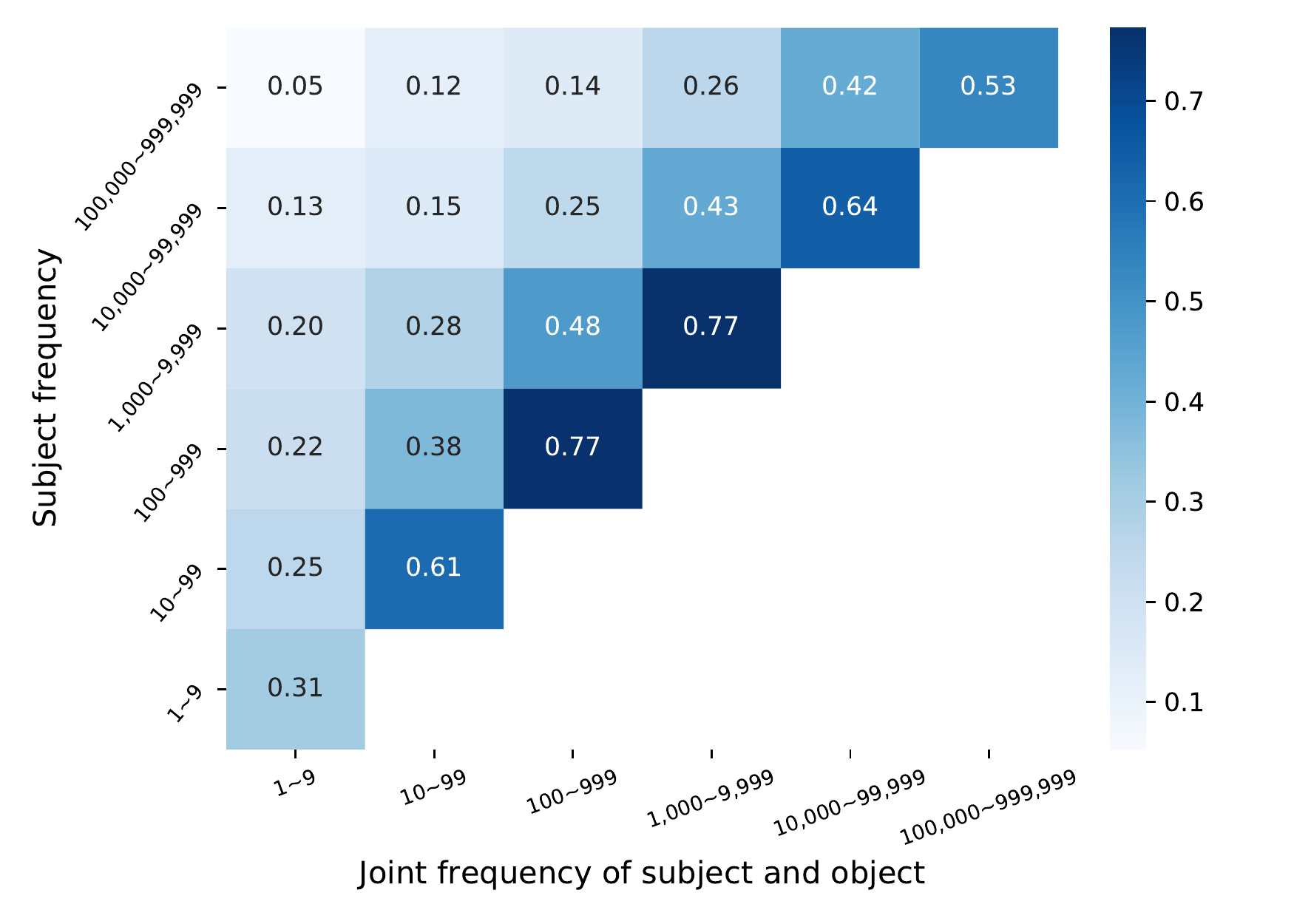}
\caption{BERT$_{base}$}
\end{subfigure}
\begin{subfigure}[b]{0.45\textwidth}
\includegraphics[width=\linewidth]{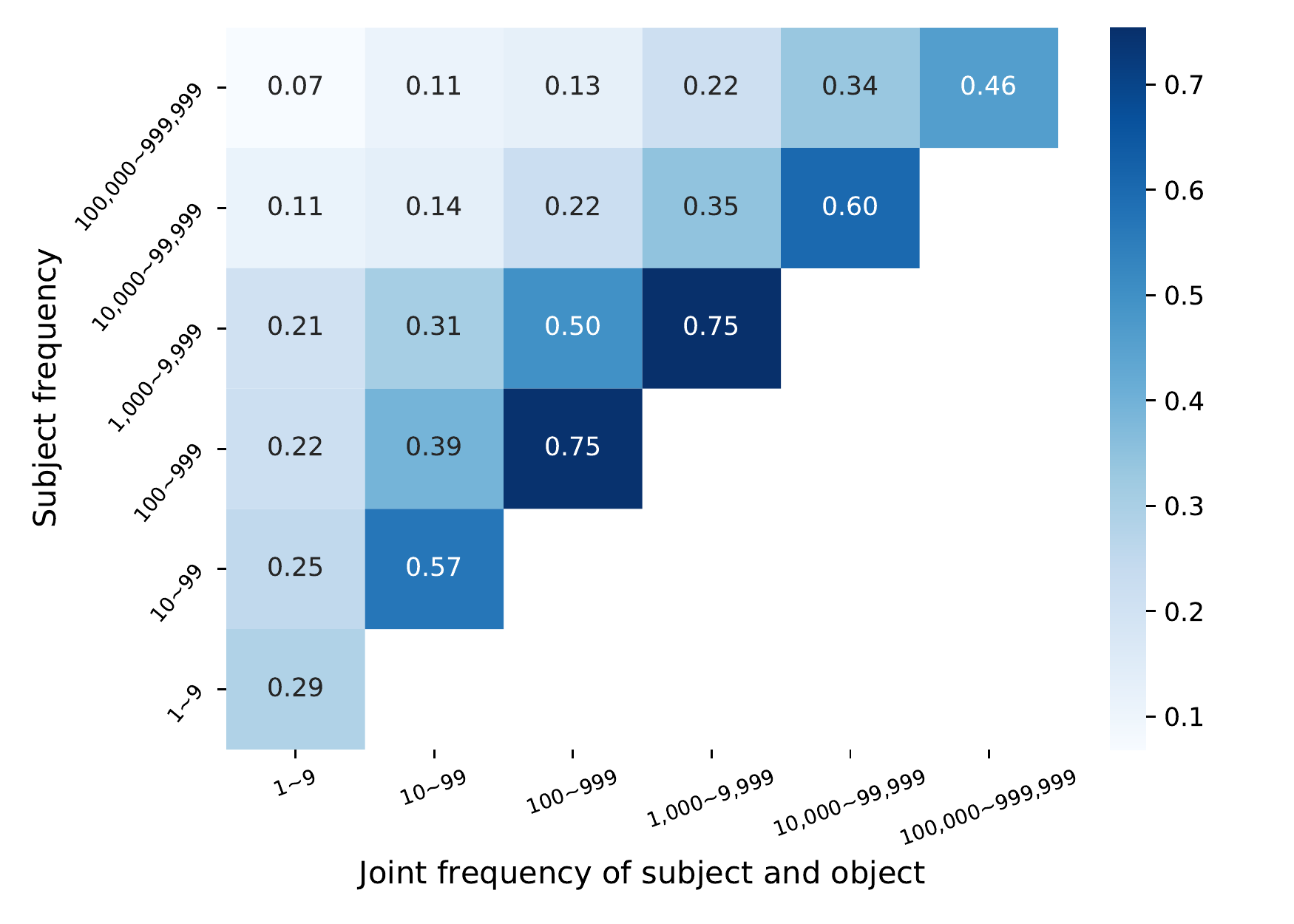}
\caption{BERT$_{large}$}
\end{subfigure}
% \begin{subfigure}[b]{0.45\textwidth}
% \includegraphics[width=\linewidth]{images/high_diff_color.pdf}
% \caption{`High'}
% \end{subfigure}
% \newline
\begin{subfigure}[b]{0.45\textwidth}
\includegraphics[width=\linewidth]{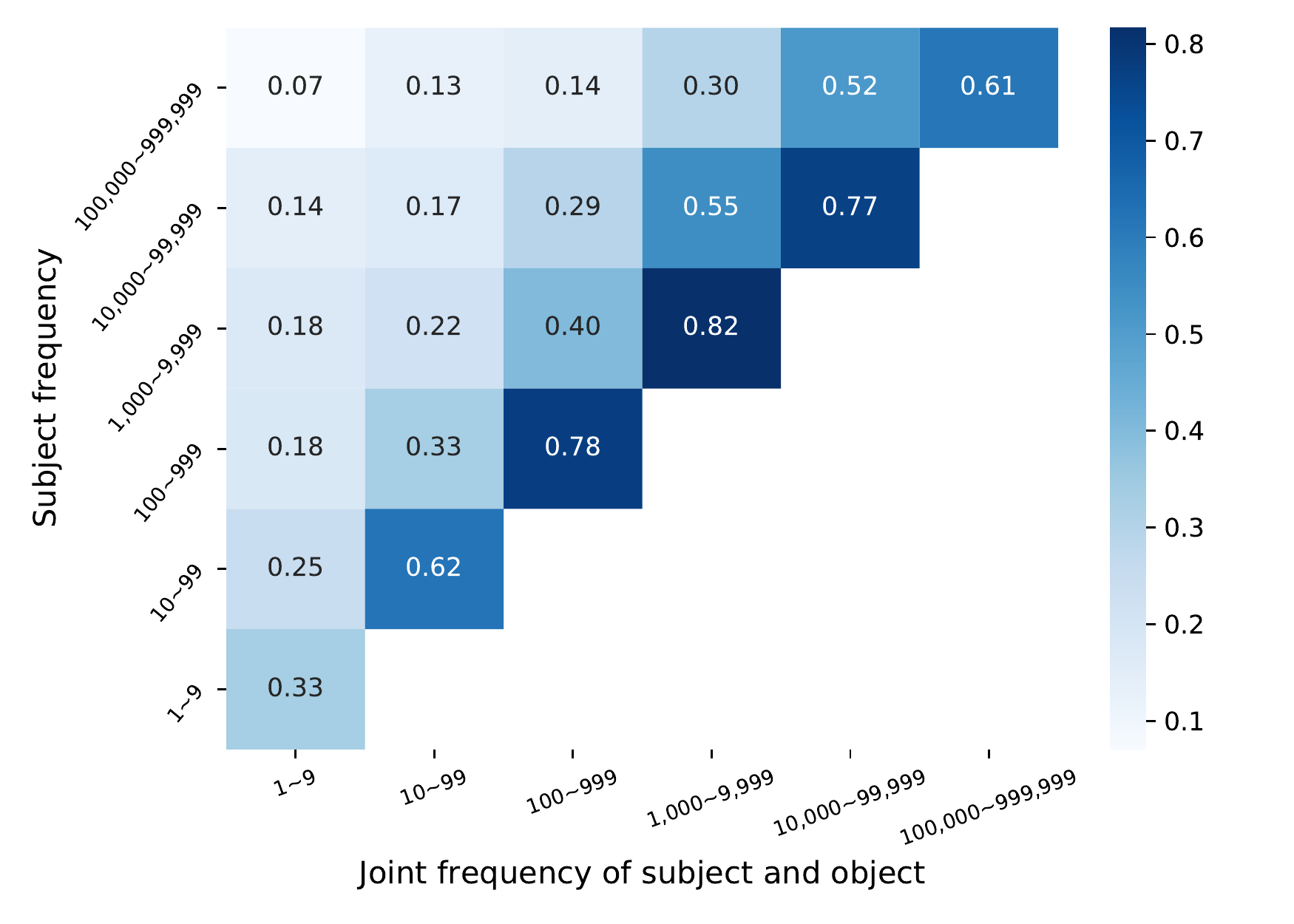}
\caption{ALBERT1$_{base}$}
\end{subfigure}
\begin{subfigure}[b]{0.45\textwidth}
\includegraphics[width=\linewidth]{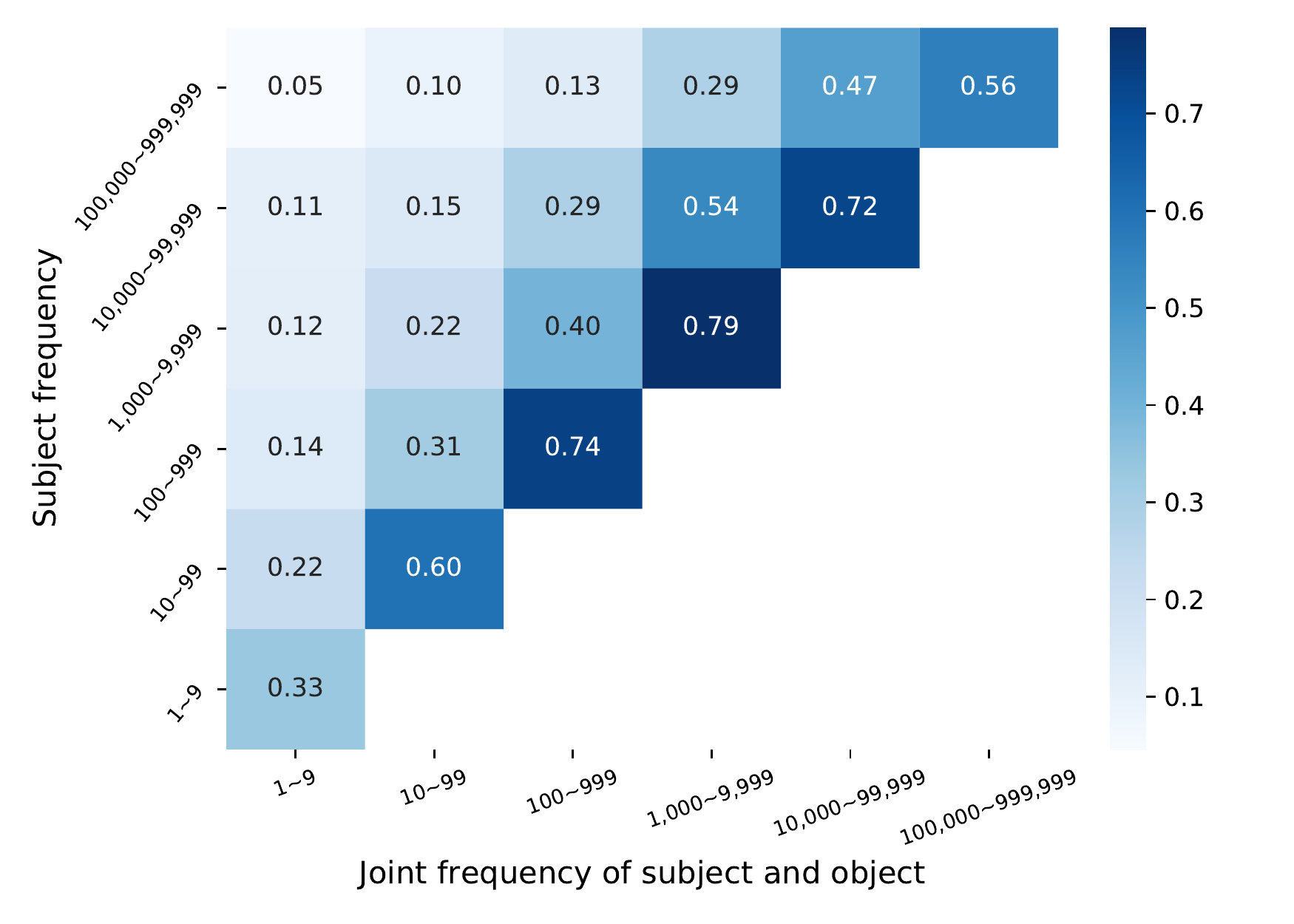}
\caption{ALBERT1$_{large}$}
\end{subfigure}
\begin{subfigure}[b]{0.45\textwidth}
\includegraphics[width=\linewidth]{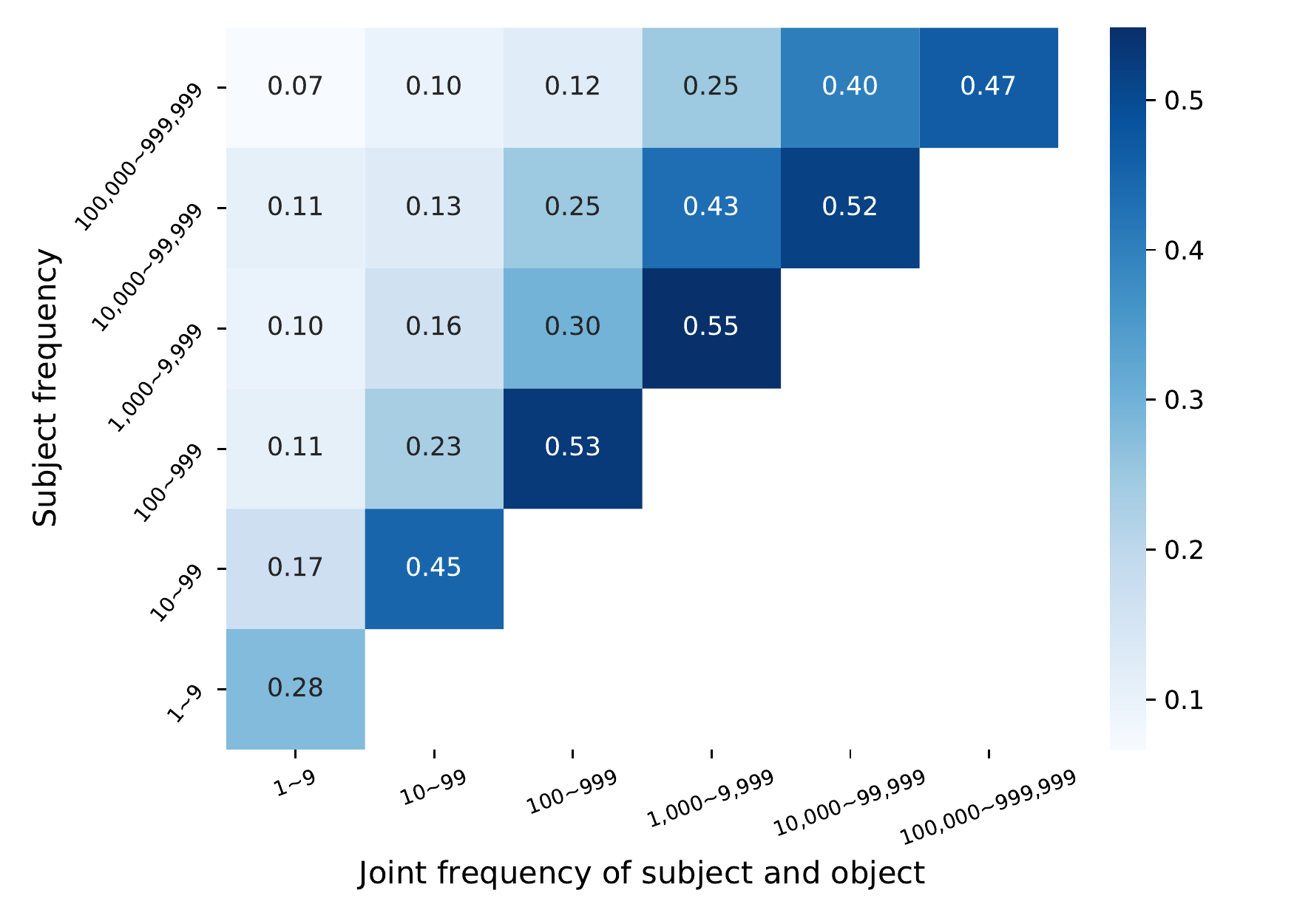}
\caption{ALBERT1$_{xlarge}$}
\end{subfigure}
\caption{\textbf{Correlations of the joint frequency of subject and object, and frequency of subject:} \textbf{(a)} BERT$_{base}$, \textbf{(b)} BERT$_{large}$, \textbf{(c)} ALBERT1$_{base}$, \textbf{(d)} ALBERT1$_{large}$, \textbf{(e)} ALBERT1$_{xlarge}$. X-axis indicates frequency with which subject and object entities are observed together. Y-axis intends the frequency of the subject. Values of heatmap means the proportion of which the answer object can be found in top 100 model’s prediction for each frequency grid.
The results show that the performance increases as the subject observation frequency is lower in the same joint frequency section of subject and object. }

\label{fig:frquency_performance_heat_map_full}
\end{figure*}

\begin{figure*}[!ht]
\centering
\begin{subfigure}[b]{0.45\textwidth}
\includegraphics[width=\linewidth]{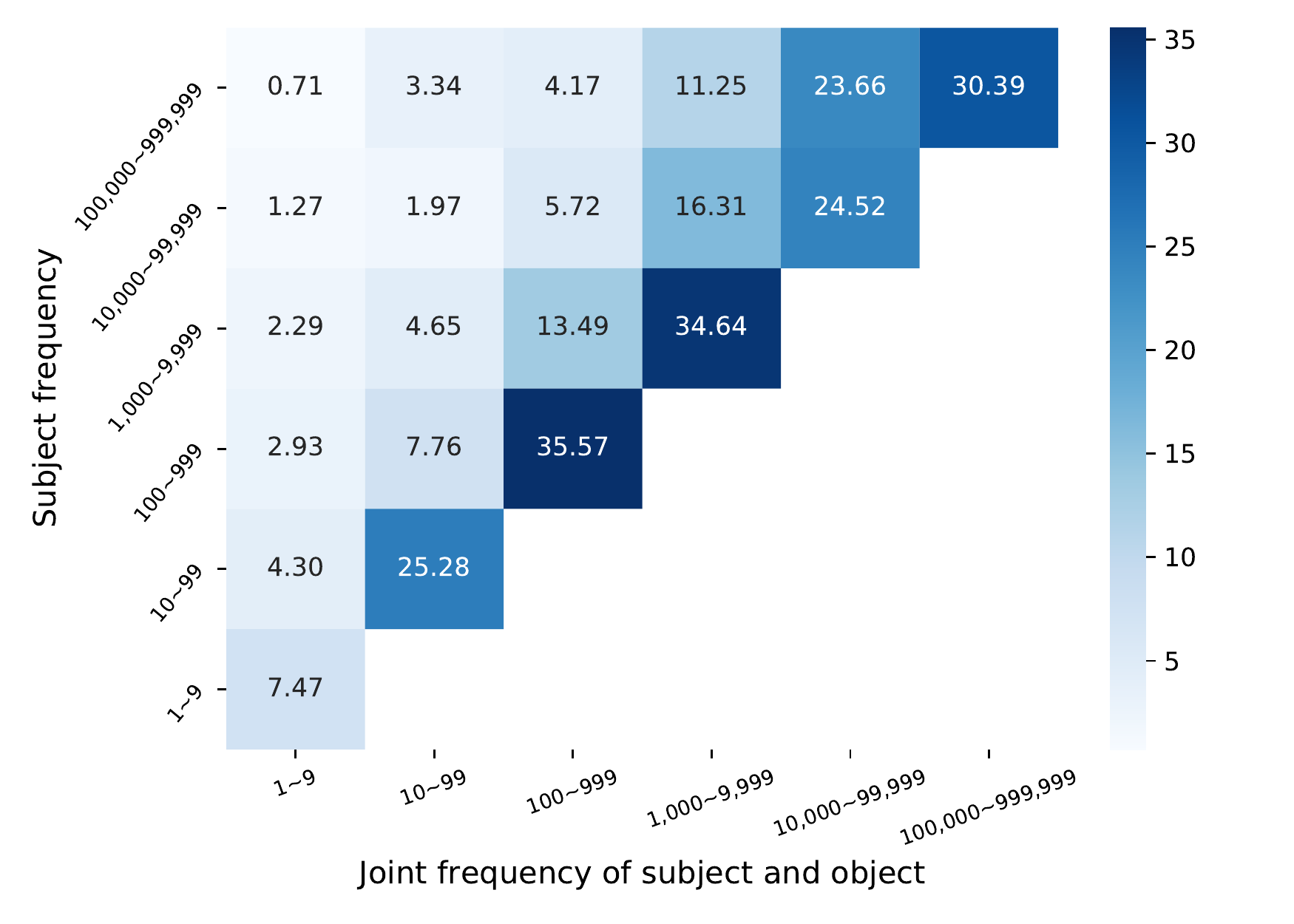}
\caption{BERT$_{base}$}
\end{subfigure}
\begin{subfigure}[b]{0.45\textwidth}
\includegraphics[width=\linewidth]{images/BERT_large_dynamic_conditional_heatmap_templates.pdf}
\caption{BERT$_{large}$}
\end{subfigure}
% \begin{subfigure}[b]{0.45\textwidth}
% \includegraphics[width=\linewidth]{images/high_diff_color.pdf}
% \caption{`High'}
% \end{subfigure}
% \newline
\begin{subfigure}[b]{0.45\textwidth}
\includegraphics[width=\linewidth]{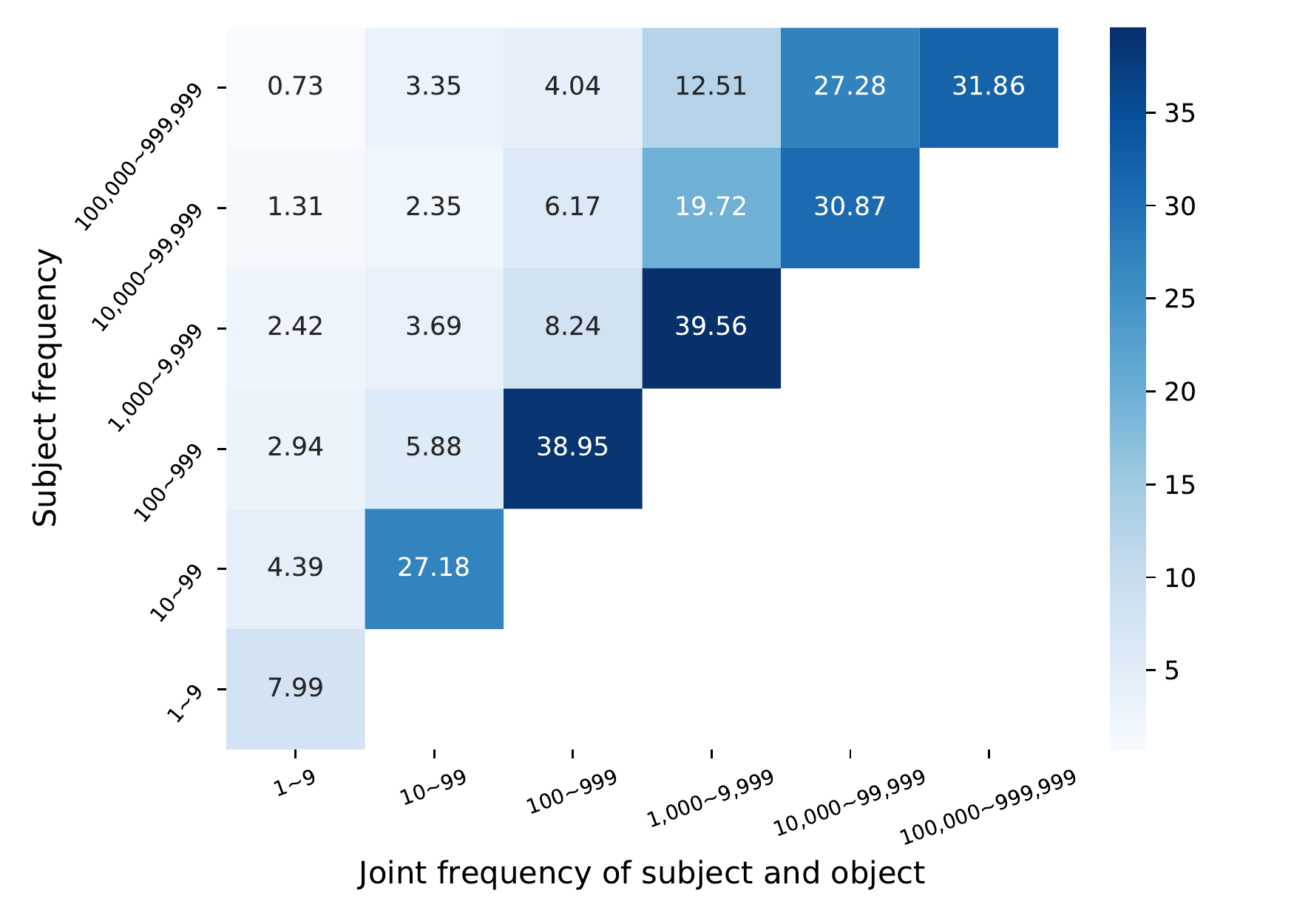}
\caption{ALBERT1$_{base}$}
\end{subfigure}
\begin{subfigure}[b]{0.45\textwidth}
\includegraphics[width=\linewidth]{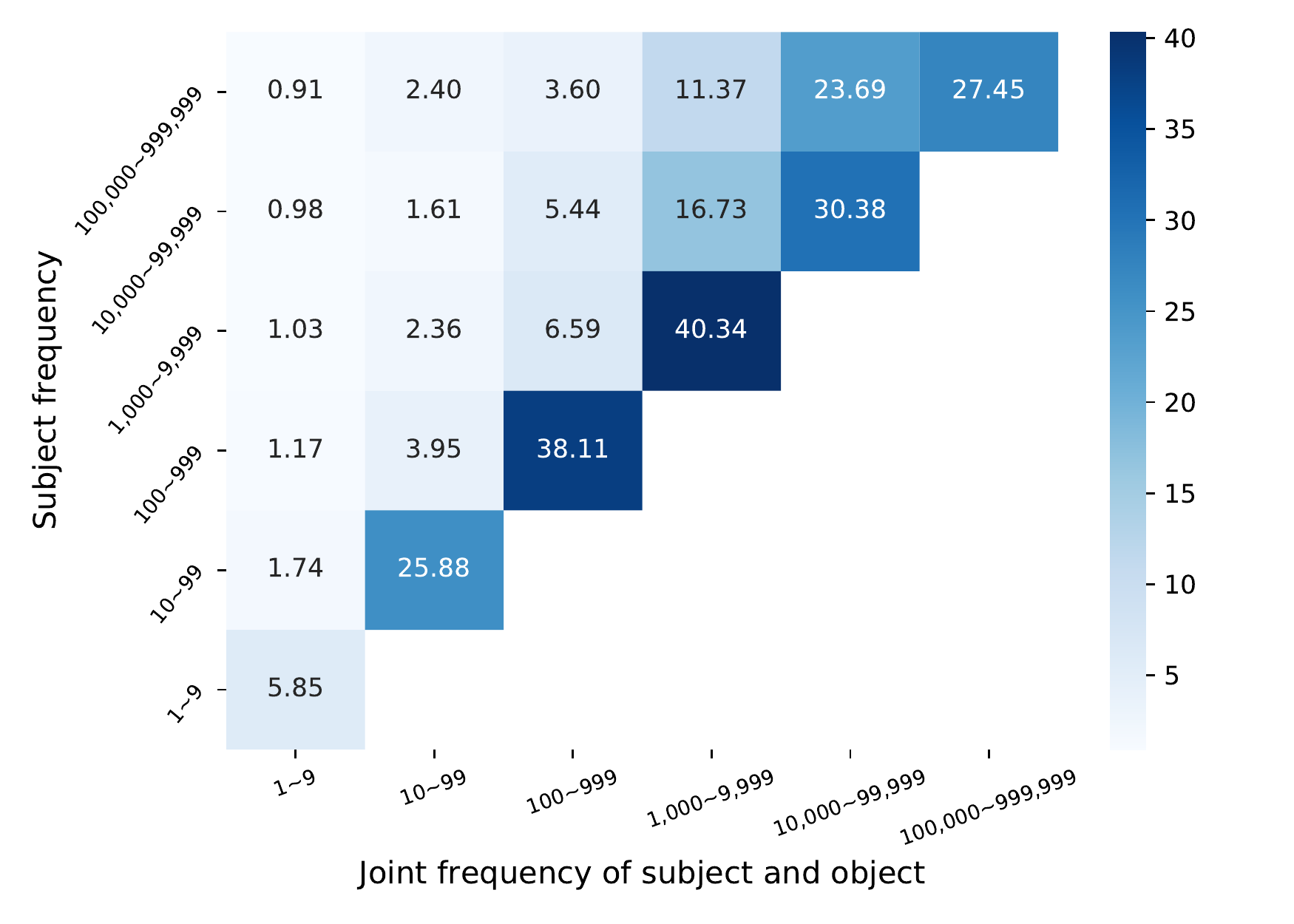}
\caption{ALBERT1$_{large}$}
\end{subfigure}
\begin{subfigure}[b]{0.45\textwidth}
\includegraphics[width=\linewidth]{images/ALBERT_xlarge_dynamic_conditional_heatmap_templates.pdf}
\caption{ALBERT1$_{xlarge}$}
\end{subfigure}
\caption{\textbf{Correlations of the joint frequency of subject and object, and frequency of subject:} \textbf{(a)} BERT$_{base}$, \textbf{(b)} BERT$_{large}$, \textbf{(c)} ALBERT1$_{base}$, \textbf{(d)} ALBERT1$_{large}$, \textbf{(e)} ALBERT1$_{xlarge}$. X-axis indicates frequency with which subject and object entities are observed together. Y-axis intends the frequency of the subject. Values of heatmap means the proportion of which the answer object can be found in top k model’s prediction for each frequency grid. Herein, the `k' indicates the number of objects of a given subject and relation pair.
The results show that the performance increases as the subject observation frequency is lower in the same joint frequency section of subject and object. }

\label{fig:frquency_performance_heat_map_full_dynamic}
\end{figure*}

\newpage
\section{Details on the Integrating External Commonsense Repository}
\label{apx:details_on_the_integrating_external_commonsense_repository_test}
\revised{This section describes the detail procedure of Section~\ref{sec:complement_NLMs_with_knowledge}. First, we initialize knowledge integrated question $Q^*$ and context $C^*$, same as input question and context (line 1, Algorithm 1). For target knowledge triples, if $Q^*$ and $C^*$ are in ReCoRD, we transfer a triple in to the sentence form via designated relation template. Then, the triple sentence is attached to the end of context with the ``@highlight'' term.
Otherwise, if a subject word is in the question, we integrate the knowledge into the question text (line 9-10, Algorithm 1). Otherwise, if a subject word is in the context, we incorporate the knowledge into the context text (line 11-12, Algorithm 1). Specifically, we find the first occurrence index of subject word of the target knowledge triple in the input text. Then, we augment the word with relational templates followed by the object word.}
%Table~\ref{tab:example_of_manually_integrating} shows the examples of our manual approach to commonsense knowledge integration. In the first example, the knowledge about uv is integrated with the template of Synonym and relative clauses. Likewise, required knowledge is integrated using the templates and relative clauses in the second and third examples.

\begin{algorithm}
    \label{algo:man_itg}
    \begin{algorithmic}[1]
        \caption{Pseudo code for the knowledge integration procedure}
        \Require An input question (\textit{Q}), An input context (\textit{C}), target knowledge (\textit{K}) and templates (\textit{T})
        \State $Q^* \gets Q$; $C^* \gets C$
        \For{$K_i$ in $K$}
            \State $Subject, Relation, Object \gets K_i$
            \If{$(Q^*, C^*)$ in ReCoRD}
                \State $T_{Relation} \gets T[Relation]$
                \State $T^*_{Relation} \gets Replace(T_{Relation},``[[SUBJ]]", Subject)$
                \State $T^*_{Relation} \gets Replace(T^*_{Relation},``[[OBJ]]", Object)$
                \State $C^* \gets C^*+``@highlight''+T^*_{Relation}$
            \ElsIf {$Subject$ in $Q$ and $Object$ in $C$}
                \State $Q^* \gets IntegrateKnowledgeIntoText(Q^*, K_i, T)$
            \ElsIf{$Subject$ in $C$ and $Object$ in $Q$}
                \State $C^* \gets IntegrateKnowledgeIntoText(C^*, K_i, T)$
            \EndIf
            % \State $C_i \gets SelectContext(w_i, W - Answer, E)$
            % \State $A_i \gets SelectContext(w_i, Answer, E)$
            % \State $S_{w_i} \gets GetSenses(w_i)$
            % \State $S_{C_i} \gets GetSenses(C_i)$
            % \State $G_i \gets ConstructSubgraph(S_{w_i}, S_{C_i}, A_i)$
            % \State $\textbf{Pr} \gets PPR(G_i)$
            % \State $\hat{s}^* \gets max_{s_j \in S_{w_i}} \textbf{Pr}(s_j)$
            % \State $Answer \gets Answers || \hat{s}^*$ 
        \EndFor             
        \State \textbf{Return} $(Q^*, C^*)$
    \end{algorithmic}
\end{algorithm}

\begin{algorithm}
    \label{algo:IKT}
    \begin{algorithmic}[1]
        \caption{Pseudo code for the \textit{IntegrateKnowledgeIntoText} function}
        \Require An input text (\textit{I}), An input knowledge triple (\textit{k}), and templates (\textit{T})
        \State $Subject, Relation, Object \gets k$
        \State $i \gets FindFirstIndex(Subject, I)$
        \State $T_{Relation} \gets T[Relation]$
        \State $T^*_{Relation} \gets Replace(T_{Relation},$``[[SUBJ]]",``which"$)$
        \State $T^*_{Relation} \gets Replace(T_{Relation},$ ``[[OBJ]]"$, Object)$
        \State $I^* \gets I[:i]+T^*_{Relation}+I[i+1:]$
            % \State $C_i \gets SelectContext(w_i, W - Answer, E)$
            % \State $A_i \gets SelectContext(w_i, Answer, E)$
            % \State $S_{w_i} \gets GetSenses(w_i)$
            % \State $S_{C_i} \gets GetSenses(C_i)$
            % \State $G_i \gets ConstructSubgraph(S_{w_i}, S_{C_i}, A_i)$
            % \State $\textbf{Pr} \gets PPR(G_i)$
            % \State $\hat{s}^* \gets max_{s_j \in S_{w_i}} \textbf{Pr}(s_j)$
            % \State $Answer \gets Answers || \hat{s}^*$ 
        \State \textbf{Return} $I^*$
    \end{algorithmic}
\end{algorithm}

\end{document}

%\noindent In this appendices, we provide supplementary materials to "".  
\onecolumn
\section{Appendix A: Details on the Templates}

\begin{table*}[!th]
  \centering
  \caption{This table presents details on the templates utilized in our paper. Here, we analyze 37 relations in ConceptNet \cite{speer2017conceptnet}. }
  \label{tab:template}
  %\begin{center}
   %\begin{adjustwidth}{-1cm}{}

  \begin{tabular}{c|m{8.5cm}|c}
    \hline
    \textbf{Relation} & \multicolumn{1}{c|}{\textbf{Template}} & \# of samples \\
    \hline\hline
        RelatedTo & [[SUBJ]] is related to [[OBJ]] . & 287,459 \\ 
    HasContext & [[SUBJ]] is used in the context of [[OBJ]] . & 113,066 \\ 
    IsA & [[SUBJ]] is a [[OBJ]] . & 74,316 \\ 
    DerivedFrom & [[OBJ]] is derived from [[SUBJ]] . &  69,510\\     Synonym & [[SUBJ]] and [[OBJ]] are same . & 28,379 \\ 
    FormOf & [[OBJ]] is the root word of [[SUBJ]] . & 27,208\\ 
    EtymologicallyRelatedTo & [[SUBJ]] is etymologically related to [[OBJ]] . & 10,187\\ 
    SimilarTo & [[SUBJ]] is similar to [[OBJ]] . & 8,384 \\ 
    AtLocation & Something you find at [[OBJ]] is [[SUBJ]] . & 7,644 \\ 
    MannerOf & [[SUBJ]] is a way to [[OBJ]] . & 6,230 \\ 
    PartOf & [[SUBJ]] is part of [[OBJ]] . & 5,320 \\ 
    Antonym & [[SUBJ]] and [[OBJ]] are opposite . & 3,932 \\ 
HasProperty & [[SUBJ]] can be [[OBJ]] . & 2,886  \\ 
    UsedFor & [[SUBJ]] may be used for [[OBJ]] . & 2,145 \\ 
    DistinctFrom & [[SUBJ]] is not [[OBJ]] . & 1,256 \\ 
    HasPrerequisite & [[SUBJ]] requires [[OBJ]] . & 1,142 \\ 
    HasSubevent & When [[SUBJ]] , [[OBJ]] . & 1,119 \\ 
    Causes & [[SUBJ]] causes [[OBJ]] . & 999 \\ 
    HasA & [[SUBJ]] contains [[OBJ]] . & 943 \\ 
    InstanceOf & [[SUBJ]] is an instance of [[OBJ]] . & 902 \\ 
    CapableOf & [[SUBJ]] can [[OBJ]] . & 697 \\ 
    ReceivesAction & [[SUBJ]] can be [[OBJ]] . & 658 \\ 
    MotivatedByGoal & You would [[SUBJ]] because [[OBJ]] . & 603 \\ 
    CausesDesire & [[SUBJ]] would make you want to [[OBJ]] . & 556 \\ 
    MadeOf & [[SUBJ]] can be made of [[OBJ]] . & 316 \\ 
    HasLastSubevent & The last thing you do when you [[SUBJ]] is [[OBJ]] . & 302 \\ 
    Entails & [[SUBJ]] entails [[OBJ]] . & 298 \\ 
    HasFirstSubevent & The first thing you do when you [[SUBJ]] is [[OBJ]] . & 280 \\ 
    Desires & [[SUBJ]] wants [[OBJ]] . & 200 \\ 
    NotHasProperty & [[SUBJ]] is not [[OBJ]] . & 161 \\ 
    CreatedBy & [[SUBJ]] is creatd by [[OBJ]] . & 118 \\ 
    DefinedAs & [[SUBJ]] can be defined as [[OBJ]] . & 80 \\ 
    NotDesires & [[SUBJ]] does not want [[OBJ]] . & 71 \\ 
    NotCapableOf & [[SUBJ]] can not [[OBJ]] . & 43 \\ 
    LocatedNear & [[SUBJ]] is typically near [[OBJ]] . & 36 \\ 
    EtymologicallyDerivedFrom & [[SUBJ]] is etymologically derived from [[OBJ]] . & 27 \\ 
    SymbolOf & [[SUBJ]] is an symbol of [[OBJ]] . & 4 \\ 

    \hline
  \end{tabular}
  %\end{center}
%  \end{adjustwidth}
\end{table*}

\newpage
\section{Appendix B: Qualitative Analysis for Probabilistic Distributions}
\begin{table*}[!ht]
 \centering
  \caption{Results of the $hits@K$ metric for each relation in ConceptNet. }
  \label{tab:results_on_each_relation}
\begin{tabular}{c|cccc|cccc}
\hline
\multicolumn{1}{c|}{\multirow{3}{*}{\textbf{Relations}}} & \multicolumn{8}{c}{$hits@K$} \\ \cline{2-9} 
\multicolumn{1}{c|}{} & \multicolumn{4}{c|}{BERT$_{base}$} & \multicolumn{4}{c}{BERT$_{large}$} \\ \cline{2-9} 
\multicolumn{1}{c|}{} & 1 & 5 & 10 & \multicolumn{1}{c|}{100} & 1 & 5 & 10 & 100 \\ \hline\hline
RelatedTo & 7.60 & 9.30 & 11.77 & 25.38 & 6.51 & 8.50 & 10.97 & 24.14 \\
HasContext & 6.79 & 16.17 & 22.38 & 48.90 & 6.91 & 15.84 & 22.13 & 47.57 \\
IsA & 0.46 & 1.56 & 2.27 & 15.57 & 0.41 & 1.19 & 1.89 & 11.67 \\
DerivedFrom & 0.14 & 5.77 & 10.70 & 31.47 & 0.11 & 3.41 & 6.90 & 23.42 \\
Synonym & 16.16 & 27.33 & 33.12 & 52.70 & 13.38 & 26.74 & 34.69 & 56.39 \\
FormOf & 0.57 & 20.10 & 28.08 & 42.41 & 2.84 & 32.39 & 38.68 & 48.76 \\
EtymologicallyRelatedTo & 5.39 & 8.35 & 10.71 & 22.45 & 3.69 & 6.59 & 9.22 & 21.70 \\
SimilarTo & 1.60 & 4.39 & 6.09 & 14.92 & 2.84 & 7.13 & 10.13 & 23.61 \\
AtLocation & 2.03 & 3.72 & 5.41 & 23.36 & 3.04 & 5.89 & 8.93 & 32.28 \\
MannerOf & 2.66 & 5.05 & 8.77 & 35.71 & 2.17 & 5.85 & 9.61 & 36.25 \\
PartOf & 21.05 & 34.37 & 40.91 & 59.43 & 24.38 & 37.18 & 43.30 & 58.97 \\
Antonym & 17.14 & 25.70 & 32.38 & 53.69 & 28.26 & 34.55 & 40.65 & 63.26 \\
HasProperty & 3.22 & 8.39 & 12.14 & 38.04 & 5.23 & 12.93 & 17.75 & 46.14 \\
UsedFor & 12.87 & 16.50 & 21.44 & 47.16 & 12.26 & 14.78 & 19.25 & 45.72 \\
DistinctFrom & 1.67 & 4.36 & 6.75 & 23.70 & 5.10 & 11.09 & 15.22 & 37.81 \\
HasPrerequisite & 11.30 & 10.56 & 14.73 & 37.29 & 13.75 & 13.35 & 17.93 & 40.54 \\
HasSubevent & 1.79 & 2.55 & 4.03 & 16.20 & 2.32 & 3.39 & 5.11 & 18.40 \\
Causes & 9.71 & 12.73 & 17.05 & 40.79 & 10.81 & 13.90 & 18.65 & 45.81 \\
HasA & 4.24 & 10.55 & 15.17 & 40.35 & 4.67 & 9.75 & 14.19 & 37.22 \\
InstanceOf & 0.00 & 5.93 & 10.29 & 22.43 & 0.11 & 4.92 & 11.12 & 31.92 \\
CapableOf & 10.04 & 17.20 & 24.27 & 53.13 & 12.34 & 22.90 & 28.19 & 52.54 \\
ReceivesAction & 12.01 & 28.12 & 36.51 & 71.44 & 14.89 & 30.52 & 38.85 & 72.45 \\
MotivatedByGoal & 0.00 & 1.07 & 2.37 & 17.90 & 0.00 & 0.17 & 0.76 & 17.74 \\
CausesDesire & 4.32 & 11.52 & 17.59 & 57.25 & 2.34 & 7.54 & 13.95 & 52.13 \\
MadeOf & 12.34 & 44.12 & 51.85 & 72.94 & 18.67 & 42.22 & 50.63 & 75.05 \\
HasLastSubevent & 8.61 & 16.30 & 22.85 & 58.73 & 10.60 & 18.04 & 25.09 & 62.30 \\
Entails & 2.01 & 4.53 & 7.38 & 22.20 & 2.35 & 4.53 & 6.88 & 24.27 \\
HasFirstSubevent & 12.86 & 23.96 & 29.38 & 63.99 & 17.50 & 29.79 & 37.56 & 71.55 \\
Desires & 4.00 & 7.52 & 7.57 & 50.90 & 7.50 & 9.47 & 11.12 & 50.17 \\
NotHasProperty & 4.35 & 14.29 & 18.32 & 42.24 & 6.83 & 23.29 & 27.64 & 60.87 \\
CreatedBy & 2.54 & 9.75 & 15.25 & 35.88 & 0.85 & 5.08 & 10.17 & 29.52 \\
DefinedAs & 0.00 & 2.50 & 3.75 & 17.92 & 2.50 & 4.17 & 10.42 & 33.75 \\
NotDesires & 1.41 & 0.28 & 2.25 & 8.74 & 1.41 & 1.69 & 3.66 & 12.94 \\
NotCapableOf & 16.28 & 32.56 & 41.86 & 73.84 & 18.60 & 27.91 & 40.12 & 76.74 \\
LocatedNear & 2.78 & 8.33 & 13.89 & 36.11 & 5.56 & 8.33 & 8.33 & 25.00 \\
EtymologicallyDerivedFrom & 0.00 & 0.00 & 0.00 & 0.00 & 0.00 & 0.00 & 0.00 & 3.70 \\ 
\multicolumn{1}{c|}{SymbolOf} & \multicolumn{1}{c}{0.00} & \multicolumn{1}{c}{50.00} & \multicolumn{1}{c}{50.00} & \multicolumn{1}{c|}{50.00} & \multicolumn{1}{c}{25.00} & \multicolumn{1}{c}{50.00} & \multicolumn{1}{c}{50.00} & \multicolumn{1}{c}{50.00}\\\hline
\end{tabular}
\end{table*}

\newpage
\section{Appendix C: Details on the Reading Comprehension Question Types}

% \begin{table*}[!ht]
% \centering
%   \caption{Results of the $hits@K$ metric for each relations of ConceptNet. }
%   \label{tab:results_on_each_relation}
% \begin{tabular}{c|c|c}

% \hlinee
\begin{table*}[!ht]
 \centering
  \caption{Examples and descriptions for the question type of the \textit{has answer} questions. The main evidences for the categorization of the questions are colored. }
  \label{tab:results_on_each_relation}
\begin{tabular}{c|m{0.30\columnwidth}|m{0.40\columnwidth}}\hline
Question Types &  \multicolumn{1}{c|}{Description} & \multicolumn{1}{c}{Example} \\\hline\hline
Synonymy & There is a clear correspondence between question and context.& \begin{tabular}{@{}p{7cm}@{}}\textbf{Question}: Which entity is the \textbf{\textcolor{red}{secondary}} legislative body?\\\textbf{Context}: ... The \textbf{\textcolor{blue}{second main}} legislative body is the Council, which is composed of different ministers of the member states. ...\end{tabular}  \\\hline
\begin{tabular}[c]{@{}c@{}}Common sense\\ knowledge\end{tabular} & Common sense knowledge is required to solve the question. & \begin{tabular}{@{}p{7cm}@{}}\textbf{Question}: Where is the \textcolor{red}{\textbf{Asian}} influence strongest in Victoria?\\\textbf{Context}: ... Many \textcolor{blue}{\textbf{Chinese}} miners worked in Victoria, and their legacy is particularly strong in Bendigo and its environs. ...\end{tabular}\\\hline
No semantic variation & There is no semantic variation such as synonymy or common sense knowledge. & \begin{tabular}{@{}p{7cm}@{}}\textbf{Question}: Who are the \textbf{\textcolor{red}{un-elected subordinates of member state governments}}?\\\textbf{Context}: ... This means Commissioners are, through the appointment process, the \textbf{\textcolor{blue}{unelected subordinates of member state governments}}. ...\end{tabular} \\\hline
Multi-sentence reasoning & Hints for solving questions are shattered in multiple sentences.  & \begin{tabular}{@{}p{7cm}@{}}\textbf{Question}: Why did \textcolor{red}{\textbf{France}} choose to give up continental lands?\\\textbf{Context}: ...  \textcolor{blue}{\textbf{France}} chose to cede the former, ... \textcolor{blue}{\textbf{They}} viewed the economic value of the Caribbean islands' sugar cane ...\end{tabular} \\\hline
Others & The labeled answer is incorrect. &  \begin{tabular}{@{}p{7cm}@{}}\textbf{Question}: Who \textcolor{red}{\textbf{won the battle}} of Lake George?\\\textbf{Context}: ...  The \textcolor{blue}{\textbf{battle ended inconclusively}}, with both sides withdrawing from the field.  ...\end{tabular} \\\hline
Typo & There exist typing errors in the question or context. & \begin{tabular}{@{}p{7cm}@{}}\textbf{Question}: What kind of measurements define \textbf{\textcolor{red}{accelerlations}}?\\\textbf{Context}... \textbf{\textcolor{blue}{Accelerations}} can be defined through kinematic measurements. ...\end{tabular}\\\hline
\end{tabular}
\end{table*}

\bibliographystyle{aaai}\bibliography{references.bib}